\documentclass[sigconf, nonacm, table]{acmart} % source: 
\pdfoutput=1

\AtBeginDocument{%
  \providecommand\BibTeX{{%
    \normalfont B\kern-0.5em{\scshape i\kern-0.25em b}\kern-0.8em\TeX}}}

\setcopyright{acmcopyright}
\copyrightyear{2020}
\acmYear{2020}
\acmDOI{TBD}

\usepackage{lipsum}
\usepackage{color, soul}

\usepackage{enumerate}
\usepackage{bbm}
\usepackage{amsmath}
\usepackage{amsmath,amsthm} % amssymb % error: https://pctex.com/kb/86.html
\usepackage{mathtools}
\usepackage{listings}
\usepackage{multirow}
\usepackage{hhline}
\usepackage{changepage} % used for adjustwidth (for large tables)
\usepackage{adjustbox}
\usepackage{subcaption} % subfigure is obsolete and doesn't work with listings; using subcaption instead
\usepackage{url}
\usepackage{wrapfig}
\usepackage{booktabs}
\usepackage{siunitx}
\newcommand{\argmin}{\mathop{argmin}}
\newcommand{\argmax}{\mathop{argmax}}

\usepackage{caption} 
\usepackage{tabularx}

\usepackage{enumitem}

\usepackage{fontawesome}

\newcommand{\tabldata}{{\small \faTable}}
\newcommand{\imagdata}{{\footnotesize \faCamera}}
\newcommand{\textdata}{{\footnotesize \faFileText}}

\newcommand{\zvec}[1]{\boldsymbol{#1}} % for normal variables

\newcommand{\zrv}[1]{\mathrm{#1}} % for random variables (assume no vectors)

\newcommand{\htheta}{h}
\newcommand{\dist}{\texttt{dist}}
\newcommand{\cost}{\texttt{cost}}

\newcommand{\variable}[1]{\textsc{#1}}

\newcommand{\xF}{\zvec{x}^\texttt{F}}
\newcommand{\xFScalar}{x^\texttt{F}}

\newcommand{\xCF}{\zvec{x}^\texttt{CF}}
\newcommand{\xCFnearest}{\zvec{x}^\texttt{*CF}}

\newcommand{\xCFeps}{\zvec{x}^{\texttt{CF}}_{\epsilon}}

\newcommand{\xSCFScalar}{x^\texttt{SCF}}

\newcommand{\xCFE}{\zvec{x}^\texttt{CFE}}

\newcommand{\xSCF}{\zvec{x}^\texttt{SCF}}

\newcommand{\xAny}{\zvec{x}}

\newcommand{\doop}{\mathrm{do}}

\newcommand{\deltaVector}{\zvec{\delta}}
\newcommand{\deltaVectorStar}{\zvec{\delta}^*}
\newcommand{\deltaScalar}{\delta}

\newcommand{\thetaVector}{\zvec{\theta}}

\newcommand{\thetaScalar}{\theta}

\newcommand{\actionVector}{\zvec{a}}
\newcommand{\actionVectorStar}{\zvec{a}^*}

    \newcommand{\amirm}[1]{}
    \newcommand{\amirr}[1]{}
\newcommand{\distance}{$\delta^*$}
\newcommand{\coverage}{$\Omega^*$}
\newcommand{\realtime}{$\tau^*$}

\usepackage{thmtools, thm-restate}

\theoremstyle{definition}

\newcommand{\yy}{\fullcircblack} % {\checkmark}
\newcommand{\ex}{\fullcircgray}
\newcommand{\pp}{\halfcircblack} % {$\sim$}
\newcommand{\xx}{\emptycircblack} % {x}

\newcommand{\citeartelt}{\citep{artelt2019efficient, artelt2019computation, artelt2020convex}}

\newcommand*\fullcircblack[1][.9ex]{\tikz\fill (0,0) circle (#1);}
\newcommand*\fullcircgray[1][.93ex]{\tikz\fill[gray!75] (0,0) circle (#1);}
\newcommand*\halfcircblack[1][.88ex]{%
  \begin{tikzpicture}
    \draw[fill][gray!75] (0,0)-- (90:#1) arc (90:270:#1) -- cycle ;
    \draw (0,0) circle (#1);
  \end{tikzpicture}
}
\newcommand*\emptycircblack[1][.9ex]{\tikz\draw[gray!75] (0,0) circle (#1);}

\usepackage{tikz}
\usetikzlibrary{shapes,decorations,arrows,calc,arrows.meta,fit,positioning}
\tikzset{
    -Latex, auto, node distance = 0.5 cm and 0.5 cm, semithick,
    state/.style = {circle, draw, minimum width = 0.6 cm},
    inter/.style = {rectangle, draw, minimum width = 0.7 cm, minimum height = 0.7 cm},
    point/.style = {circle, draw, inner sep = 0.04cm, fill, node contents = {}},
    bidirected/.style = {Latex-Latex,dashed},
    el/.style = {inner sep=2pt, align=left, sloped}
}

\usetikzlibrary{backgrounds}

\usepackage{tikz}
\usetikzlibrary{trees,calc,fadings,decorations.pathreplacing, intersections}

\tikzset{%
  >=latex, % option for nice arrows
  inner sep=0pt,%
  outer sep=2pt,%
  mark coordinate/.style={inner sep=0pt,outer sep=0pt,minimum size=3pt,
    fill=black,circle}%
}

\newcommand{\ballF}{\mathcal{B}^\Delta_{\xF}}

\begin{document}

\rowcolors{2}{gray!20}{white}

% \title{SoK: A technical review of Algorithmic Recourse:\\Counterfactual Explanations and Consequential Interventions}
% \title{SoK: an overview of algorithmic recourse:\\definitions, formulations, solutions, and prospects}
% \title{Systematization of knowledge of algorithmic recourse:\\definitions, formulations, solutions, and prospects}
% \title{A survey of algorithmic recourse:\\definitions, formulations, solutions, and prospects}
\title{A survey of algorithmic recourse:\\contrastive explanations and consequential recommendations}


% \author{author}
% \email{email}
% % % \orcid{1234-5678-9012}
% \affiliation{%
%   \institution{institution}
%   \city{city}
%   \country{country}
% }

\author{Amir-Hossein Karimi}
% \email{amirhkarimi@gmail.com}
\affiliation{%
    \institution{MPI for Intelligent Systems}
    % \institution{MPI-IS}
    % \country{Germany}
    \country{}
}
\affiliation{
    \institution{ETH Z\"urich}
    % \country{Switzerland}
    \country{}
}
\author{Gilles Barthe}
\affiliation{%
   \institution{MPI for Security and Privacy}
%   \institution{MPI-SP}
%   \country{Germany}
    \country{}
}
\author{Bernhard Sch\"olkopf}
\affiliation{%
   \institution{MPI for Intelligent Systems}
%   \institution{MPI-IS}
%   \country{Germany}
    \country{}
}
\author{Isabel Valera}
\affiliation{
    \institution{MPI for Intelligent Systems}
    % \country{Germany}
    \country{}
}
\affiliation{
    \institution{Saarland University}
    % \country{Germany}
    \country{}
}

\renewcommand{\shortauthors}{Karimi, Barthe, Sch\"olkopf, Valera}

\begin{abstract}
Machine learning is increasingly used to inform decision-making in sensitive situations where decisions have consequential effects on individuals' lives.
% , e.g., pre-trial bail, loan approval, resume filtering, or prescription of life-altering medication.
%
In these settings, in addition to requiring models to be accurate and robust, socially relevant values such as fairness, privacy, accountability, and explainability play an important role in the adoption and impact of said technologies. % in determining the adoption...
In this work, we focus on \emph{algorithmic recourse}, which is concerned with providing \emph{explanations} and \emph{recommendations} to individuals who are unfavorably treated by automated decision-making systems.
We first perform an extensive literature review, and align the efforts of many authors by presenting unified \emph{definitions}, \emph{formulations}, and \emph{solutions} to recourse.
% \footnote{NeurIPS 2020 Workshop: ML Retrospectives, Surveys & Meta-Analyses (ML-RSA).}
% \footnote{The FAccT program chairs confirmed upon request that survey/position papers are within scope.}
%
% Then, we provide an overview of \hl{the \emph{promises} and \emph{problems} of existing recourse} approaches, and present \emph{prospects} towards which the community may engage, making explicit connections to \hbox{other ethical challenges such as security, privacy, and fairness.} % holistic summary
Then, we provide an overview of the \emph{prospective} research directions towards which the community may engage, challenging existing assumptions and making explicit connections to other ethical challenges such as security, privacy, and fairness. % holistic summary

% \footnotetext{In email correspondences with the FAccT Program Chairs, we confirmed that the material presented in this SoK is within the scope of this year's CFP.}
% \footnotetext{The FAccT program chairs confirmed upon request that survey/position papers are within scope.}
\end{abstract}

\maketitle

\section{Introduction}
\label{sec:010introduction}
% \bernhard{I wonder if we have to add something, e.g. a footnote, to the effect that you have asked the program chair and survey papers are indeed welcome. Otherwise some clever reviewer might try to construct a case why the paper should be rejected (and in such cases, people are reluctant to change their scores even if they are corrected).}\amir{Thank you for the comment, Bernhard. Does my footnote addition suffice?}
Consider the following setting: a 28-year-old female professional working as a software engineer seeks a mortgage to purchase a home.
Consider further that the loan-granting institution (e.g., bank) uses a binary classifier and denies the individual the loan based on her attributes.
Naturally, in this context, answering the following questions become relevant to the individual:

% \vspace{-3mm}
\begin{enumerate}
    \itemsep0em
    \item[\textbf{Q1.}] Why was I rejected the loan?
    \item[\textbf{Q2.}] What can I do in order to get the loan in the future?
    % \item How can I change my situation to get the loan in the future?
\end{enumerate}
% \vspace{-3mm}

\noindent In the setting of the example above, unless the policy of the bank is relaxed, the individual must expend effort to change their situation to be favourably treated by the decision-making system.
Examples such as the above are prevalent not only in finance \citep{mukerjee2002multi, barocas2017fairness} but also in justice (e.g., pretrial bail) \citep{angwin2016machine}, healthcare (e.g., prescription of life-altering medication) \citep{bastani2017interpretability, grote2020ethics, begoli2019need}, and other settings (e.g., hiring) \citep{nabi2018fair, cohen2019efficient, schumann2020we}
broadly classified as \emph{consequential decision-making} settings \citep{barocas2017fairness, karimi2020model, burrell2016machine, corbett2018measure}.
Given the rapid adoption of automated decision-making systems in these settings, designing models that not only have high objective accuracy but also afford the individual with \emph{explanations} and \emph{recommendations} to favourably change their situation is of paramount importance, and even argued to be a legal necessity (GDPR \citep{voigt2017eu}). % change their\newpage\noindent situation
This is the concern of \emph{algorithmic recourse}.
\amirm{timely problem}
% Along with a \hl{rush to adopt} ML-based systems in consequential decision-making settings , there has been a flurry of philosophical discussion and technical approaches for answering the above questions to endow individuals with recourse.
%
% Given the timeliness of this increased attention (see GDPR \citep{voigt2017eu}),

% In reviewing the literature, there seems to be a need to consolidate definitions, construct technical baselines for future comparison, and situate recourse in the broader ethical ML literature.
%
% \hl{Our work identifies a number of points where the situation could be made better and contributes in this direction.}
%
% Specifically, inconsistent use of definitions and notation, a lack of technical baselines, and an absence of discussion around situating recourse in the broader ethical ML literature may impede necessary progress in this topic.
% \amir{TODO: add small graph of number of papers?}

%  granting individuals recourse from (semi-)automated systems \hl{cite Jenna Burell?}

% \input{025figure}

% \newpage
% \subsection{Survey Scope \& Process}
% \vspace{2mm}
\amirm{contributions}
\textbf{Contributions:}
Our review brings together the plethora of recent works on algorithmic recourse.
In reviewing the literature, we identify opportunities to consolidate definitions, construct technical baselines for future comparison, and situate recourse in the broader ethical ML literature. % there seemed to be a need
The primary contribution is thus a unified overview of the definitions (\S\ref{sec:020background}), formulations (\S\ref{sec:030formulation}) and technical solutions (\S\ref{sec:040solution}) for computing \emph{(nearest) contrastive explanations} and \emph{(minimal) consequential recommendations}, across a broad range of setups.
A visual summary is curated and presented in Table ~\ref{table:technical_summary}.
Clearly distinguishing between the two recourse offerings above and presenting their often overlooked different technical treatment is a common theme throughout this work.
%
% The secondary contribution of our review is to provide a holistic \hl{summary of the \emph{promises} and \emph{problems} of existing} approaches to recourse, and to present \emph{prospects} towards which the community may engage (\S\ref{sec:050discussion}).
% The secondary contribution of our review is to
Additionally, we identify a number of prospective research directions which challenge the assumptions of existing setups, and present extensions to better situate recourse in the broader ethical ML literature (\S\ref{sec:050discussion}).

\textbf{Who is this document for?}
% covers a number of topics related to recourse and
This document aims to engage different audiences: practitioners (\variable{P}), researchers (\variable{R}), and legal scholars (\variable{L}).
\S\ref{sec:020background} builds on the rich literature in the philosophy of science and presents a clear distinction between contrastive explanations and consequential recommendations based on the causal history of events (\variable{R}, \variable{L}).
Table \ref{table:technical_summary} presents an overview of 50+ technical papers that offer recourse in a broad range of setups (\variable{P}, \variable{R}).
\S\ref{sec:030formulation} formulates the recourse problem as constrained optimization and present the variety of constraints needed to model real-world settings (\variable{R}).
\S\ref{sec:040solution} presents the difficulty in solving for recourse in general settings, and summarizes existing approximate solutions that trade-off various desirable properties (\variable{P}, \variable{R}, \variable{L}).
Finally, \S\ref{sec:050discussion} argues that recourse should be considered in relation to other ethical ML considerations (\variable{L}), and outlines future research directions (\variable{R}). % broadens the scope and argues

% \amirm{remove?}
% \noindent \textbf{Survey Process:} Our literature survey consisted of a search through past works for terms such as ``algorithmic recourse'', ``counterfactual explanation'', ``contrastive explanation'', or ``consequential recommendation''.
% % in their title or abstract.
% %
% For this, we used various search engines, primarily GScholar,\footnote{\url{https://scholar.google.com/}} arXiv,\footnote{\url{https://arxiv.org}} Connected Papers,\footnote{\url{https://www.connectedpapers.com/}} and dblp,\footnote{\url{https://dblp.uni-trier.de/}} arriving at an initial set of \hl{??} papers.
% %
% From this initial set of papers, we retain those that formulate the problem and provide an algorithm with accompanying experiments, and present them jointly in Table~\ref{table:technical_summary}.
% %
% Most remaining papers were intermittently used to guide definitions and discussions.

\amirm{scope; what is \& what is not}
\textbf{Scope:}
% in addition to surveying the recourse literature,
To adequately present the material, we provide brief introductions to various related topics, such as explainable machine learning (or XAI), causal and counterfactual inference, and optimization, among others.
Our work is not meant to be a comprehensive review of these topics, and merely aims to situate algorithmic recourse in the context of these related fields.
For the interested reader, an in-depth discussion on the social, philosophical, cognitive, and psychological aspects of explanations can be found at \citep{miller2019explanation, barocas2020hidden, venkatasubramanianphilosophical, byrne2019counterfactuals}.
% which we briefly present in \S\ref{sec:050discussion}.
%
For XAI, see \citep{mcgarry2005survey, freitas2014comprehensible, lipton2018mythos, biran2017explanation, gilpin2018explaining, abdul2018trends, guidotti2018survey, adadi2018peeking, doshi2017towards, montavon2018methods}, for causal and counterfactual inference, see \citep{pearl2010foundations, gelman2011causality, pearl2016causal, scholkopf2019causality, moraffah2020causal, guo2018survey, sep-causation-mani}, and for optimization, see \citep{nocedal2006numerical, boyd2004convex, sra2012optimization}.

% Given the breadth of topics involved, it is quite likely that we have inadvertently missed key references relating to either the main or supplementary topics.
% %
% We apologize in advance for such shortcomings, and look forward to hearing your feedback on how to improve the manuscript.

% \amir{would be useful to consult some folks to properly define \emph{recourse}, \emph{agency}, \emph{explanation}, \emph{reasoning}, \emph{attribution}, etc.
% \begin{itemize}
%     \item [Recourse] Suresh V., Salon Barocas, Berk Ustun
%     \item [Philos.] Tim Miller, Eric Raidl, David Danks, Tobias Gerstenberg
%     \item [Legal] Sandra W., Jenna Burrel, Finale D.-V., Cynthia Rudin
%     \item [Other] Raha Moraffah, Seyed-Mohsen Dezfooli, Krikamol, Mijung, Samira, Miriam, Julius, Kiarash, Patrick, Waleed, Manuel, Annika, Adrian
% \end{itemize}
% }

% % \noindent \textbf{Who is this document for?}
% % This document aims to
% \noindent \textbf{This document aims to:}
% provide guidance to practitioners in using existing recourse methods in their pipelines, inform researchers of shortcomings and to suggest ideas for extending to new domains, and assist policy-makers in better understanding the technical requirements and challenges for offering algorithmic recourse.
% % practitioners and model deployers who aim to

% \clearpage
\section{Background}
\label{sec:020background}

\subsection{Recourse definitions}
% \zmessage{recourse is guaranteed when two questions are answered.}
\amirm{3 def'ns}
In its relatively young field, \emph{algorithmic recourse} has been defined as, e.g.,
``an actionable set of changes a person can undertake in order to improve their outcome'' \citep{joshi2019towards};
``the ability of a person to obtain a desired outcome from a fixed model'' \citep{ustun2019actionable};
or ``the systematic process of reversing unfavorable decisions by algorithms and bureaucracies across a range of counterfactual scenarios'' \citep{venkatasubramanianphilosophical}.
{\color{black} Similarly, \emph{legal recourse} pertains to actions by individuals or corporations to remedy legal difficulties \citep{wallin1992legal}.}
{\color{black} Parallel to this, \emph{explanations} aim to, at least in part, assist data-subjects to ``understand what could be changed to receive a desired result in the future'' \citep{wachter2017counterfactual}.}
{\color{black} This plurality of overlapping definitions, and the evolving consensus therein, motivates us to attempt to present a unified definition under the umbrella of recourse.}
%
% Such definitions build o legal recourse 
% ADD A legal recourse is an action that can be taken by an individual or a corporation to attempt to remedy a legal difficulty (from wiki)
% From this : %https://www.jstor.org/stable/248023?seq=1#metadata_info_tab_contents
%
% {\color{black}the term recourse has been defined not only as an explanation \citep{wachter2017counterfactual} but also as the ``ability'' [171], the ``process'' [174], and the ``recommended changes'' [21, 87] to obtain the desired result.}

We submit that recourse can be achieved by an affected individual if they can \emph{understand} \citep{wachter2017counterfactual, joshi2019towards, downscruds} and accordingly \emph{act} \citep{karimi2020mintrecourse, karimi2020caterecourse} to alleviate an unfavorable situation, thus exercising temporally-extended agency \citep{venkatasubramanianphilosophical}.
\amirm{2 qqs}
Concretely, in the example of the previous section, recourse is offered when the loan applicant is given answers to both questions:
provided \emph{explanation(s)} as to why the loan was rejected (\textbf{Q1});
and offered \emph{recommendation(s)} on how to obtain the loan in the future (\textbf{Q2}).
Below, we describe the {\color{black} similarities and} often overlooked differences between these questions and the different set of assumptions and tools needed to sufficiently answer each in general settings.
%
% {\color{black} As we shall see, presenting a clear-cut distinction between recourse explanations and recommendations, without a notion of causality, remains elusive.}
% {\color{black} Such a distinction is requires a causal perspective, which we summarize below.}
{\color{black} Such a distinction is made possible by looking at the context from a causal perspective, which we summarize below.}

% Below we briefly review and clarify terminology, and describe the different set of assumptions needed for providing recourse.
% situating our argument in rich philosophical literature.

% \subsection{Philosophical foundations of recourse}
% \subsection{Recourse through \hl{a causal lens}}
%\subsection{Recourse through \hl{a philosophy lens}}
\subsection{Recourse and causality}
\label{sec:causal_recourse}

% In order to answer the two questions for recourse, we first remark on the form of explanations and recommendations used broadly in XAI (which subsumes consequential decision-making that focuses primarily on the data-subjects).

\subsubsection{\textbf{Contrastive Explanations}}
\amirm{def'n}
% \st{To provide explanations, we must first dissect the explanandum.}.\bernhard{I would tone it down a bit, and simply remove this (kind of) sentence - this will not help getting favorable reviews.}
%
In an extensive survey of the social science literature, \citet{miller2019explanation} concluded that when people ask ``Why P?'' questions {\color{black}(e.g., \textbf{Q1})}, they are typically asking ``Why P rather than Q?'', where Q is often implicit in the context \citep{hilton1990conversational, robeer2018contrastive}.
A response to such questions is commonly referred to as a \emph{contrastive explanation}, and is appealing for two reasons.
Firstly, contrastive \emph{questions} provide a `window' into the questioner's mental model, identifying what they had expected (i.e., Q, the contrast case), and thus, the \emph{explanation} can be better tuned towards the individual's uncertainty and gap in understanding \citep{miller2018contrastive}.
%
% Secondly, as will become evident below
Secondly, providing contrastive explanations may be ``simpler, more feasible, and cognitively less demanding" \citep{miller2018contrastive} than offering recommendations. % \st{to both questioner and explainer}'' \bernhard{simpler than what? Or is this a (meta-) contrastive explanation of what a contrastive explanation is? ;)}\amir{thank you (and isabel) for pointing this out. in reviewing the source, while not clear, it seems the author means that contrastive explanations are simpler relative to \emph{causal attribution}-based explanations, which also aim to explain the causal relations between variables. Perhaps a small fix (with a bit of extrapolation) would be: ``simpler, more feasible, and cognitively less demanding'' than providing recommendations.}
% from a practical perspective and as we shall see below
% are asked when an individual is surprised of an event (e.g., loan rejection), where the contrast case identifies what the individual expected and is thus

\subsubsection{\textbf{Consequential Recommendations}}
\amirm{def'n}
Providing an affected individual with recommendations {\color{black} (e.g., response to \textbf{Q2})} amounts to suggesting \emph{a set of actions} {\color{black}(a.k.a. \emph{flipsets} \citep{ustun2019actionable})} that should be performed to achieve a favourable outcome in the future.
%inferred actions directly from
In this regard, several works have used contrastive explanations to directly infer actionable recommendations \cite{joshi2019towards, ustun2019actionable, sharma2019certifai, wachter2017counterfactual} {\color{black} where actions are considered as independent shifts to the feature values of the individual (P) that results in the contrast (Q).}
%
% {\color{orange} however, \citet{karimi2020mintrecourse} show that contrastive explanations may result in infeasible or suboptimal actions in general settings.}
%
{\color{black} Recent work, however, has cautioned against this approach, citing the implicit assumption of independently manipulable features as a potential limitation that may lead to suboptimal or infeasible actions} \citep{venkatasubramanianphilosophical, barocas2020hidden, mahajan2019preserving, karimi2020mintrecourse}.
{\color{black} To overcome this,} \citet{karimi2020mintrecourse} suggest that actions may instead be interpreted as \emph{interventions} in a causal model of the world in which actions will take place{\color{black}, and not as independent feature manipulations derived from contrastive explanations.}
Formulated in this manner, e.g., using a structural causal model (SCM) \citep{pearl2000causality}, the down-stream effects of interventions on other variables in the model (e.g., descendants of the intervened-upon variables) can directly be accounted for when recommending actions \citep{barocas2020hidden}.
%
% \citet{karimi2020mintrecourse} suggest that actions can be interpreted as \emph{interventions} in a causal model of the inter-variable relations, e.g., in a structural causal model (SCM) \citep{pearl2000causality}.
%
% Interventions may be performed additively \citep{eberhardt2007interventions} or structurally \citep{pearl2016causal}, and may stochastically change the value of multiple variables conditioned on a variety of pre-/post-intervention feasibility conditions. % in an additive manner
%
% Irrespective of the \emph{form}, \emph{feasibility}, and \emph{scope} of interventions, actions as interventions may have down-stream effects on other variables in the model (e.g., descendants of the intervened-upon variables) which should be accounted for when recommending actions \citep{barocas2020hidden}.
%
Thus a recommended set of actions for recourse, in a world governed by a SCM, are referred to as \emph{consequential recommendations} \citep{karimi2020mintrecourse}.

\subsubsection{\textbf{Clarifying terminology: contrastive, consequential, \\ and counterfactual}}
\label{sec:clarifying_terminology}

To summarize, recourse explanations are commonly sought in a contrastive manner, and recourse recommendations can be considered as interventions on variables modelled using an SCM.
%we ask how such explanations/recommendations are generated?
% we can define
% \emph{contrastive explanations} -- ``how the world would have (had) to be different for a desirable outcome to occur'' --
% %
% and \emph{consequential recommendations} -- ``a recommendable set of actions that would result in a counterfactual instance with a favourable output when performed''.
%
Thus, we can rewrite the two recourse questions as:
\amirm{rewrite 2 qqs}
\begin{enumerate}
    % \item[\textbf{Q1.}] Who should I become in order to receive the loan? % \footnote{What profile would have received the loan?}
    \item[\textbf{Q1.}] What profile would have led to receiving the loan?
    \item[\textbf{Q2.}] What actions would have led me to develop this profile? \footnote{\color{black}A common assumption when offering recommendations is that the world is stationary; thus, actions that \emph{would have} led me to develop this profile had they been performed in the past, \emph{will} result in the same were they to be performed now. This assumption is challenged in \citep{rawal2020can, venkatasubramanianphilosophical} and discussed further in \S\ref{sec:beyond_individualized_recourse}.} % and changes to the individual's circumstances arise only through the actions of the individual
\end{enumerate}

% In layman's terms, recourse is offered when we ``inform an individual where they need to get to, and how to get there'' \citep{karimi2020mintrecourse}.
% %
% For instance, a diabetic patient may know that a lower blood pressure would reduce their risk of heart failure, but only a doctor with better understanding of the underlying processes could effectively suggest ``interventions that will move the patient out of [a high blood pressure] group''~\citep{wachter2017counterfactual}. % an at-risk
%
\amirm{intuitively}
Viewed in this manner, both contrastive explanations and consequential recommendations can be classified as a \emph{counterfactual}~\citep{mcgill1993contrastive}, in that each considers the alteration of an entity in the history of the event P, where P is the undesired model output.
Thus, responses to \textbf{Q1} (resp. \textbf{Q2}) may also be called \emph{counterfactual explanations} (resp. \emph{counterfactual recommendations}), meaning what could have been (resp. what could have been done) \citep{byrne2019counterfactuals}.

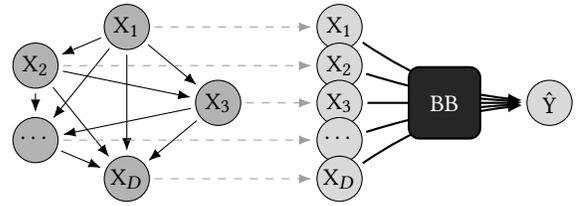
\begin{figure}[t]
    \begin{tikzpicture}
        [auto,
        block/.style ={rectangle, draw=black, thick, fill=black!85, text width=3em, text centered, text=white, rounded corners, minimum height=3em},
        line/.style ={draw, thick, -latex}]

        \matrix [column sep=6mm,row sep=0mm]
        {
        & \node[state, fill=gray!60] (x1g) {$\zrv{X}_1$} ; & & [4mm] \node[state, fill=gray!30] (x1b) {$\zrv{X}_1$} ; &                         &                                                       \\ [-1mm]% [-2mm]
        \node[state, fill=gray!60] (x2g) {$\zrv{X}_2$} ; & & & [4mm] \node[state, fill=gray!30] (x2b) {$\zrv{X}_2$} ; &                         &                                                       \\ [-3mm] % [-6mm]
        & & \node[state, fill=gray!60] (x3g) {$\zrv{X}_3$} ; & [4mm] \node[state, fill=gray!30] (x3b) {$\zrv{X}_3$} ; & \node[block] (bb) {BB}; & \node[state, fill=gray!30] (yhat) {$\hat{\zrv{Y}}$} ; \\ [-3mm] % [-6mm]
        \node[state, fill=gray!60] (xog) {$\cdots$} ;    & & & [4mm] \node[state, fill=gray!30] (xob) {$\cdots$} ;    &                         &                                                       \\ [-1mm]% [-2mm]
        & \node[state, fill=gray!60] (xDg) {$\zrv{X}_D$} ; & & [4mm] \node[state, fill=gray!30] (xDb) {$\zrv{X}_D$} ; &                         &                                                       \\
        };

        \path[solid] (x1g) edge (x2g);
        \path[solid] (x1g) edge (x3g);
        \path[solid] (x1g) edge (xog);
        \path[solid] (x1g) edge (xDg);

        \path[solid] (x2g) edge (x3g);
        \path[solid] (x2g) edge (xog);
        \path[solid] (x2g) edge (xDg);

        \path[solid] (x3g) edge (xog);
        \path[solid] (x3g) edge (xDg);

        \path[solid] (xog) edge (xDg);

        \begin{scope}[every path/.style=line, on background layer]
            % \path (x1b.east) -- ([yshift=+3mm]bb.west);
            % \path (x2b.east) -- ([yshift=+1mm]bb.west);
            % \path (x3b.east) -- ([yshift= 0mm]bb.west);
            % \path (xob.east) -- ([yshift=-1mm]bb.west);
            % \path (xDb.east) -- ([yshift=-3mm]bb.west);
            % \path (bb.east) -- (yhat.west);
            \path (x1b) edge [bend left  = -14] (yhat);
            \path (x2b) edge [bend left  =  -7] (yhat);
            \path (x3b) edge [bend left  =   0] (yhat);
            \path (xob) edge [bend left  =  +7] (yhat);
            \path (xDb) edge [bend left  = +14] (yhat);
        \end{scope}

        \path[dashed, gray!75] (x1g) edge (x1b);
        \path[dashed, gray!75] (x2g) edge (x2b);
        \path[dashed, gray!75] (x3g) edge (x3b);
        \path[dashed, gray!75] (xog) edge (xob);
        \path[dashed, gray!75] (xDg) edge (xDb);

    \end{tikzpicture}
    \vspace{-3mm}
    \caption{
    % \small % ARXIV
    \footnotesize % CSUR
    % {\color{black} placeholder placeholder placeholder placeholder }
    % % Illustration of the recourse setting;
    Variables $\{\zrv{X}_i\}_{i=1}^D$ capture observable characteristics of an individual which are fed into a blackbox (BB) decision-making system (\emph{mechanism}) yielding the prediction, $\hat{Y}$. % (dashed arrows are identity mechanisms),
    %
    % generating contrastive explanations relies only on the parametric form of a human-designed (albeit blackbox) decision-making system (which is fully specified), while generating consequential recommendations requires accurate knowledge of the SCM or a causal graph as a model of the nature itself (both of which are non-identifiable from observational data alone \citep{peters2017elements}).
    %
    Contrastive explanations are obtained via interventions on the BB inputs which can be seen as independent feature shifts that
    do not affect other variables.
    {\color{black} Conversely, consequential recommendations are interventions on a causal model of the world in which actions take place, and may have down-stream effects on other variables before being passed into the BB.
    Unlike the former, generating the latter relies on accurate knowledge of the SCM or causal graph as a model of nature itself (both of which are non-identifiable from observational data alone~\citep{peters2017elements}).
    In turn, offering consequential recommendations also implies contrastive explanations, but not vice versa (see \S3 for technical details).
    }
    %
    % Conversely, consequential recommendations are interventions on a model of the world in which actions will take place and thus rely on accurate knowledge of the SCM or causal graph as a model of nature itself (both of which are non-identifiable from observational data alone~\citep{peters2017elements}).
    % %
    % As a result, recourse recommendations are more difficult to generate than recourse explanations.
    }
    % relies only on the parametric form of a human-designed (albeit blackbox) decision-making system (which is fully specified), while generating consequential recommendations requires accurate knowledge of the SCM or a causal graph as a model of the nature itself
    %
    % Here, dashed arrows correspond to identity functions.
    % different levels of historical causal knowledge are needed to generate contrastive explanations and consequential recommendations (see text for details).}}
    \label{fig:contrastive_vs_consequential}
    \vspace{-3mm}
    % \vspace{-6mm}
\end{figure}

To better illustrate the difference between contrastive explanations and consequential recommendations, we refer to Figure \ref{fig:contrastive_vs_consequential}.
According to \citet[p. 217]{lewis1986causal}, ``to explain an event is to provide some information about its causal history''.
\citet[p. 256]{lipton1990contrastive} argues that in order to explain why P rather than Q, ``we must cite a causal difference between P and not-Q, consisting of a cause of P and the absence of a corresponding event in the history of not-Q''.
%
% In the recourse setting, P and Q can be considered as the fixed model's output, i.e., loan approval and rejection, respectively.
%
% Moreover
\amirr{causal history}
% In the XAI\bernhard{define - explainable AI?}
In the algorithmic recourse setting, because the model outputs are determined by its inputs (which temporally precede the prediction), the input features may be considered as the causes of the prediction.
The determining factor in whether one is providing contrastive explanations as opposed to consequential recommendations is thus the level at which the \emph{causal history} \citep{ruben2015explaining} is considered:
% Importantly, this definition
% Following this, the explainer does not need to reason about or even know about all causes of the fact — only those relative to the contrast case \citet{miller2018contrastive}.
%
% Specifically, whereas generating the former does not require knowledge of the causal relations between variables, generating the latter does.
% while explanations usually only go as far back as the model inputs, accounting for the consequences of actions requires modeling as far back as the causal generative process of the inputs including its dependence upon available actions. \iv{I would rewrite this sentence in terms of "while providing explanations only require information on the relationship between the model inputs and predictions, recommendations require information on the causal relationships among inputs."}\amir{thank you isabel!}
while providing explanations only requires information on the relationship between the model inputs, $\{\zrv{X}_i\}_{i=1}^D$, and predictions, $\hat{\zrv{Y}}$, recommendations require information as far back as the causal relationships among inputs. % (captured, e.g., in an SCM or a causal graph, both of which are non-identifiable from observational data alone \citep{peters2017elements}
The reliance on fewer assumptions \citep{miller2019explanation, lipton1990contrastive} thus explains why generating recourse explanations is easier than generating recourse recommendations {\color{black}\citep{wachter2017counterfactual, miller2018contrastive}}.

Next, we summarize the technical literature in Table \ref{table:technical_summary}.
%
% % ONLY FOR CSUR, NOT FOR ARXIV
% We find that the vast majority of the recourse literature has focused on generating contrastive explanations rather than consequential recommendations (c.f., \S\ref{sec:020background}).
% %
% Differentiable models are the most widely supported class of models, and many constraints are only sparsely supported (c.f., \S\ref{sec:030formulation}).
% %
% All tools generate solutions that to some extent trade-off desirable requirements, e.g., optimality, perfect coverage, efficient run-time, and access (c.f., \S\ref{sec:040solution}), resulting in a lack of unifying comparison (c.f., \S\ref{sec:050discussion}).
% %
% This table does not aim at serving to rank or be a qualitatively comparison of surveyed methods, and one has to exercise caution when comparing different setups.
% %
% % \hl{Furthermore, curating such a table is prone to errors, and we invite authors of the surveyed works to share corrections where applicable.}
% %
% %
% As a systematic organization of knowledge, we believe the table may be useful to practitioners looking for methods that satisfy certain properties, and useful for researchers that want to identify open problems and methods to further develop.
%
Offering recourse in diverse settings with desirable properties remains an open challenge, which we explore in the following sections.

% https://eprint.iacr.org/2019/1393.pdf

\renewcommand{\arraystretch}{0.98}

\begin{table*}%[htbp!]
  \setlength{\tabcolsep}{1.4pt} % ARXIV
  \caption{\footnotesize
  An overview of recourse algorithms for consequential decision-making settings is presented.
  Ordered chronologically, we summarize the \emph{goal}, \emph{formulation}, \emph{solution}, and \emph{properties} of each algorithm.
  %
  % With the over-arching goal of providing recourse, each paper either primarily offers contrastive explanations, \ee, or consequential recommendations, \rr.
  %
  % Table cells are populated with
  Symbols are used to indicate supported settings in the experimental section of the paper (\yy), settings that are natural extensions of the presented algorithm\protect\footnotemark[1] (\ex), settings that are partially supported\protect\footnotemark[2] (\pp\!\!), and settings that are not supported (\xx).
  The models cover a broad range of tree-based (TB), kernel-based (KB), differentiable (DF), or other (OT) types.
  Actionability contraints (unconditional or conditional), plausibility constraints (domain-, density-, and prototypical-consistency), and additional constraints (diversity, sprasity) are also explored.
  While the primary datatypes used in consequential settings are tabular \tabldata\, (involving a mix of numeric, binary, categorical, and ordinal variables), we also include additional works that generate recourse for non-tabular (images \imagdata \, and document \textdata) datasets.
  %
  % Furthermore, algorithms may additionally emphasize their aim to achieve optimality (opt.), perfect coverage (cov.), efficient run-time (rtm.) solutions. %  \divrsity, \sparsity
  Furthermore, papers that present analysis of such properties as optimality (opt.), coverage (cov.), and run-time complexity (rtm.) are specified in the table. %  \divrsity, \sparsity
  \hbox{Finally, we make note of those papers that provide \emph{open-source} implementations of their algorithm.}
  \\ \\ % ARXIV ONLY, MOVE TO MAIN BODY FOR CSUR
  Executive summary:
  % Some high-level trends are nonetheless observed:
  %
  the vast majority of the recourse literature has focused on generating contrastive explanations rather than consequential recommendations (c.f., \S\ref{sec:020background}).
  Differentiable models are the most widely supported class of models, and many constraints are only sparsely supported (c.f., \S\ref{sec:030formulation}).
  All tools generate solutions that to some extent trade-off desirable requirements, e.g., optimality, perfect coverage, efficient run-time, and access (c.f., \S\ref{sec:040solution}), resulting in a lack of unifying comparison (c.f., \S\ref{sec:050discussion}).
  This table does not aim at serving to rank or be a qualitatively comparison of surveyed methods, and one has to exercise caution when comparing different setups.
  %
  % \hl{Furthermore, curating such a table is prone to errors, and we invite authors of the surveyed works to share corrections where applicable.}
  %
  %
  As a systematic organization of knowledge, we believe the table may be useful to practitioners looking for methods that satisfy certain properties, and useful for researchers that want to identify open problems and methods to further develop.
  }
  \label{table:technical_summary}
   \footnotesize % ARXIV
%   \tiny % CSUR
  \begin{minipage}{\textwidth}
    \raggedright
    \begin{tabular}{ l l     c c c c  c c c  c c  c c c  c c    c c c c c c }
      \rowcolor{white}
      \toprule
      \multicolumn{2}{c}{\multirow{3}{*}{Algorithm}}
      & \multicolumn{14}{c}{Formulation}
      & \multicolumn{6}{c}{Solution} \\
      \rowcolor{white}
      \cmidrule(r){3-16}\cmidrule(r){17-22}
      \morecmidrules
      \cmidrule(r){3-16}\cmidrule(r){17-22}
                &
                & \multicolumn{4}{c}{Model} & \multicolumn{2}{c}{Actionability} & \multicolumn{3}{c}{Plausibility} & \multicolumn{2}{c}{Extra} & \multicolumn{3}{c}{Data types} % & \multirow{2}{*}{Goal}
                & \multirow{2}{*}{Tools} & \multirow{2}{*}{Access} & \multicolumn{3}{c}{Properties} & \multirow{2}{*}{Code} \\
                \rowcolor{white}
                \cmidrule(r){3-6}
                \cmidrule(r){7-8}
                \cmidrule(r){9-11}
                \cmidrule(r){12-13}
                \cmidrule(r){14-16}
                \cmidrule(r){19-21}
                \rowcolor{white}
                &                                                         & TB  & KB  & DF  & OT    & uncond. & cond. & dom. & dens. & proto. & diver.    & spar.     & \tabldata & \imagdata & \textdata &                      &                  & opt.      & cov.      & rtm.      &     \\ %      &
      \midrule
      % (??.??) & ?? \citep{??}                                           & \qq & \qq & \qq & \qq   & \qq     & \qq   & \qq  & \qq   & \qq    & \qq       & \qq       & \qq       & \qq       & \qq       & ??                   & ??               & \qq       & \qq       & \qq       & \qq \\ %  \ee &
      (2014.03) & SEDC \citep{martens2014explaining}                      & \ex & \yy & \ex & \ex   & \xx     & \xx   & \xx  & \xx   & \xx    & \xx       & \xx       & \xx       & \xx       & \yy       & heuristic            & query            & \yy       & \xx       & \yy       & \yy \\ %  \ee &
      (2015.08) & OAE \citep{cui2015optimal}                              & \yy & \xx & \xx & \xx   & \yy     & \ex   & \xx  & \xx   & \xx    & \xx       & \xx       & \yy       & \xx       & \xx       & ILP                  & white-box        & \yy       & \xx       & \yy       & \xx \\ %  \ee &
      (2016.05) & HCLS \citep{lash2017budget, lash2017generalized}        & \yy & \yy & \yy & \ex   & \yy     & \yy   & \yy  & \yy   & \xx    & \xx       & \xx       & \pp       & \xx       & \xx       & grad opt/heuristic   & gradient/query   & \yy       & \xx       & \yy       & \yy \\ %  \ee &
      (2017.06) & Feature Tweaking \citep{tolomei2017interpretable}       & \yy & \xx & \xx & \xx   & \xx     & \xx   & \xx  & \xx   & \xx    & \xx       & \xx       & \yy       & \xx       & \xx       & heuristic            & white-box        & \xx       & \xx       & \yy       & \yy \\ %  \ee &
      (2017.11) & CF Expl. \citep{wachter2017counterfactual}              & \xx & \xx & \yy & \xx   & \yy     & \xx   & \xx  & \xx   & \xx    & \yy       & \yy       & \pp       & \xx       & \xx       & grad opt             & gradient         & \xx       & \xx       & \xx       & \xx \\ %  \ee &
      (2017.12) & Growing Spheres \citep{laugel2017inverse}               & \yy & \yy & \ex & \ex   & \xx     & \xx   & \xx  & \xx   & \xx    & \xx       & \yy       & \pp       & \yy       & \xx       & heuristic            & query            & \xx       & \xx       & \xx       & \xx \\ %  \ee &
      (2018.02) & CEM \citep{dhurandhar2018explanations}                  & \xx & \xx & \yy & \xx   & \xx     & \xx   & \yy  & \xx   & \xx    & \xx       & \yy       & \pp       & \yy       & \xx       & FISTA                & class prob.      & \xx       & \xx       & \xx       & \yy \\ %  \ee &
      (2018.02) & POLARIS \citep{zhang2018interpreting}                   & \xx & \xx & \yy & \xx   & \yy     & \xx   & \xx  & \xx   & \xx    & \xx       & \xx       & \yy       & \yy       & \xx       & heuristic            & gradient         & \xx       & \xx       & \yy       & \yy \\ %  \ee &
      (2018.05) & LORE \citep{guidotti2018local}                          & \yy & \yy & \yy & \ex   & \xx     & \xx   & \yy  & \xx   & \xx    & \xx       & \xx       & \yy       & \xx       & \xx       & gen alg + heuristic  & query            & \yy       & \xx       & \yy       & \yy \\ %  \ee &
      (2018.06) & Local Foil Trees \citep{van2018contrastive}             & \yy & \yy & \yy & \ex   & \xx     & \xx   & \xx  & \xx   & \xx    & \xx       & \xx       & \yy       & \xx       & \xx       & heuristic            & query            & \xx       & \xx       & \yy       & \yy \\ %  \ee &
      % (2028.06) & Multi-modal CFs \citep{hendricks2018generating}       & \qq & \qq & \qq & \qq   & \qq     & \qq   & \qq  & \qq   & \qq    & \qq       & \qq       & \qq       & \yy       & \yy       & \qq                  & \qq              & \qq       & \qq       & \qq       & \qq \\ %  \ee &
      (2018.09) & Actionable Recourse \citep{ustun2019actionable}         & \xx & \xx & \yy & \xx   & \yy     & \yy   & \yy  & \xx   & \xx    & \xx       & \xx       & \pp       & \xx       & \xx       & ILP                  & white-box        & \yy       & \xx       & \xx       & \yy \\ %  \ee &
      (2018.11) & Weighted CFs \citep{grath2018interpretable}             & \yy & \yy & \yy & \ex   & \xx     & \xx   & \xx  & \xx   & \xx    & \xx       & \yy       & \pp       & \xx       & \xx       & heuristic            & query            & \xx       & \xx       & \xx       & \xx \\ %  \ee &
      (2019.01) & Efficient Search \citep{russell2019efficient}           & \xx & \xx & \yy & \xx   & \yy     & \xx   & \yy  & \xx   & \xx    & \yy       & \yy       & \yy       & \xx       & \xx       & MILP                 & white-box        & \xx       & \xx       & \xx       & \yy \\ %  \ee &
      (2019.04) & CF Visual Expl. \citep{goyal2019counterfactual}         & \xx & \xx & \yy & \xx   & \xx     & \xx   & \xx  & \xx   & \xx    & \xx       & \xx       & \xx       & \yy       & \xx       & greedy search        & white-box        & \yy       & \xx       & \yy       & \xx \\ %  \ee &
      (2019.05) & MACE \citep{karimi2020model}                            & \yy & \ex & \yy & \ex   & \yy     & \yy   & \yy  & \xx   & \xx    & \yy       & \yy       & \yy       & \xx       & \xx       & SAT                  & white-box        & \yy       & \yy       & \yy       & \yy \\ %  \ee &
      (2019.05) & DiCE \citep{mothilal2019explaining}                     & \xx & \xx & \yy & \xx   & \yy     & \xx   & \ex  & \xx   & \xx    & \yy       & \xx       & \yy       & \xx       & \xx       & grad opt             & gradient         & \xx       & \xx       & \xx       & \yy \\ %  \ee &
      (2019.05) & CERTIFAI \citep{sharma2019certifai}                     & \yy & \yy & \yy & \ex   & \yy     & \xx   & \yy  & \xx   & \xx    & \yy       & \xx       & \yy       & \yy       & \xx       & gen alg              & query            & \xx       & \xx       & \xx       & \xx \\ %  \ee &
      (2019.06) & MACEM \citep{dhurandhar2019model}                       & \yy & \ex & \ex & \ex   & \xx     & \xx   & \yy  & \yy   & \xx    & \xx       & \yy       & \yy       & \xx       & \xx       & FISTA                & query            & \xx       & \xx       & \xx       & \xx \\ %  \ee &
      (2019.06) & Expl. using SHAP \citep{rathi2019generating}            & \yy & \yy & \yy & \yy   & \xx     & \xx   & \xx  & \xx   & \xx    & \xx       & \xx       & \pp       & \xx       & \xx       & heuristic            & query            & \xx       & \xx       & \xx       & \yy \\ %  \ee &
      (2019.07) & Nearest Observable \citep{wexler2019if}                 & \ex & \ex & \yy & \ex   & \ex     & \ex   & \yy  & \xx   & \yy    & \yy       & \xx       & \yy       & \yy       & \xx       & brute force          & dataset          & \xx       & \xx       & \xx       & \yy \\ %  \ee &
      (2019.07) & Guided Prototypes \citep{van2019interpretable}          & \ex & \ex & \yy & \ex   & \ex     & \xx   & \yy  & \yy   & \yy    & \xx       & \yy       & \yy       & \yy       & \xx       & grad opt/FISTA       & gradient/query   & \xx       & \xx       & \yy       & \yy \\ %  \ee &
      (2019.07) & REVISE \citep{joshi2019towards}                         & \xx & \xx & \yy & \xx   & \yy     & \xx   & \yy  & \yy   & \xx    & \xx       & \xx       & \yy       & \yy       & \xx       & grad opt             & gradient         & \xx       & \xx       & \xx       & \xx \\ %  \ee &
      (2019.08) & CLEAR \citep{white2019measurable}                       & \ex & \yy & \yy & \ex   & \xx     & \xx   & \yy  & \yy   & \xx    & \xx       & \xx       & \yy       & \xx       & \xx       & heuristic            & query            & \xx       & \xx       & \xx       & \yy \\ %  \ee &
      (2019.08) & MC-BRP \citep{lucic2020does}                            & \ex & \ex & \yy & \ex   & \xx     & \xx   & \xx  & \xx   & \xx    & \xx       & \xx       & \pp       & \xx       & \xx       & heuristic            & query            & \xx       & \xx       & \xx       & \yy \\ %  \ee &
      (2019.09) & FACE \citep{poyiadzi2019face}                           & \ex & \ex & \yy & \ex   & \yy     & \yy   & \ex  & \yy   & \yy    & \xx       & \xx       & \yy       & \yy       & \xx       & graph + heuristic    & query            & \xx       & \yy       & \yy       & \yy \\ %  \ee &
      (2019.09) & Equalizing Recourse \citep{gupta2019equalizing}         & \ex & \yy & \ex & \ex   & \yy     & \xx   & \xx  & \xx   & \xx    & \xx       & \xx       & \yy       & \xx       & \xx       & ILP/heuristic        & white-box/query  & \xx       & \xx       & \xx       & \xx \\ %  \ee &
      (2019.10) & Action Sequences \citep{ramakrishnan2019synthesizing}   & \xx & \xx & \yy & \xx   & \yy     & \yy   & \xx  & \xx   & \xx    & \xx       & \xx       & \yy       & \yy       & \xx       & program synthesis    & class prob.      & \yy       & \yy       & \yy       & \yy \\ %  \ee &
      (2019.10) & C-CHVAE \citep{pawelczyk2019towards}                    & \ex & \yy & \yy & \ex   & \yy     & \xx   & \yy  & \yy   & \xx    & \xx       & \xx       & \yy       & \xx       & \xx       & grad opt + heuristic & query + gradient & \xx       & \yy       & \yy       & \yy \\ %  \ee &
      (2019.11) & FOCUS \citep{lucic2019actionable}                       & \yy & \xx & \xx & \yy   & \xx     & \xx   & \xx  & \xx   & \xx    & \xx       & \xx       & \pp       & \xx       & \xx       & grad opt + heuristic & white-box        & \yy       & \yy       & \xx       & \yy \\ %  \ee &
      (2019.12) & Model-based CFs \citep{mahajan2019preserving}           & \xx & \xx & \yy & \xx   & \yy     & \yy   & \xx  & \yy   & \xx    & \yy       & \xx       & \yy       & \yy       & \xx       & grad opt             & gradient         & \xx       & \xx       & \yy       & \yy \\ %  \ee &
      (2019.12) & LIME-C/SHAP-C \citep{ramon2019counterfactual}           & \ex & \yy & \yy & \ex   & \yy     & \xx   & \xx  & \xx   & \xx    & \xx       & \xx       & \pp       & \xx       & \yy       & heuristic            & query            & \yy       & \yy       & \yy       & \yy \\ %  \ee &
      (2019.12) & EMAP \citep{chapman2019emap}                            & \yy & \ex & \ex & \ex   & \xx     & \xx   & \xx  & \xx   & \xx    & \xx       & \xx       & \yy       & \yy       & \xx       & grad opt             & dataset/query    & \xx       & \xx       & \yy       & \xx \\ %  \ee &
      (2019.12) & PRINCE \citep{ghazimatin2020prince}                     & \ex & \ex & \ex & \yy   & \xx     & \xx   & \xx  & \xx   & \xx    & \xx       & \xx       & \yy       & \xx       & \xx       & graph + heuristic    & query            & \yy       & \yy       & \yy       & \yy \\ %  \ee &
      (2019.12) & LowProFool \citep{ballet2019imperceptible}              & \xx & \xx & \yy & \xx   & \xx     & \xx   & \yy  & \xx   & \xx    & \xx       & \xx       & \pp       & \xx       & \xx       & grad opt             & gradient         & \yy       & \xx       & \xx       & \xx \\ %  \ee &
      (2020.01) & ABELE \citep{guidotti2019black}                         & \yy & \xx & \yy & \xx   & \xx     & \xx   & \xx  & \yy   & \yy    & \xx       & \xx       & \xx       & \yy       & \xx       & gen alg + heuristic  & query + data     & \xx       & \xx       & \xx       & \yy \\ %  \ee &
      (2020.01) & SHAP-based CFs \citep{fernandez2020explaining}          & \ex & \ex & \yy & \ex   & \xx     & \xx   & \xx  & \xx   & \xx    & \xx       & \yy       & \yy       & \xx       & \xx       & heuristic            & query            & \xx       & \xx       & \xx       & \yy \\ %  \ee &
      (2020.02) & CEML \citeartelt                                        & \ex & \ex & \yy & \yy   & \xx     & \xx   & \xx  & \xx   & \yy    & \xx       & \yy       & \yy       & \yy       & \xx       & grad opt/heuristic   & gradient/query   & \yy       & \xx       & \yy       & \yy \\ %  \ee &
      (2020.02) & MINT \citep{karimi2020mintrecourse}                     & \yy & \ex & \yy & \ex   & \yy     & \yy   & \yy  & \yy   & \xx    & \yy       & \yy       & \yy       & \xx       & \xx       & SAT                  & white-box        & \yy       & \yy       & \xx       & \yy \\ %  \rr &
      (2020.03) & ViCE \citep{gomez2020vice}                              & \ex & \yy & \ex & \ex   & \yy     & \xx   & \xx  & \xx   & \xx    & \xx       & \xx       & \pp       & \xx       & \xx       & heuristic            & query            & \xx       & \xx       & \xx       & \yy \\ %  \ee &
      (2020.03) & Plausible CFs \citep{barredo2020plausible}              & \xx & \xx & \yy & \xx   & \xx     & \xx   & \xx  & \yy   & \xx    & \xx       & \xx       & \xx       & \yy       & \xx       & grad opt + gen alg   & dataset          & \yy       & \xx       & \xx       & \xx \\ %  \ee &
      (2020.04) & SEDC-T \citep{vermeire2020explainable}                  & \ex & \ex & \yy & \ex   & \xx     & \xx   & \xx  & \xx   & \xx    & \xx       & \xx       & \xx       & \yy       & \xx       & heuristic            & query            & \xx       & \xx       & \yy       & \yy \\ %  \ee &
      (2020.04) & MOC \citep{dandl2020multi}                              & \yy & \yy & \yy & \yy   & \yy     & \yy   & \yy  & \yy   & \xx    & \yy       & \yy       & \yy       & \xx       & \xx       & gen alg              & query            & \yy       & \xx       & \xx       & \yy \\ %  \ee &
      (2020.04) & SCOUT \citep{wang2020scout}                             & \xx & \xx & \yy & \xx   & \xx     & \xx   & \xx  & \xx   & \xx    & \xx       & \xx       & \xx       & \yy       & \xx       & grad opt             & gradient         & \xx       & \xx       & \yy       & \xx \\ %  \ee &
      (2020.04) & ASP-based CFs \citep{bertossi2020asp}                   & \ex & \ex & \ex & \ex   & \yy     & \yy   & \xx  & \xx   & \xx    & \xx       & \xx       & \yy       & \xx       & \xx       & answer-set prog.     & query            & \xx       & \xx       & \xx       & \xx \\ %  \ee &
      (2020.05) & CBR-based CFs \citep{keane2020good}                     & \ex & \ex & \yy & \ex   & \xx     & \xx   & \yy  & \yy   & \xx    & \xx       & \yy       & \yy       & \xx       & \xx       & heuristic            & query + data     & \xx       & \xx       & \xx       & \xx \\ %  \ee &
      (2020.06) & Survival Model CFs \citep{kovalev2020counterfactual}    & \yy & \ex & \ex & \yy   & \yy     & \xx   & \xx  & \xx   & \yy    & \xx       & \xx       & \pp       & \xx       & \xx       & gen alg              & query            & \yy       & \xx       & \xx       & \xx \\ %  \ee &
      (2020.06) & Probabilistic Recourse \citep{karimi2020caterecourse}   & \yy & \ex & \yy & \ex   & \yy     & \xx   & \yy  & \yy   & \xx    & \ex       & \xx       & \pp       & \xx       & \xx       & grad opt/brute force & gradient/query   & \yy       & \xx       & \xx       & \yy \\ %  \rr &
      (2020.06) & C-CHVAE \citep{pawelczyk2020counterfactual}             & \yy & \xx & \yy & \xx   & \yy     & \xx   & \xx  & \yy   & \xx    & \xx       & \xx       & \pp       & \xx       & \xx       & grad opt             & gradient         & \xx       & \xx       & \xx       & \yy \\ %  \ee &
      (2020.07) & FRACE \citep{zhao2020fast}                              & \xx & \xx & \yy & \xx   & \xx     & \xx   & \yy  & \yy   & \xx    & \xx       & \xx       & \xx       & \yy       & \xx       & grad opt             & gradient         & \xx       & \xx       & \yy       & \xx \\ %  \ee &
      (2020.07) & DACE \citep{kanamoridace}                               & \yy & \xx & \yy & \xx   & \yy     & \ex   & \yy  & \yy   & \xx    & \yy       & \xx       & \yy       & \xx       & \xx       & MILP                 & white-box        & \xx       & \xx       & \yy       & \xx \\ %  \ee &
      (2020.07) & CRUDS \citep{downscruds}                                & \ex & \ex & \yy & \ex   & \yy     & \yy   & \xx  & \yy   & \xx    & \yy       & \xx       & \pp       & \xx       & \xx       & grad opt             & gradient/data    & \xx       & \yy       & \xx       & \xx \\ %  \ee &
      (2020.07) & Gradient Boosted CFs \citep{aguilar2020cold}            & \yy & \xx & \xx & \xx   & \xx     & \xx   & \xx  & \xx   & \xx    & \yy       & \xx       & \yy       & \xx       & \xx       & heuristic            & data             & \xx       & \xx       & \xx       & \yy \\ %  \ee &
      % (2020.07) & OCSE \citep{fernandez2020random}                      & \yy & \xx & \xx & \xx   & \qq     & \qq   & \qq  & \qq   & \qq    & \qq       & \qq       & \qq       & \qq       & \qq       & \qq                  & \qq              & \qq       & \qq       & \qq       & \qq \\ %  \ee &
      (2020.08) & Gradual Construction \citep{kang2020counterfactual}     & \xx & \xx & \yy & \xx   & \ex     & \xx   & \ex  & \ex   & \xx    & \xx       & \xx       & \pp       & \yy       & \yy       & heuristic            & class prob.      & \xx       & \xx       & \xx       & \xx \\ %  \ee &
      (2020.08) & DECE \citep{cheng2020dece}                              & \xx & \xx & \yy & \xx   & \yy     & \yy   & \xx  & \yy   & \xx    & \yy       & \yy       & \yy       & \xx       & \xx       & grad opt             & gradient         & \xx       & \xx       & \xx       & \yy \\ %  \ee &
      (2020.08) & Time Series CFs \citep{ates2020counterfactual}          & \ex & \xx & \ex & \xx   & \xx     & \xx   & \xx  & \xx   & \xx    & \xx       & \xx       & \pp       & \xx       & \xx       & heuristic            & query            & \xx       & \yy       & \yy       & \xx \\ %  \ee &
      (2020.08) & PermuteAttack \citep{hashemi2020permuteattack}          & \yy & \ex & \ex & \ex   & \yy     & \xx   & \xx  & \xx   & \xx    & \yy       & \xx       & \yy       & \xx       & \xx       & gen alg              & query            & \xx       & \xx       & \xx       & \xx \\ %  \ee &
      (2020.10) & Fair Causal Recourse \citep{von2020fairness}            & \yy & \ex & \yy & \ex   & \yy     & \xx   & \yy  & \yy   & \xx    & \ex       & \xx       & \pp       & \xx       & \xx       & grad opt/brute force & gradient/query   & \yy       & \xx       & \xx       & \yy \\ %  \rr &
      (2020.10) & Recourse Summaries \citep{rawal2020interpretable}       & \yy & \yy & \yy & \ex   & \yy     & \xx   & \xx  & \xx   & \xx    & \xx       & \xx       & \yy       & \xx       & \xx       & itemset mining alg   & query            & \xx       & \xx       & \xx       & \xx \\ %  \ee &
      (2020.10) & Strategic Recourse \citep{chen2020strategic}            & \xx & \xx & \yy & \xx   & \yy     & \yy   & \xx  & \xx   & \xx    & \xx       & \xx       & \yy       & \xx       & \xx       & Nelder-Mead          & query            & \yy       & \yy       & \xx       & \yy \\ %  \ee &
      (2020.11) & PARE \citep{ross2020ensuring}                           & \xx & \xx & \yy & \xx   & \xx     & \xx   & \xx  & \xx   & \xx    & \xx       & \xx       & \pp       & \xx       & \yy       & grad opt + heuristic & query            & \yy       & \xx       & \xx       & \xx \\ %  \ee &
      \bottomrule
    \end{tabular}
    % ARXIV ONLY, REMOVE FOR CSUR
    \footnotetext[1]{\ex\, E.g., a model-agnostic query-based algorithm supports all models, even if the experiments were only conducted on a subset of those presented in the table.}
    \footnotetext[2]{\pp\, E.g., an algorithm may support numeric and binary variables, but not categorical.}
  \end{minipage}
\end{table*}

% \clearpage
% \clearpage
\section{Formulation}
\label{sec:030formulation}
% In this section, we begin with the barebone structure of the optimization problem (for either contrastive explanations or consequential recommendations), and then define various objective measures and incrementally append various constraints used in the literature. The following section will then review ways to solve the formulations defined here.

\amirm{intro formulations}
% In the previous section, we learned that contrastive explanations and consequential recommendations were jointly necessary to offer recourse.
%
% In this section, we summarize the optimization problem formulations for such recourse and define various objective measures and constraints used in the literature. The following section will then review ways to solve the formulations defined here.
%
\amirm{intro contras. explan.}
Given a fixed predictive model, commonly assumed to be a binary classifier, $\htheta : \mathcal{X} \rightarrow \{0, 1\}$, with $\mathcal{X} = \mathcal{X}_1 \times \cdots \times \mathcal{X}_D$, we can define the set of \emph{contrastive explanations} for a (factual) input $\xF \in \mathcal{X}$ as \hbox{$\mathcal{E} \coloneq \{\xCF \in \mathcal{P}(\mathcal{X}) ~|~ \htheta(\xCF) \not= \htheta(\xF) \}$}.
Here, $\mathcal{P}(\mathcal{X}) \subseteq \mathcal{X}$ is a \emph{plausible} subspace of $\mathcal{X}$, according to the distribution of training data (see \S\ref{sec:plausibility}). % where inputs are likely according to
Descriptively, contrastive explanations identify alternative feature combinations (in nearby worlds \citep{lewis1973counterfactuals}) that result in a favourable prediction from the fixed model.
Assuming a notion of dissimilarity between instances, represented as $\dist(\cdot, \cdot) : \mathcal{X} \times \mathcal{X} \rightarrow \mathbb{R}_+$, one can identify \emph{nearest contrastive explanations} (a.k.a. counterfactual explanations) as follows:

\vspace{-3mm}
\begin{equation}
  \label{eq:explanation}
  \begin{aligned}
    \xCFnearest \in \argmin_{\xCF \in \mathcal{X}(\mathcal{P})} && \dist(\xAny, \xF)          \\
                                                     \text{s.t.} && \htheta(\xCF) \not= \htheta(\xF) \\
                                                                 && \xCF = \xF + \deltaVector              \\
                                                             %   && \xAny \in \mathcal{A}(\xF)        \\
                                                             %   && \xAny \in \mathcal{P}(\xF)        \\
  \end{aligned}
\end{equation}
\vspace{-3mm}

% \noindent It is common to assume that $\xAny = \xF + \delta$ is a perturbation of the factual instance.
\noindent where $\deltaVector$ is the perturbation applied independently to the feature vector $\xF$ to obtain the counterfactual instance $\xCFE$.
\amirm{need explan. + recomm.}
As discussed in \S\ref{sec:020background}, although contrastive explanations identify the \emph{feature vectors} that would achieve recourse, in general, the \emph{set of actions} that would need to be performed to realize these features are not directly implied from the explanation~\citep{karimi2020mintrecourse}. 
\amirm{intro recomm.}
Thus,
% to distinguish between \emph{explanations} and their counterparts, \emph{recommendations},
a \emph{consequential recommendation} for (factual) input $\xF \in \mathcal{X}$ is defined as \hbox{$\mathcal{R} \coloneq \{a \in \mathcal{A}(\xF) ~|~ \htheta\big(\xSCF(\actionVector; \xF)\big) \not= \htheta(\xF) \}$}.
Here, $\mathcal{A}(\xF)$ is the set of \emph{feasible} actions that can be performed by the individual seeking recourse (see \S\ref{sec:actionability}).
Approaching the recourse problem from a causal perspective within the structural causal model (SCM) framework \citep{karimi2020mintrecourse}, actions are considered as interventions of the form $\actionVector = \doop(\{\zrv{X}_i := \xFScalar_i + \thetaScalar_i \}_{i\in\mathcal{I}}) \in \mathcal{A}(\xF)$, and $\xSCF(\actionVector; \xF)$ denotes the structural counterfactual of $\xF$ had action $\actionVector$ been performed, all else being equal \citep{pearl2000causality}.
Finally, given a notion of cost of actions, capturing the effort expended by the individual as $\cost(\cdot; \cdot) : \mathcal{A} \times \mathcal{X} \rightarrow \mathbb{R}_+$, one can identify \emph{minimal consequential recommendations} as follows:

\vspace{-3mm}
\begin{equation}
  \label{eq:recommendation}
  \begin{aligned}
    \actionVectorStar \in \argmin_{\actionVector \in \mathcal{A}(\xF)} && \cost(\actionVector; \xF)  \\
                                                           \text{s.t.} && \htheta(\xCF) \not= \htheta(\xF) \\
                                                                       && \xCF = \xSCF(\actionVector; \xF) \\
  \end{aligned}
\end{equation}
\vspace{-3mm}

% \amirr{causality matters}
% \noindent where $\actionVectorStar$ is the set of interventions $\doop(\{\zrv{X}_i := \actionScalar_i\}_{i\in\mathcal{I}})$ that the individual, $\xF$, should perform to achieve recourse.
%

% Importantly, $\xCF = \xF + \deltaVectorStar$ and $\xCF = \xSCF(\actionVectorStar; \xF)$ are both \emph{counterfactual} instances (see \S\ref{sec:causal_recourse}) and constitute a contrastive explanation.
% %
% Evidently, solving for consequential recommendations (i.e., the solution of \eqref{eq:recommendation}) also yields a contrastive explanation (i.e., by construction $\xCF = \xSCF(\actionVectorStar; \xF)$).
% %
% However, this requires additional information about the causal generative process (captured in the SCM) and is thus more difficult to compute (as discussed in \S\ref{sec:causal_recourse}).\footnotemark

\amirr{(2) gives (1), but not vice-versa}
Importantly, the solution of \eqref{eq:explanation} yields a nearest contrastive explanation (i.e., $\xCF = \xF + \deltaVectorStar$), with no direct mapping to a set of (minimal) consequential recommendations \citep{karimi2020mintrecourse}.
Conversely, solving \eqref{eq:recommendation} yields both a minimal consequential recommendation (i.e., $\actionVectorStar$) and a contrastive explanation (i.e., by construction $\xCF = \xSCF(\actionVectorStar; \xF)$).\footnotemark\\
\amirr{strong stance: (2) > (1)}
Our position is that, with the aim of providing recourse, the primary goal should be to provide minimal consequential recommendations that result in a (not necessarily nearest) contrastive explanation when acted upon.
Offering nearest contrastive explanations that are not necessarily attainable through \emph{minimal} effort is of secondary importance to the individual.
In practice, however, due to the additional assumptions needed to solve \eqref{eq:recommendation} (specifically for computing $\xSCF$), the literature often resorts to solving \eqref{eq:explanation}. % i.e., information about the causal generative process (captured in the SCM)

% However, the latter requires additional information about the causal generative process and is thus more difficult to compute (as discussed in \S\ref{sec:causal_recourse}).

\footnotetext{
Relatedly, the counterfactual instance that results from performing optimal actions, $\actionVectorStar$, \emph{need not} correspond to the counterfactual instance resulting from optimally and independently shifting features according to $\deltaVectorStar$; see \citep[][prop. 4.1]{karimi2020mintrecourse} and \citep[][Fig. 1]{barocas2020hidden}.
This discrepancy may arise due to, e.g., minimal recommendations suggesting that actions be performed on an ancestor of those variables that are input to the model.
}

% they need not correspond to the same instance. \amir{TODO: give illustrative example of this... overlap in indep worlds.. add figure to page 1}. Thus, while performing a minimal consequential recommendation, $\actionVectorStar$, always results in a counterfactual instance (i.e., by construction $\xCF = \xSCF(\actionVectorStar; \xF)$), it may not correspond to the nearest contrastive explanation in general.

In the remainder of this section, we provide an overview of the objective function and constraints used in \eqref{eq:explanation} and \eqref{eq:recommendation}, followed by a description of the datatypes commonly used in recourse settings. % followed by a description of plausibility and feasibility constraints and the datatypes commonly employed in recourse settings.
Finally, we conclude with related formulations.
%
% The following section will then review the tools used to solve the formulations defined here.
Then in Section~\ref{sec:040solution}, we review the tools used to solve the formulations defined here.

% \newpage
\subsection{Optimization objective}
\label{sec:objective}

\amirm{defining obj. is hard}
Generally, it is difficult to define dissimilarity (\dist) between individuals, or \cost\, functions for effort expended by individuals.
Notably, this challenge was first discussed in the algorithmic fairness literature~\citep{dwork2012fairness, ilvento2019metric}, and later echoed throughout the algorithmic recourse community~\citep{venkatasubramanianphilosophical, barocas2020hidden}.
In fact, ``the law provides no formal guidance as to the proper metric for determining what reasons are most salient'' \citep{selbst2018intuitive}.
In spite of this, existing works have presented various ad-hoc formulations with sensible intuitive justifications or practical allowance, which we review below.
%
% We review the most commonly used examples below.

\subsubsection{\textbf{On \dist}}

\amirm{some example defns}
\citet{wachter2017counterfactual} define \dist\, as the Manhattan distance, weighted by the inverse median absolute deviation (MAD):
\vspace{-3mm}
\begin{equation}
    \begin{aligned}
        \dist(\xAny, \xF) &= \sum_{k \in [D]} \frac{|\xAny_k - \xF_k|}{\text{MAD}_k} \\
        \text{MAD}_k            &= \text{median}_{j \in P} (|X_{j,k} - \text{median}_{l \in P}(X_{l,k})|)
    \end{aligned}
\end{equation}

This distance has several desirable properties, including accounting and correcting for the different ranges across features through the MAD heuristic, robustness to outliers with the use of the median absolute difference, and finally, favoring sparse solutions through the use of $\ell_1$ Manhattan distance.
% for the natural \hl{deviation/spread} in the dataset and

\citet{karimi2020model} propose a weighted combination of $\ell_p$ norms as a flexible measure across a variety of situations.
The weights, $\alpha, \beta, \gamma, \zeta$ as shown below, allow practitioners to balance between sparsity of changes (i.e., through the $\ell_0$ norm), an elastic measure of distance (i.e., through the $\ell_1, \ell_2$ norms)~\citep{dhurandhar2018explanations}, and a maximum normalized change across all features (i.e., through the $\ell_\infty$ norm):

\vspace{-3mm}
\begin{equation}
    \begin{aligned}
        \dist(\xAny; \xF) = \alpha || \deltaVector ||_0 + \beta || \deltaVector ||_1 + \gamma || \deltaVector ||_2 + \zeta || \deltaVector ||_\infty
    \end{aligned}
\end{equation}

\noindent where $\deltaVector = [\deltaScalar_1, \cdots, \deltaScalar_D]^T$ and $\delta_k: \mathcal{X}_k \times \mathcal{X}_k \rightarrow [0,1] ~ \forall ~ k \in [D]$. This measure accounts for the variability across heterogeneous features (see \S\ref{sec:datatypes}) by independently normalizing the change in each dimension according to its spread. Additional weights may also be used to relative emphasize changes in specific variables.
Finally, other works aim to minimize dissimilarity on a graph manifold \citep{poyiadzi2019face}, in terms of Euclidean distance in a learned feature space \citep{pawelczyk2019towards, joshi2019towards}, or using a Riemannian metric in a latent space \citep{arvanitidis2017latent, arvanitidis2020geometrically}.

\subsubsection{\textbf{On \cost}}

\amirm{more example defns}
Similar to \citep{karimi2020model}, various works explore $\ell_p$ norms to measure cost of actions. \citet{ramakrishnan2019synthesizing} explore $\ell_1, \ell_2$ norm as well as constant cost if specific actions are undertaken; \citet{karimi2020mintrecourse, karimi2020caterecourse} minimize the $\ell_2$ norm between $\xF$ and the action $\actionVector$ assignment (i.e., $||\thetaVector||_2$); and \citet{cui2015optimal} explore combinations of $\ell_0, \ell_1, \ell_2$ norms over a user-specified cost matrix. Encoding individual-dependent restrictions is critical, e.g., obtaining \variable{credit} is more difficult for an foreign students compared to local resident. % \citet{karimi2020mintrecourse, karimi2020caterecourse} minimize the normalized $\ell_2$ norm

Beyond $\ell_p$ norms, the work of \citet{ustun2019actionable} propose the total- and maximum-log percentile shift measures, to automatically account for the distribution of points in the dataset, e.g.,

\vspace{-3mm}

\begin{equation}
    \begin{aligned}
        \cost(\actionVector; \xF) &= \max_{k \in [D]} |Q_k(\xF_k + \thetaScalar_k) - Q_k(\xF_k)|
    \end{aligned}
\end{equation}

\noindent where $Q_k(\cdot)$ is the CDF of $x_k$ in the target population. This type of metric naturally accounts for the relative difficulty of moving to unlikely (high or low percentile) regions of the data distribution.
For instance, going from a 50 to 55th percentile in \variable{school grades} is simpler than going from 90 to 95th percentile.

\subsubsection{\textbf{On the relation between \dist\, and \cost}}
% In an independent world, where actions are performed as interventions that correspond to a separate $\delta$ shift of each feature, \dist\, and \cost\, can be used interchangeably. 

% In a world in which changing one variable does not affect others, \dist\, and \cost\, \emph{may} be used interchangeably.
% %
% % This can be seen by noting that in an \emph{independent} world,
% In such settings, we note that $\xCF = \xSCF(\actionVector; \xF) = \xF + \thetaVector$ in \eqref{eq:recommendation} mirrors the $\xCF = \xF + \deltaVector$ relation in \eqref{eq:explanation}.
% %
% \amirr{causality matters}
% % In \emph{dependent} world settings where
% Conversely, in a world in which actions may consequently change other features, $\xCF = \xSCF(\actionVector; \xF)$ need not correspond to $\xF + \thetaVector$ and thus \dist\, and \cost\, definitions may no longer overlap \citep{karimi2020mintrecourse}.
% %
% % \hl{for instance} changing a \variable{job} may affect \variable{salary} and ...
% %
% % We explore more nuanced relations in \S\ref{sec:cost_of_recourse}.
% % \amir{TODO: relate $\deltaVector$ and $\actionVector$?}
% %

In a world in which changing one variable does not affect others, one can see a parallel between the counterfactual instance of \eqref{eq:explanation}, i.e., $\xCF = \xF + \deltaVector$, and that of \eqref{eq:recommendation}, i.e., $\xCF = \xSCF(\actionVector; \xF) = \xF + \thetaVector$.
This mirroring form suggests that definitions of dissimilarity between individuals (i.e., \dist) and effort expended by an individual (i.e., \cost) \emph{may} be used interchangeably.
Following \citep{karimi2020mintrecourse}, however, we do not consider a general \hbox{1-1} mapping between \dist\, and \cost.
For instance, in a two-variable system with \variable{medication} as the parent of \variable{headache}, an individual that consumes more than the recommended amount of medication may not recover from the headache, i.e., higher cost but smaller \emph{symptomatic} distance relative to another individual who consumed the correct amount.
Furthermore, while dissimilarity is often considered to be symmetric (i.e., $\dist(\xF_A, \xF_B) = \dist(\xF_B, \xF_A)$), the effort needed to go from one profile to another need not satisfy symmetry, e.g., spending money is easier than saving money (i.e., $\cost(\actionVector = \doop(\zrv{X}_{\$} := \xF_{A,\$} - \$500); \xF_A) \le \cost(\actionVector = \doop(\zrv{X}_{\$} := \xF_{B,\$} + \$500); \xF_B)$.
These example illustrate that the interdisciplinary community must continue to engage to define the distinct notions of \dist\, and \cost, and such definitions cannot arise from a technical perspective alone.

% \newpage
\subsection{Model and counterfactual constraints}
\label{sec:model_and_counterfactual}
\subsubsection{\textbf{Model}}
\amirm{types of models}
A variety of fixed models have been explored in the literature for which recourse is to be generated. As summarized in Table \ref{table:technical_summary}, we broadly divide them in four categories:
i) tree-based (TB);
ii) kernel-based (KB);
iii) differentiable (DF); and
iv) other (OT) types (e.g., generalized linear models, Naive Bayes, k-Nearest Neighbors). % , Quadratic Discriminant (QDA) and Vector Quantization (LVQ) models
%
% \amirm{model constraint}
\amirm{multi-class \& regress.}
While the literature on recourse has primarily focused on binary classification settings, most formulations can easily be extended to multi-class classification or regression settings.
%
% \hl{give examples}
%
Extensions to such settings are straightforward, where the model constraint is replaced with $\htheta(\xCF) = k$ for a target class, or $\htheta(\xCF) \in [a,b]$ for a desired regression interval, respectively.
Alternatively, soft predictions may be used in place of hard predictions, where the goal may be, e.g., to increase the prediction gap between the highest-predicted and second-highest-predicted class, i.e.,
$\text{Pred}(\xCF)_i - \text{Pred}(\xCF)_j$ where $i = \argmax_{k \in K} \text{Pred}(\xCF)_k$, $j = \argmax_{k \in K \setminus i} \text{Pred}(\xCF)_k$.
\amirm{arbitrarily complex}
In the end, the model constraint representing the change in prediction may be arbitrarily non-linear, non-differentiable, and non-monotone \citep{barocas2020hidden}, which may limit the applicability of solutions (c.f. \S\ref{sec:040solution}).

% \footnotetext{This is similar to targeted mis-classification attacks; we discuss this parallel in \S\ref{sec:related}.}

\subsubsection{\textbf{Counterfactual}}
\amirr{causality matters}
The counterfactual constraint depends on the type of recourse offered. Whereas $\xCF = \xF + \deltaVector$ in \eqref{eq:explanation} is a linear constraint, computing $\xCF = \xSCF(\actionVector; \xF)$ in \eqref{eq:recommendation} involves performing the three step abduction-action-prediction process of \citet{pearl2016causal} and may thus be non-parametric and arbitrary involved. A closed-form expression for deterministically computing the counterfactual in additive noise models is presented in \citep{karimi2020mintrecourse}, and probabilistic derivations for more general SCMs are presented in \citep{karimi2020caterecourse}.
%
% \amir{TODO: add?}

% While the formulations in \eqref{eq:explanation} and \eqref{eq:recommendation} are agnostic to the underlying model class, $\htheta$, we shall see in \S\ref{sec:040solution} that the choice of optimizer and properties of the ensuing solutions are heavily influenced by the model, and specifically whether the model is linear, monotonic, convex, or differentiable.
% A non-monotonic relation is particularly problematic if a data-subject is asked to increase a feature... \citep{barocas2020hidden}

% \newpage
\subsection{Actionability and plausibility constraints}

\subsubsection{\textbf{Plausibility}}
\label{sec:plausibility}
% Whereas
% \emph{actionable feasibility} constraints (i.e., $\mathcal{A}(\xF)$) restrict actions to those that are possible to do,
% \emph{plausibility} constraints (i.e., $\mathcal{P}(\mathcal{X})$) require that the resulting counterfactual instance be possibly true, believable, or realistic.
%
% When generating explanations/recommendations, it is desirable to make suggestions that correspond to likely alternative states for the individual (i.e., $\mathcal{P}(\mathcal{X})$).

\amirm{3 types of plaus.}
Existing literature has formalized plausibility constraint as one of three categories: (i) \emph{domain}-consistency; (ii) \emph{density}-consistency; and (iii) \emph{prototypical}-consistency.
Whereas domain-consistency restricts the conterfactual instance to the range of admissible values for the domain of features \citep{karimi2020model}, density-consistency focuses on likely states in the (empirical) distribution of features \citep{laugel2019issues, joshi2019towards, pawelczyk2019towards, kang2020counterfactual, dhurandhar2018explanations, dhurandhar2019model} identifying instances close to the data manifold.
A third class of plausibility constraints selects counterfactual instances that are either directly present in the dataset \citep{wexler2019if, poyiadzi2019face}, or close to a prototypical example of the dataset \citep{artelt2019computation, artelt2019efficient, van2019interpretable, kovalev2020counterfactual, laugel2019issues}.
%
% \amir{TODO: add equations from von loovern and mention what type of constraints.}

\subsubsection{\textbf{Actionability (Feasibility)}}
\label{sec:actionability}
\amirm{2 layers}
% Whereas \emph{plausibility} constraints (i.e., $\mathcal{P}(\mathcal{X})$) require that the resulting counterfactual instance be possibly true, believable, or realistic,
% \emph{actionable feasibility} constraints (i.e., $\mathcal{A}(\xF)$) restrict actions to those that are possible to do.
%
\amirr{causality matters}
The set of feasible actions, $\mathcal{A}(\xF)$, is the set of interventions $\doop(\{\zrv{X}_i := \xFScalar_i + \thetaScalar_i\}_{i\in\mathcal{I}})$ that the individual, $\xF$, is able to perform. % to achieve recourse.
To determine $\mathcal{A}(\xF)$, we must identify the set of variables upon which interventions are possible, as well as the pre-/post-conditions that the intervention must satisfy.
The actionability of each variable falls into three categories \cite{lash2017budget, karimi2020mintrecourse}:

\amirm{layer \#1}
\begin{enumerate}[label=\Roman*.]
\item actionable (and mutable), e.g., \variable{bank balance};
\item mutable but non-actionable, e.g., \variable{credit score};
\item immutable (and non-actionable), e.g., \variable{birthplace}.
\end{enumerate}
Intuitively, mutable but non-actionable variables are not directly actionable by the individual, but may change as a consequence of a change to its causal ancestors (e.g., \variable{regular debt payment}).%\footnote{Of course, this example assumes that a bank would not recommend fraudulent actions that would require direct intervention on \variable{credit-score}.}

Having identified the set of actionable variables, an intervention can change the value of a variable \emph{unconditionally} (e.g., \variable{bank balance} can increase or decrease), or \emph{conditionally} to a specific value~\citep{karimi2020mintrecourse} or in a specific direction~\citep{ustun2019actionable}.
%
% In the most general sense,
\citet{karimi2020mintrecourse} present the following examples to show that the actionable feasibility of an intervention on $\zrv{X}_i$ may be contingent on any number of conditions:
%
% \amir{TODO: change examples so they're not CC of MINT paper}

\amirm{layer \#2}
\begin{enumerate}[label=\roman*.]
\item pre-intervention value of the intervened variable (i.e., $\xFScalar_i$);
e.g., an individual's \variable{age} can only increase,
i.e., $[\xSCFScalar_\variable{age} \ge \xFScalar_\variable{age}]$;
\item pre-intervention value of other variables (i.e., $\{ \xFScalar_j \}_{j \subset [d] \setminus i}$);
e.g., an individual cannot apply for \variable{credit} on a temporary \variable{visa},
i.e.,  $[\xFScalar_\variable{visa} = \texttt{PERMANENT}] \ge [\xSCFScalar_\variable{credit} = \texttt{TRUE}]$;
\item post-intervention value of the intervened variable (i.e., $\xSCFScalar_i$);
e.g., an individual may undergo heart surgery (an additive intervention) only if they won't remiss due to sustained \variable{smoking habits},
i.e., $[\xSCFScalar_\variable{heart} \not= \texttt{REMISSION}]$
\item post-intervention value of other variables (i.e., $\{ \xSCFScalar_j \}_{j \subset [d] \setminus i}$);
e.g., an individual may undergo heart surgery only \emph{after} their blood pressure (\variable{bp}) is regularized due to medicinal intervention,
i.e., $[\xSCFScalar_\variable{bp} = \texttt{O.K.}] \ge [\xSCFScalar_\variable{heart} = \texttt{SURGERY}]$
\end{enumerate}

\amirm{simple to encode constr.}
All such feasibility conditions can easily be encoded as Boolean/logical constraint into $\mathcal{A}(\xF)$ and jointly solved for in the constrained optimization formulations \eqref{eq:explanation}, \eqref{eq:recommendation}.
An important side-note to consider is that $\mathcal{A}(\xF)$ is \emph{not} restricted by the SCM assumptions, but instead, by individual-/context-dependent consideration that determine the \emph{form}, \emph{feasibility}, and \emph{scope} of actions \citep{karimi2020mintrecourse}. 
% \hl{talk about causal structure here, and mention that feasibility of actions/interventions are not captured in the SCM itself, but are added to the optimization formulation given our understanding of the variables (~ conditional interventions)}
%
%We describe the solvers and tools in \S\ref{section:tools_access}.
%
% Beyond actions that can are directly performed by the individual seeking recourse, changes can arise as a result of actions performed by a representative/fiduciary \citep{venkatasubramanianphilosophical}, or changes in background conditions (e.g., a law change that grants multiple-entry visas to professionals in certain research fields) \citep{karimi2020mintrecourse}.

\subsubsection{\textbf{On the relation between actionability \& plausibility}}

\amirm{distinct concepts}
While seemingly overlapping, \emph{actionability} (i.e., $\mathcal{A}(\xF)$) and \emph{plausibility} (i.e., $\mathcal{P}(\mathcal{X})$) are two distinct concepts:
% that may be used in conjunction. % that need not be used exclusively
%
whereas the former restrict actions to those that are \emph{possible to do}, the latter require that the resulting counterfactual instance be \emph{possibly true, believable, or realistic}.
%
% OTHER EXAMPLE: loan... i am a 28 y.o. phd student, there may be 25 year olds (plausible) but can't suggest that i become one
Consider a Middle Eastern PhD student who is denied a U.S. visa to attend a conference.
While it is quite likely for there to be favorably treated foreign students from other countries with similar characteristics (\emph{plausible} \variable{gender}, \variable{field of study}, \variable{academic record}, etc.), % but favorable treatment, 
%
% Whereas students from other countries may be granted an entry visa, 
it is impossible for our student to act on their \variable{birthplace} for recourse (i.e., a \emph{plausible} explanation but an \emph{infeasible} recommendation).
Conversely, an individual may perform a set of \emph{feasible} actions that would put them in an \emph{implausible} state (e.g., small $p(\xCF)$; not in dataset) where the model fails to classify with high confidence. % As a result, the model may not be able to classify this \emph{implausible} state with high confidence. %
% Here, the failure of recourse is due to the system and not the inability of the individual, and as such, should be communicated to the individual. 
%
Thus, actionability and plausibility constraints may be \hbox{used in conjunction depending on the recourse setting they describe.}

% In an world in which features are independently manipulable
% \hl{Just as \dist\, and \cost\, overlapped in an independent world, so do \they overlap in an independent world}

% \clearpage% \newpage
\subsection{Diversity and sparsity constraints}

\subsubsection{\textbf{Diversity}} % \divrsity}}
\label{sec:diverse}

\amirm{diversity as a work-around}
% As discussed in \S\ref{sec:objective}, \dist\, and \cost\, functions depend on an accurate understanding of individual-/context-dependent factors, which is often difficult to capture in practice.
%
% One work-around is to generate multiple recourse explanations/recommendations, allowing the individual to choose between the offered options.
%
% This requirement can also be encoded as an additional constraint in \eqref{eq:explanation} and \eqref{eq:recommendation}, whereby successive runs of the optimizer would identify diverse recourse.
%
Diverse recourse is often sought in the presence of uncertainty, e.g., unknown user preferences when defining \dist\, and \cost.
\amirm{2 approaches}
Approaches seeking to generate diverse recourse generally fall in two categories:
i) diversity through multiple runs of the same formulation; or
ii) diversity via explicitly appending diversity constraints to prior formulations.
% i) diversity via returning explanations/recommendations with similar \dist/\cost; or
% ii) diversity via using multiple objective functions.

\amirm{appr. \#1}
In the first camp, \citet{wachter2017counterfactual} show that different runs of their gradient-based optimizer over a non-convex model (e.g., multilayer perceptron) results in different solutions as a result of different random seeds. \citet{sharma2019certifai} show that multiple evolved instances of the genetic-based optimization approach can be used as diverse explanations, hence benefiting from not requiring multiple re-runs of the optimizer.
\citet{downscruds, mahajan2019preserving, pawelczyk2019towards} generate diverse counterfactuals by passing multiple samples from a latent space that is shared between factual and counterfactual instances through a decoder, and filtering those instances that correctly flip the prediction.

\amirm{appr. \#2}
In the second camp, \citet{russell2019efficient} pursue a strategy whereby subsequent runs of the optimizer would prevent changing features in the same manner as prior explanations/recommendations.
\citet{karimi2020model} continue in this direction and suggest that subsequent recourse should not fall within an $\ell_p$-ball surrounding any of the earlier explanations/recommendations.
\citet{mothilal2019explaining, cheng2020dece} present a differentiable constraint that maximizes diversity among generated explanations by maximizing the determinant of a (kernel) matrix of the generated counterfactuals. % \hl{according to a predefined metric}.
% \hl{...} \amir{TODO: add equations from mothilal? constraint type?}

\subsubsection{\textbf{Sparsity}} %/Interpretability
\label{sec:sparse}

\amirm{sparsity}
\amirr{causality matters}
It is often argued that sparser solutions are desirable as they emphasize fewer changes (in explanations) or fewer variables to act upon (in recommendations) and are thus more interpretable for the individual \citep{miller1956magical}.
While this is not generally accepted \citep{van2019interpretable, pawelczyk2020counterfactual}, one can formulate this requirement as an additional constraint, whereby, e.g., $||\deltaVector||_0 \le s$, or $||\thetaVector ||_0 \le s$. 
Formulating sparsity as an additional (hard) constraint, rather than optimizing for it in the objective, grants the flexibility to optimize for a different object while ensuring that a solution would be sparse.

% \newpage
\subsection{Datatypes and encoding}
\label{sec:datatypes}

\amirm{tabular data}
A common theme in consequential decision-making settings is the use of datatypes that refer to real-world attributes of individuals.
As a result, datasets are often tabular \tabldata, comprising of heterogeneous features, with a mix of numeric (real, integer), binary, categorical, or ordinal variables.
Most commonly, the Adult \citep{adult_dataset}, Australian Credit \citep{dua2019uci}, German Credit \citep{bache2013uci}, GiveMeCredit \citep{yeh2009comparisons}, COMPAS \citep{propublica_compas}, HELOC \citep{holter2017fico}, etc. are used, which are highly heterogeneous.
% Fannie Mae Single Family Loan Performance \citep{mae2014fannie}

% \amir{TODO: see also page 6 of commented link}
% https://www.researchgate.net/publication/337158615_Imperceptible_Adversarial_Attacks_on_Tabular_Data/link/5dce6b704585156b3513f009/download

Different feature types obey different statistical properties, e.g., the integer-based \variable{heart rate}, real-valued \variable{BMI}, categorical \variable{blood type}, and ordinal \variable{age group} differ drastically in their range. % of permissible values.
Thus, heterogeneous data requires special handling in order to preserve their semantics.
A common approach is to encode each variable according to a predetermined strategy, which preprocesses the data before model training and consequently during recourse generation.
For instance, categorical and ordinal features may be encoded using one-hot encoding and thermometer encoding, respectively.
\amirm{tabular constr.}
To preserve the semantics of each variable during recourse generation, we must also ensure that the generated explanations/recommendations result in counterfactual instances that also satisfy the encoding constraints.
For instance, Boolean and linear constraints of the form $\sum_j \xAny_{i,j} = 1 ~ \forall ~ \xAny_{i,j} \in \{0,1\}$ are used to ensure that multiple categories are be simultaneously active, and thermometer-encoded ordinal variables are required to satisfy $\xAny_{i,j} \ge \xAny_{i,j+1} ~ \forall ~ \xAny_{i,j} \in \{0,1\}$.
% For instance, ensuring that multiple categories cannot be simultaneously active can be encoded as a combination of Boolean and linear constraints: $\sum_j \xAny_{i,j} = 1 ~ \forall ~ \xAny_{i,j} \in \{0,1\}$, as can the constraint for ordinal variables: $\xAny_{i,j} \ge \xAny_{i,j+1} ~ \forall ~ \xAny_{i,j} \in \{0,1\}$.
%
For a detailed overview, we refer to the work of \citet{nazabal2020handling}.

% Relatedly, given the difference in feature ranges between, e.g., real-valued \variable{bank-balance}, categorical \variable{race}, and ordinal \variable{education-level}, the dataset may be pre-processed to normalize features across dimensions \hl{cite}. Alternatively, a proper choice of dist/cost objective may directly account for and correct such disparities \hl{cite}, as we discussed in \S~\ref{sec:objective}.
% %

\amirm{image \& text data}
In addition to \tabldata\, tabular data, one may require contrastive explanations for \imagdata\, image-based or \textdata\, text-based datasets, as summarized in Table \ref{table:technical_summary}. 
For image-based datasets, the algorithm may optionally operate on the raw data, or on super-pixel or other forms of extracted features, e.g., a hidden representation in a neural network.
Text-based datasets are also commonly encoded as vectors representing GloVe \citep{pennington2014glove} or bag-of-words embeddings \citep{ribeiro2016should}.

\vspace{-1mm}
% \subsection{Relation to adversarial perturbations, inverse classification, counterexample generation, etc.x}
% \label{sec:adv}

% \newpage
\subsection{Related formulations}
\label{sec:related}

The problem formulation for recourse generation, and specifically that of contrastive explanations, \eqref{eq:explanation}, is broadly related to several other problems in data mining and machine learning. % both in the case of contrastive explanations and for consequential recommendations,
For instance, \emph{cost-minimizing inverse classification problem} \citep{laugel2017inverse, aggarwal2010inverse, lash2018prophit, lash2017budget, lash2017generalized, mannino2000cost}, aim to identify the ``minimum required change to a case in order to reclassify it as a member of a different preferred class?'' \citep{mannino2000cost}.
\emph{Actionable knowledge extraction} is employed in data mining to suggest ``behaviors which render a state of an instance into a preferred state'' \citep{du2011efficient} according to a classifier \citep{cao2006domain, cao2007knowledge, cao2009flexible, karim2013decision, yang2006extracting}.
Finally, \emph{adversarial perturbations} are small imperceptible changes to the input of a classifier that would alter the output prediction to a false and highly confident region \citep{papernot2017practical, moosavi2017universal, carlini2017towards, moosavi2016deepfool, papernot2016limitations, nguyen2015deep, goodfellow2014explaining, szegedy2013intriguing}.
An additional parallel shared by the above methods is in their assumption of a fixed underlying model. Extensions of the above, in which model designers anticipate and aim to prevent mailicious behavior, exist in the \emph{strategic classification} \citep{hardt2016strategic, dong2018strategic, milli2019social, miller2020strategic, kleinberg2020classifiers, liu2020disparate, hu2019disparate} and \emph{adversarial robustness} \citep{cohen2019certified, carmon2019unlabeled, fawzi2015fundamental, xie2019feature} literature.

% \citep{cohen2019certified, carmon2019unlabeled, fawzi2015fundamental, xie2019feature, carlini2019evaluating}

Whereas there exists strong parallels in their formulations, the differences arise in their intended use-cases and guarantees for the stakeholders involved. % and guarantees for various stakeholders.
For example, as opposed to recourse which aims to build trust with affected individuals, the primary use-case cited in the actionable knowledge extraction literature is to deliver cost-effective actions to maximize profit or other business objectives.
%
% Another manner in which contrastive explanations and adversarial perturbations differ is in their intended use-case.
%
Furthermore, whereas a contrastive explanation aims to inform an individual about ways in which their situation would have led to a desirable outcome, an adversarial perturbation aims to fool the human by being imperceptable (e.g., by leaving the data distribution).
In a sense, imperceptability is the anti-thesis of explainability and trust.
%
% \hl{Additionally, as we have seen in this section, the domain of consequential decision-making goes beyond those considered in other literature, mandating that recourse be offered for various deployed models and objective functions, satisfying plausibility of counterfactuals and feasibility of actions, and over heterogeneous data.}
%
% Additionally, the domain of consequential decision-making typically emphasizes human-centric constraints (e.g., feasibility of actions) and operates over heterogenous data domains, which are milder requirements in related fields. % \hl{anthropocentric}
%
Finally, building on the presentation in \S\ref{sec:020background}, offering consequential recommendations relies on a causal modelling of the world, which is largely ignored by other approaches. % (see \citep{yang2019causal} for recent efforts).

% \clearpage
\section{Solution}
\label{sec:040solution}
\amirm{hard general formulation}
% By definition, recourse is offered when an individual is presented with (nearest) contrastive explanations and (minimal) consequential recommendations, as solutions to the optimization problems in \eqref{eq:explanation} and \eqref{eq:recommendation}, respectively.
By definition, recourse is offered when an individual is presented with contrastive explanations and consequential recommendations, which can be obtained by solving \eqref{eq:explanation} and \eqref{eq:recommendation}, respectively.
Notably, the objective that is to be minimized (i.e., \dist\, or \cost) may be non-linear, non-convex, or non-differentiable.
Furthermore, without restricting the classifier family, the model constraint also need not be linear, monotonic, or convex.
Finally, based on individual-/context-specific restrictions, the problem setting may require optimizing over a constrained set of plausible instances, $\mathcal{P}(\mathcal{X})$, or feasible actions, $\mathcal{A}(\xF)$.\footnote{Optimization terminology refers to both of these constraint sets as \emph{feasibility} sets.}
Thus, a distance/cost-agnostic and model-agnostic solution with support for plausibility, feasibility, sparsity, and diversity constraints over heterogeneous datasets will in general require complex approaches, trading-off various desirable properties in the process.
Below, we discuss the importance of these properties, and provide an overview of utilized solutions. % we define and discuss % optimization approaches. 

% and describe the assumptions and tools of each approach, as well as their domain of applicability. 
%
% Below, we describe these tradeoffs in detail, and present an overview of solutions ranging from model-specific approaches with simplifying assumptions to more general approaches.

% \hl{see page 5 of} \url{https://arxiv.org/pdf/1911.07749.pdf}
% \hl{see ares paper} + \citep{khuller1999budgeted}
% \hl{see} \citep{cui2015optimal}

\subsection{Properties}
\label{sec:solutions_properties}

\amirm{trade-offs}
% Before presenting the optimization approaches,
We remark that the optimizer and the resulting solutions should ideally satisfy some desirable properties, as detailed below.
In practice, methods typically trade-off optimal guarantee \distance, perfect coverage \coverage, or efficient runtime \realtime, and may otherwise require prohibitive \emph{access} to the underlying data or predictive model.

% \subsubsection{\textbf{Optimal Distance/Cost \hl{\distance}}}
\subsubsection{\textbf{Optimality}} % \hl{\distance}}}
\label{sec:optimal}
\amirr{causality matters}
Identified counterfactual instances should ideally be \emph{proximal} to the factual instance, corresponding to a small change to the individual's situation.
When optimizing for minimal \dist\, and \cost\, in \eqref{eq:explanation} and \eqref{eq:recommendation}, the objective functions and constraints determine the \emph{existence} and \emph{multiplicity} of recourse.
For factual instance $\xF$, there may exist zero, one, or multiple\footnotemark~optimal solutions and an ideal optimizer should thus identify (at least) one solution (explanation or recommendation, respectively) if one existed, or terminate and return \texttt{N/A} otherwise.

\footnotetext{The existence of multiple equally costly recourse actions is commonly referred to as the Rashoman effect \citep{breiman2001statistical}.}
% In fact, it is often desirable for a recourse method to provide multiple recourse options to the individual (as we saw in \S\ref{sec:diverse}).}

% Consider, for example, a health model the relies heavily on \variable{exercise} to better the conditions of its users. Upon being asked about their commitments, a user may indicate an inability to devote time to regular workout sessions (i.e., an actionability constraint). Thus, the model should either return the least costly suggestion under this constraint, or return \texttt{N/A} if no solution (or an unreasonably costly \citep{venkatasubramanianphilosophical} recourse) exists according to the setup.

% In contrast, consider a bank that grants loans based on \variable{salary} and \variable{bank balance}. Whereas one solution offered to a rejected applicant would be to seek to increase their \variable{salary} (perhaps through working over-time, or pursuing a raise), another equally costly suggestion may involve paying off debt before reapplying. Ideally, the user is presented with multiple options, up to a tolerance on \dist/\cost\, optimality.\footnote{See also discussion on diversity in \S\ref{sec:diverse}.} 

\subsubsection{\textbf{Perfect coverage}} % \hl{\coverage}}}
\label{sec:coverage}
Coverage is defined as the number of individuals for which the algorithm can identify a plausible counterfactual instance (through either recourse type), if at least one solution existed \citep{karimi2020model}.
% (either through contrastive explanations or consequential recommendations)
%
Communicating the domain of applicability to users is critical for building trust \citep{wachter2017counterfactual, mittelstadt2019explaining}.

\subsubsection{\textbf{Efficient runtime}} % \hl{\realtime}}}
\label{sec:runtime}
Because explanations/recommendations are likely to be offered in conversational settings \citep{sokol2018conversational, hilton1990conversational, miller2019explanation, huk2018multimodal, weller2017challenges}, it is desirable to generate recourse in near-real-time. Thus, algorithms with efficient and interactive run-time are preferred.

\subsubsection{\textbf{Access}}
\label{sec:access}
Different optimization approaches may require various levels of access to the underlying dataset or model. Access to the model may involve \emph{query access} (where only the label is returned), \emph{gradient access} (where the gradient of the output with respect to the input is requested), or \emph{class probabilities access} (from which one can infer the confidence of the prediction), or complete \emph{white-box access} (where all the model params are known).

Naturally, there are practical implications to how much access is permissible in each setting, which further restricts the choice of tools. % \hl{optimization}
Consider an organization that seeks to generate recourse for their clients. Unless these algorithms are ran in-house by said organization, it is unlikely that the organization would hand over training data, model parameters, or even a non-rate-limited API of their models to a third-party to generate recourse.%\footnote{We further elaborate on the concern of recourse and model theft in \S~\ref{sec:recourse_security}.}

\vspace{-3mm}

\subsection{Tools}
\label{sec:tools}
\amirm{optimiz. is out of scope}

We consider the richly explored field of optimization \citep{nocedal2006numerical, boyd2004convex, sra2012optimization} out of scope of this work and suffice to briefly review the tools used specifically for recourse generation, highlighting their domain of applicability, and relegating technical details to appropriate references.
\amirm{really hard; NP-hard}
Not only is solving \eqref{eq:explanation} and \eqref{eq:recommendation} difficult in general settings \citep{lash2017budget}, it has even been shown to be NP-complete or NP-hard in restricted settings, e.g., solving for integer-based variables \citep{artelt2019computation}, solving for additive tree models \citep{cui2015optimal, tolomei2017interpretable, ates2020counterfactual} or neural networks \citep{katz2017reluplex}, and solving for quadratic objectives and constraints \citep{artelt2019computation, park2017general, boyd2004convex}.
%
% Thus, except for exhaustive search over a potentially uncountable set to identify \emph{exact} solutions, most works pursue \emph{approximate} solutions in restricted settings.
Thus, except for exhaustive search over a potentially uncountable set of solutions, most works pursue \emph{approximate} solutions in restricted settings, trading-off the desirable properties above (see Table \ref{table:technical_summary}).
Solutions can be broadly categorized as
% \emph{optimization-based},
\emph{gradient-based-optimization},
\emph{model-based},
\emph{search-based},
\emph{verification-based}, and
\emph{heuristics-based}.

% \amirm{diff \& convex}
% Under differentiability and convexity of the objective and constraints, fast proximal gradient methods such as FISTA \citep{beck2009fast} as done in \citep{dhurandhar2018explanations, dhurandhar2019model, van2019interpretable}.
% %
% \amirm{diff \& nonconv}
% For differentiable and non-convex target functions (\hl{lagrangian}), first-order methods such as projected gradient-descent or (L-)BFGS may be used to identify local optima in reasonable time.
% %
% % For linear objective and constraints (e.g., logistic regression) with convex feasibility and plausibiltiy constraints, the Simplex algorithm is used \hl{cite}
% %

% %
% % (i.e., differentiable Lagrangian of objective and constraints)

% Under differentiability and convexity of the objective and constraints, \emph{gradient-optimization-based} solutions such as FISTA \citep{beck2009fast} are employed \citep{dhurandhar2018explanations, dhurandhar2019model, van2019interpretable} to find globally optimal solutions.
% %
% \amirm{diff \& nonconv}
% For differentiable and non-convex target functions, first-order methods such as projected gradient-descent or (L-)BFGS may be used to identify local optima in reasonable time. % and convexity

Under differentiability of the objective and constraints, \emph{gradient-optimization-based} solutions such as FISTA \citep{beck2009fast} are employed \citep{dhurandhar2018explanations, dhurandhar2019model, van2019interpretable} to find globally optimal solutions under convex Lagrangian, and first-order methods such as (L-)BFGS or projected gradient-descent may be used to identify local optima otherwise. % and convexity
Relatedly, rather than solving recourse for each individual independently, some works pursue a \emph{model-based} approach, whereby a mapping from factual to counterfactual instances is learned through gradient optimization \cite{pawelczyk2019towards, mahajan2019preserving}.
These methods enjoy efficient runtimes at the cost of coverage loss and poor handling of heterogeneous data.%  and non-optimal cost.

\amirm{non-diff}
For non-differentiable settings, branch-and-bound-based \citep{lawler1966branch} approaches split the \emph{search} domain into smaller regions within which a solution may be easier to find. % smaller (perhaps convex) regions
\amirm{MILP}
Under linearity of the objectives and constraints, integer linear programming (ILP) algorithms may be used when datatypes are discrete \citep{cui2015optimal, ustun2019actionable}, and mixed-integer linear programming (MILP) extensions are utilized when some variables are not discrete \citep{russell2019efficient, kanamoridace}. (M)ILP formualations are solved using powerful off-the-shelf solvers such as CPLEX \citep{cplex2009v12} and Gurobi \citep{optimization2014inc}.
\amirm{SAT}
% Just as most gradient-based methods are iterative, so can
One may also use a combination of iterative binary \emph{search} and \emph{verification} tools to obtain solutions to \eqref{eq:explanation} and \eqref{eq:recommendation}. Here, the problem is reformulated as a constrained satisfaction problem, where the constraint corresponding to the objective (\dist\, or \cost) is updated in each iteration to reflect the bounds in which a solution is obtainable \citep{karimi2020model, karimi2020mintrecourse, mohammadi2020scalable}. As with (M)ILP, this approach benefits from the existence of off-the-shelf solvers such as Z3 \citep{de2008z3}, CVC4 \citep{barrett2011cvc4}, and pySMT \citep{gario2015pysmt}.
%, rendering this approach model-agnostic, datatype-agnostic,
%
\amirm{prog. synth.}
The problem may also be cast and solved as program synthesis \citep{ramakrishnan2019synthesizing} or answer-set programming \citep{bertossi2020asp}.
% \amir{TOOD: add program synthesis?}
The methods above typically offer optimality and perfect coverage while relying on white-box access to the fixed model parameters.

A number of \emph{heuristics-based} approaches are also explored, e.g, 
finding the shortest path (Dijkstra's algorithm \citep{cormen2009introduction}) between $\xF$ and potential $\xCF$s on an empirical graph where edges are placed between similar instances (according to, e.g., Gaussian kernel) \citep{poyiadzi2019face}.
%
% \amir{TODO: briefly describe one more paper}; and
%
\amirm{heuristics}
Finally, genetic-based approaches \citep{whitley1994genetic, zitzler1998evolutionary} find solutions over different evolutions of candidate solutions according to various heuristics \citep{guidotti2018local, sharma2019certifai, dandl2020multi, kovalev2020counterfactual, barredo2020plausible}, and benefit from being model-/datatype-/norm-agnostic via only requiring query access to the model. 
\section{Prospects}
\label{sec:050discussion}
% \input{050discussion.tex}
% \input{051promise.tex} % \newpage
% \input{052problems.tex} % \clearpage
% \subsection{Prospects}

In the previous sections we covered the definitions, formulations, and solutions of existing works aiming to offer algorithmic recourse.
We showed that generating recourse explanations and recommendations required counterfactual reasoning based on different levels of causal knowledge. % at various levels of the causal history.
Counterfactual reasoning has roots not only in the philosophy of science \citep{lipton1990contrastive, hilton1986knowledge, hilton1990conversational, lewis1973counterfactuals, lewis1986causal, woodward2005making}, but also in the psychology of human agents \citep{miller2018contrastive, miller2019explanation, byrne2019counterfactuals}, and benefits from strong technical foundations \citep{halpern2005causes, bareinboim2020pearl}.
User studies have demonstrated that causal relationships are assessed by evaluating counterfactuals \citep{mcgill1993contrastive}, and counterfactual simulation is used to predict future events \citep{gerstenberg2017eye}.
Specifically in the context of XAI, it has been shown that counterfactuals can ``make the decisions of inscrutable systems intelligible to developers and users'' \citep{byrne2019counterfactuals}, and that people perform better at predicting model behavior when presented with counterfactual instances \citep{lage2019evaluation}.
Organizations seek to deploy counterfactual-based explanations citing their easy-to-understand nature \citep{bhatt2020explainable, bhatt2020machine} and GDPR-compliance \citep{wachter2017counterfactual}.
Finally, from a practitioner's standpoint, not only does algorithmic recourse benefit from the widely exercised practice of sharing open-source implementations (see Table~\ref{table:technical_summary}), various graphical interfaces have also been developed to assist the on-boarding of non-technical stakeholders \citep{wexler2019if, cheng2020dece, gomez2020vice}.

% While already quite general,
There are, however, a number of implicit assumptions made in existing setups, e.g., that the world dynamics are known and do not change, the predictive (supervised) model is fixed, and that changes only arise due to the actions of the individual seeking recourse. % do not change (stationary) / (simultaneous) actions
%
% Specifically, the assumption that predictive model is fixed and is a supervised model, recourse is deterministic, that the \dist/\cost\, are known, that optimization-based approaches are sufficiently able to generate recourse with few trade-offs, and the absence of benchmarks and studies on how other stakeholders are affected is lacking.
%
Moreover, in the multi-agent settings considered (with e.g., bank and loan seeker), agents are assumed to act truthfully with no gaming or false reporting of features, and agents are aligned in the aim to minimize an agreed-upon objective function.
% Cost and dissimilarity can be measured 
%
% Finally, technical solutions assume, at the very least, a non-rate-limited access to the fixed model, and that progress can be measured against a set of baseline approaches.
% Progress can be measured (baselines)
% Recourse offered by model owner (non rate limited access)
%
Below, we explore settings in which these assumptions do not hold, and offer potential \hbox{solutions for extending to more realistic recourse settings}. % for overcoming shortcomings and...

% \subsection{\textbf{From deterministic to probabilistic recourse}}
\subsection{\textbf{Beyond deterministic recourse}}
In \eqref{eq:recommendation}, we saw that minimal consequential recommendations are generated subject to the constraint that the counterfactual instance, $\xCF$, is assigned to be the structural counterfactual of $\xF$ under hypothetical actions $\actionVector$, i.e., $\xCF = \xSCF(\actionVector; \xF)$ \citep{karimi2020mintrecourse}.
Computing the structural counterfactual exactly, however, relies on strong assumptions (i.e., the true SCM is an additive noise model and is known).
\citet{karimi2020caterecourse} show that without complete knowledge of the true SCM, counterfactual analysis cannot be done exactly and thus recourse cannot be offered deterministically.
Although they then present methods that offer recourse with high probability, they do so under specification of a causally sufficient graph.
\amirm{uncertainty}
Future research in this direction may explore less strict settings, perhaps accounting for hidden confounders, or partially observed graphs \citep{angrist1996identification, tian2001causal, cooper1999causal}, further adding to the uncertainy of recourse recommendations.
\amirm{other randomness}
Alternatively, sources of stochasticity may enter the recourse process via a non-deterministic decision-making system. For example, it has been demonstrated that for models trained on \emph{selective labels}, fair and optimal decisions should be made stochastically~\citep{kilbertus2020fair, bechavod2019equal, tsirtsis2020decisions}.
%
% Alternatively, consider a resource-limited social service, e.g., tax audits~\citep{tan2014can}, whereby the output of the model is a risk score for the individual/company, after which they are audited if they are randomly chosen based their score among a pool of candidates.
%
% Finally, the limited resource may be the labels upon which the decision-making system is trained, where it has been demonstrated that fair and optimal decisions are made stochastically \citep{kilbertus2020fair}. Generating recourse for such models would also entail addition considerations.
% % may only be optimal and fair under when trained in a selective labelsource
% % One may also be working with a decision-making system that 
% % Stochastic decision-making settings such as the above have been shown to be optimal under

\subsection{\textbf{Beyond supervised recourse}}
In \S\ref{sec:model_and_counterfactual} we discussed how the standard binary classification setting could be extended to support multi-class classification and regression.
\amirm{matching}
Beyond these classic supervised learning settings, an individual may be subject to an automated decision maker that determines a matching of applicants to resources across a population, e.g., kindergarten assignment for children, housing for low-income families. %, public housing for low-income families.
\amirm{RL}
Alternatively, one can expect to generate explanations in more interactive settings, such as for the actions and policies of a reinforcement learning agent \citep{madumal2020distal, van2018contrastive, madumal2019explainable, rosenfeld2020predictions} or for recommender systems \citep{ghazimatin2020prince, dean2020recommendations}. %, for a single \citep{ghazimatin2020prince} or multiple users.
\amirm{online data}
Finally, explanations may also be generated for time-series data \citep{aguilar2020cold, ates2020counterfactual, lucic2020does}, which can be extended to support online data streams and models that change over time \citep{pawelczyk2020counterfactual, barocas2020hidden, venkatasubramanianphilosophical}.

% \subsection{\textbf{Beyond fixed models}}

\subsection{\textbf{Beyond individualized recourse}}
\label{sec:beyond_individualized_recourse}
So far, the presented formulations aimed to offer recourse explanations pertaining to a single individual, and assumed that recourse recommendations would be undertaken by that individual.
However, it is natural to extend the notion of recourse beyond the data-subject in question, or beyond a single individual in the population.

\amirm{fiduciary}
An example of the former setting is when the family member of a patient decides on a treatment on their behalf when the patient cannot directly exercise their agency due to incapacitation \citep{venkatasubramanianphilosophical}. One may also consider common cases in judicial processes where a legal counsel represents and seeks recourse for their client which may then be exercised by another fiduciary. In such settings, the formulation of cost and feasibility of actions may need to be adjusted to account for restrictions on both the subject and the actor. % \hl{ex?}.

% \amirm{subpopulation}
% \hl{Another manner in which recourse can be extended beyond individualized settings is as} a result of uncertainties in the features of an individual \citep{karimi2020caterecourse}, where recourse would instead be generated for a group of individuals that share features with the individual in a certain manner.
% %
% Alternatively, we may desire to assay other properties of the system, e.g., fairness, \citep{karimi2020model, gupta2019equalizing, ustun2019actionable, cheng2020dece} \hl{add ares paper}, which we explore further below.
% %
% % The need for such analysis may arise due to uncertainty in assumptions \citep{karimi2020caterecourse} or as an intentional study of other properties of the system, e.g., the fairness \citep{karimi2020model, gupta2019equalizing, ustun2019actionable, cheng2020dece} \hl{add ares paper}.
% % %
% % As in the fairness literature, such formulations depend on more elaborate definitions of the objective and constraints in \S\ref{sec:030formulation} that may be required to hold both on an individual and on a subpopulation level.

\amirm{society}
Alternatively, recourse may be achieved through the collective action of a group of people, rather than that of a single individual \citep{karimi2020mintrecourse}.
For instance, the efforts of social and political activists may culminate in a \variable{law} change that offers better conditions for a group of individuals.
%grants multiple-entry visas to professionals in certain research fields.
%
In such settings, a (background) variable which is non-actionable (or incurs high cost) on an individual level may be rendered as actionable on a group level, which may in turn bring down the cost for all members of the group.
This example also suggests that background variables may capture contextual information (e.g., \variable{economy}) that are not characteristics of, but nonetheless affect, the individual.
Furthermore, the individual may not have control over these macro variables that change over time and violate the stationarity assumption of the world.
% , and thus may change from the time recourse is offered to the time t and may change as a result of .
%
Modelling such considerations is an open problem and relegated to future work.
% \hl{not clear whether recourse is a purely counterfactual notion.}
%
Finally, the need to analyze recourse on a sub-population level may arise due to uncertainty in assumptions \citep{karimi2020caterecourse} or as an intentional study of other properties of the system, e.g., fairness \citep{karimi2020model, gupta2019equalizing, ustun2019actionable, cheng2020dece, rawal2020interpretable}, which we explore further below.

\amirm{other desirable properties}
% \subsection{\hl{\textbf{Beyond recourse}}}
\subsection{\textbf{On the interplay of recourse and ethical ML}}
% Recourse is generally seen as a \emph{modally robust good} \citep{venkatasubramanianphilosophical} which we seek to offer to an affected data-subject. \hl{beyond the individual}
%
The research questions above have primarily focused on one stakeholder: the affected individual.
However, giving the right of recourse to individuals should not be considered in a vacuum and independently of the effect that providing explanations/recommendations may have on other \emph{stakeholders} (e.g., model deployer and regulators), or in relation to \emph{other desirable properties} (e.g., fairness, security, privacy, robustness), broadly referred to as ethical ML.
%
% Below, we explore the interplay of algorithmic recourse and other desirable system properties broadly referred to as ethical ML.
% Below, we broadly explore the interplay of algorithmic recourse and ethical ML.
We explore this interplay below.

\subsubsection{\textbf{Recourse and fairness}}
\label{sec:recourse_fairness}

Fairness in ML is a primary area of study for researchers concerned with uncovering and correcting for potentially discriminatory behavior of machine learning models.
In this regard, prior work has informally used the concept of \emph{fairness of recourse} as a means to evaluate the \emph{fairness of predictions}.
For instance, \citet{ustun2019actionable} look at comparable male/female individuals that were denied a loan and show that a disparity can be detected if the suggested recourse actions (namely, \emph{flipsets}) require relatively more effort for individuals of a particular sub-group. % (e.g., females).
Along these lines, \citet{sharma2019certifai} evaluate group fairness via aggregating and comparing the cost of recourse (namely, \emph{burden}) over individuals of different sub-populations.
\citet{karimi2020model} show that the addition of feasibility constraints (e.g., non-decreasing \variable{age}) that results in an increase in the cost of recourse indicates a reliance of the fixed model on the sensitive attribute \variable{age}, which is often considered as legally and socially unfair. %\hl{AReS?}
Here we clarify that these notions are distinct and would benefit from a proper mathematical study of the relation between them.

The examples above suggest that evidence of discriminatory recourse (e.g., reliance on \variable{race}) may be used to uncover unfair classification.
We show, however, that the contrapositive statement does not hold: 
consider, for example, a 2-D dataset comprising of two sub-groups (i.e., $s \in \{0,1\}$), where $p(x|s) = \mathcal{N}(0, 10^s)$.
Consider a binary classifier, $\htheta : \mathbb{R} \times \mathbb{S} \rightarrow \{0,1\}$, where $\htheta(x, s) = \text{sign}(x)$.
While the distribution of predictions satisfies demographic parity, the (average) recourse actions required of negatively predicted individuals in $s=1$ is larger than those in $s=0$. % $p(x|s=0) = \mathcal{N}(0,1)$ and $p(x|s=1) = \mathcal{N}(0,10)$
Thus, we observe unfair recourse even when the predictions are demographically-fair.

This contradiction (the conditional and contrapositive not holding simultaneously) can be resolved by considering a new and distinct notion of fairness, i.e., \emph{fairness of recourse}, that does not imply or is not implied by the \emph{fairness of prediction}.
\amirm{fair recourse}
In this regard, \emph{Equalizing Recourse} was recently presented by \citet{gupta2019equalizing} which offered the first recourse-based (and prediction-independent) notion of fairness. % outcome-independent notion of fairness.
The authors demonstrate that one can directly calibrate for the average distance to the decision boundary to be equalized across different subgroups during the training of both linear and nonlinear classifiers.
A natural extension would involve considering the \emph{cost of recourse actions}, as opposed to the \emph{distance to the decision boundary}, in flipping the prediction across subgroups. % by considering causal relations between variables as in \citep{karimi2020mintrecourse}.
%
% Exploring the relation between such definitions of fair recourse and fair predictions is relegated to future work.
In summary, recourse may trigger new definitions of fairness to ensure non-discriminatory behavior of ML models, as in \citep{von2020fairness}.

\subsubsection{\textbf{Recourse and robustness}}
\label{sec:recourse_robustness}
% \subsubsection{\textbf{On the robustness of recourse}}
Robustness often refers to our expectation that model outputs should not change as a result of (small) changes to the input.
In the context of recourse, we expect that similar individuals should receive similar explanations/recommendations, or that recourse suggestions for an individual should be to some extent invariant to the underlying decision-making system trained on the same dataset \citep{rudin2019stop}. % or that model predictions and recourse suggestions
In practice, however, the stability of both gradient-based \citep{alvarez2018robustness, melis2018towards, ghorbani2019interpretation, dombrowski2019explanations} and counterfactual-based \citep{laugel2019issues, laugel2019dangers} explanation systems has been called into question.
Interestingly, it has been argued that it is possible for a model to have \emph{robust predictions} but \emph{non-robust explanations} \citep{hancox2020robustness}, and vice versa \citep{laugel2019issues} (similar to relation between fair predictions and fair recourse).
Parallel studies argue that the sparsity of counterfactuals may contribute to non-robust recourse when evaluating explanations generated under different fixed models \citep{pawelczyk2020counterfactual}.
%
% Finally, in the case of consequential recommendations, the underlying causal generative process (full \citep{karimi2020mintrecourse} or partial \citep{karimi2020caterecourse}) critically determine whether or not the recommended actions will result in recourse.
Finally, in the case of consequential recommendations, robustness will be affected by assumptions of the causal generative process (see Figure~\ref{fig:contrastive_vs_consequential}).
Carefully reviewing assumptions and exploring such robustness issues in more detail is necessary to build trust in the recourse system, and in turn, in the algorithmic decision-making system.

\subsubsection{\textbf{Recourse, security, and privacy}}
\label{sec:recourse_security}
\def\centerarc[#1](#2)(#3:#4:#5)% Syntax: [draw options] (center) (initial angle:final angle:radius)
    { \draw[#1, -] ($(#2)+({#5*cos(#3)},{#5*sin(#3)})$) arc (#3:#4:#5); }
\amirm{intro to model theft}
% \hl{describe problem}
%
Model extraction concerns have been raised in various settings for machine learning APIs \citep{tramer2016stealing, lowd2005adversarial, wang2018stealing, reith2019efficiently}. % hosted by Amazon, Microsoft, and Google
In such settings, an adversary aims to obtain a surrogate model, $\hat{f}$, that is similar (e.g., in fidelity) to the target model, $f$:

\vspace{-2mm}

\begin{equation}
    \begin{aligned}
        f \approx \hat{f} = \mathrm{arg}\min_{\hat{f} \in \mathcal{F}} \mathbb{E}_{\xAny \sim P(\xAny)}\Big[\mathcal{L}\big(f(\xAny), \hat{f}(\xAny)\big)\Big].
    \end{aligned}
\end{equation}

\vspace{-2mm}

% In the API settings, it is assumed that
Here, an adversary may have access to various model outputs (e.g., classification label \citep{lowd2005adversarial}, class probabilities \citep{tramer2016stealing}, etc.) under different query budgets (unlimited, rate-limited, etc. \citep{ilyas2018black, chen2020boosting}).
%
% Model extraction may be accelerated when gradients of outputs w.r.t. inputs are available \citep{milli2019model}, as is commonly available for explanation methods based on saliency/attribution maps, especially in the image domain.
% %
% An interesting line of future research (and of practical concern \citep{ustun2019actionable, sokol2019counterfactual, selbst2018intuitive, barocas2020hidden}) is to study the ability of an adversary with access to a recourse API on their ability to extract the model, which we formalize below:
Model extraction may be accelerated in presence of additional information, such as gradients of outputs w.r.t. inputs\footnote{A large class of explanation methods rely on the gradients to offer saliency/attribution maps, especially in the image domain.} \citep{milli2019model}, or contrastive explanations \citep{aivodji2020model}.
Related to recourse, and of practical concern \citep{ustun2019actionable, sokol2019counterfactual, selbst2018intuitive, barocas2020hidden}, is a study of the ability of an adversary with access to a recourse API in extracting a model. %, which we formalize below:
Specifically, we consider a setting in which the adversary has access to a \emph{prediction API} and a \emph{recourse API} which given a factual instance, $\xF$, returns a nearest contrastive explanation, $\xCFnearest$, using a known distance function, $\Delta : \mathcal{X} \times \mathcal{X} \rightarrow \mathbb{R}_+$.\footnotemark~How many queries should be made to these API to perform a functionally equivalent extraction of $f(\cdot)$?

\footnotetext{Explanation models such as MACE~\citep{karimi2020model} provide optimal solutions, $\xCFeps$, where $f(\xF) \not= f(\xCFeps),~ \Delta(\xF, \xCFeps) \le \Delta(\xF, \xCFnearest) + \epsilon$, where $\xCFnearest$ is the optimal nearest contrastive explanation. In practice, $\epsilon=1e-5$ which in turn results in $\xCFeps \approx \xCFnearest$.}

\begin{figure}[t!]
  \begin{tikzpicture}[xscale=0.4, yscale=0.4]

    \draw[step=1cm, gray, ultra thin, opacity=0.5] (0,0) grid (8,8);
    \draw[thick] (0,0) rectangle (8,8);
    \draw[black, line width = 0.06cm, dashed, -] (0,0) .. controls (0,8) and (8,0) .. (8,8);
    \draw[black, line width = 0.06cm, dashed, -] (4,8) .. controls (0,8) and (8,0) .. (8,8);

    % olive circles
    \draw[olive, fill=olive, opacity=0.4, ultra thick] (1.9,1.75) circle (1.65cm);
    \draw[olive, fill=olive, opacity=0.4, ultra thick] (3,3) circle (0.95cm);
    \draw[olive, fill=olive, opacity=0.4, ultra thick] (5,2.5) circle (1.5cm);
    \draw[olive, fill=olive, opacity=0.4, ultra thick] (7.5,4.5) circle (0.47cm);
    \centerarc[olive, fill=olive, opacity=0.4, ultra thick](7,1)(72.5:197.5:3.3);
    \fill[olive, opacity=0.4] (8,4.155) -- (8,0) -- (3.845,0) -- cycle; % triangle under arc; needed because filling an arc leaves empty space

    % teal circles
    \draw[teal, fill=teal, opacity=0.4, ultra thick] (.5,3.5) circle (.47cm);
    \draw[teal, fill=teal, opacity=0.4, ultra thick] (2,5.3) circle (1.4cm);
    \draw[teal, fill=teal, opacity=0.4, ultra thick] (3.55,4.85) circle (0.7cm);
    \draw[teal, fill=teal, opacity=0.4, ultra thick] (5,4.35) circle (0.3cm);
    \centerarc[teal, fill=teal, opacity=0.4, ultra thick](1,7)(-122:32:1.88);
    \fill[teal, opacity=0.4] (0,5.4) -- (0,8) -- (2.6,8) -- cycle; % triangle above arc; needed because filling an arc leaves empty space

    % magenta circles
    \draw[magenta, fill=magenta, opacity=0.4, ultra thick] (4,7) circle (1cm);
    \draw[magenta, fill=magenta, opacity=0.4, ultra thick] (6,6.1) circle (1.6cm);
    \draw[magenta, fill=magenta, opacity=0.4, ultra thick] (7.2,7.2) circle (0.745cm);

  \end{tikzpicture}\quad
%   \begin{tikzpicture}[xscale=0.36, yscale=0.36, >=latex]
\begin{tikzpicture}[xscale=0.22, yscale=0.22, >=latex]

    % axis
    \draw[very thick,->] (0.3,-3.5) -- +(0,7)   node[yshift=5pt] {};
    \draw[very thick,->] (0.3,-3.5) -- +(220:4) node[yshift=-5pt,xshift=-5pt] {};
    \draw[very thick,->] (0.3,-3.5) -- +(12,0)  node[xshift=6pt] {};

    % borders of the surface2
    \path[draw, ultra thick, name path=border3, dashed, -] (-2.5,-5) to[out=20,in=220] (5,4);
    \path[draw, ultra thick, name path=border6, dashed, -] (5,4)     to[out=10,in=160] (12,4);
    \path[draw, ultra thick, name path=border4, dashed, -] (12,4)    to[out=190,in=40] (4,-7);
    \path[draw, ultra thick, name path=border5, dashed, -] (4,-7)    to[out=120,in=0] (-2.5,-5);

    % draw the surface2
    \shade[top color=gray!10, bottom color=gray!90, opacity=.30]
      (-2.5,-5)
      to[out=20,in=220] (5,4)
      to[out=10,in=160] (12,4)
      to[out=190,in=40] (4,-7)
      to[out=120,in=0]  (-2.5,-5);

    % draw factual balls
    \filldraw[ball color=magenta, opacity=0.3] (2,2)    circle (2);
    \filldraw[ball color=magenta, opacity=0.3] (3,0)    circle (1);
    \filldraw[ball color=magenta, opacity=0.3] (-1,-1)  circle (3);
    \filldraw[ball color=magenta, opacity=0.3] (2,-2.5) circle (2);
    \filldraw[ball color=magenta, opacity=0.3] (5,1)    circle (1.5);
    \filldraw[ball color=magenta, opacity=0.3] (7,4)    circle (2);
    \filldraw[ball color=magenta, opacity=0.3] (10,5)   circle (1.5);

    \filldraw[ball color=olive, opacity=0.3] (10,2)      circle (1.5);
    \filldraw[ball color=olive, opacity=0.3] (11.5,3.5)  circle (.5);
    \filldraw[ball color=olive, opacity=0.3] (9,-1)      circle (3);
    \filldraw[ball color=olive, opacity=0.3] (0,-5)      circle (1);
    \filldraw[ball color=olive, opacity=0.3] (1,-6)      circle (1.5);
    \filldraw[ball color=olive, opacity=0.3] (4,-6)      circle (2);
    \filldraw[ball color=olive, opacity=0.3] (5,-4)      circle (1.25);
    \filldraw[ball color=olive, opacity=0.3] (4.6,3)     circle (.5);
    \filldraw[ball color=olive, opacity=0.3] (5.5,-1.75) circle (1.25);

  \end{tikzpicture}
%   \vspace{-3mm}
  % \caption{\small Depiction of objective in 2D, 3D. How many factual balls (centers sampled uniformly in $\mathbb{R}^d$) are needed to maximally pack all decision regions?}
  \caption{\small Here we illustrate the model stealing process in 2D and 3D using hypothetical non-linear decision boundaries. ``How many optimal contrastive explanations are needed to extract the decision regions of a classifier?'' can be formulated as ``How many factual balls are needed to maximally pack all decision regions?''}
  \label{fig:ball_packing}
  \vspace{-3mm}
\end{figure}
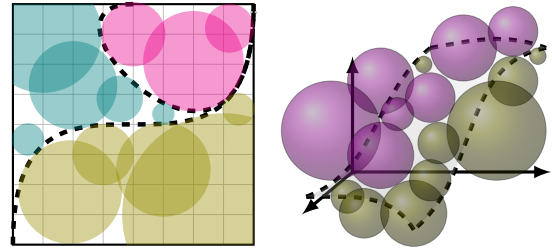
% \end{wrapfigure} 

\amirm{attack 1}
% \paragraph{Attack Strategy \#1}
In a first attack strategy, one could learn a surrogate model on a dataset where factual instances and labels (form the training set or randomly sampled) are augmented with counterfactual instances and counterfactual labels.
This idea was explored by \citep{aivodji2020model} where they demonstrated that a high fidelity surrogate model can be extracted even under low query budgets.
While easy to understand and implement, this attack strategy implicitly assumes that constructed dataset has i.i.d. data, and thus does not make use of the relations between factual and counterfactual pairs.

% \amirm{attack 2}
% % \paragraph{Attack Strategy \#2}
% An alternative attack strategy may aim to use the relation between factual and counterfactual instances.
% %
% Specifically,  consider that for linear classifiers, a single nearest counterfactual instance is sufficient to identify the normal to the separating hyperplane (i.e, by computing the vector difference between instances). 
% %
% Building on this observation, one may then proceed to compute bounds on the number of queries needed to reconstruct the Voronoi diagram ??

\amirm{attack 2}
An alternative attack strategy considers that the model $f$ can be fully represented by its decision boundaries, or the complementary \emph{decision regions} $\{\mathcal{R}_1, \cdots, \mathcal{R}_l\}$. %\footnote{Note that $l$ may be larger than $k$ if there are non-singly-connected decision regions.}
%
% Every contrastive explanation returned from the recourse API informs us that $\not\exists ~ \xAny; ~ \Delta(\xF, \xAny) < \Delta(\xF, \xCFnearest), f(\xF) \not= f(\xAny) \implies f(\xF) = f(\xAny) ~ \forall ~ \xAny \in \ballF$, i.e., all points $\xAny$ within the ball centered at $\xF$ and with radius $\Delta(\xF, \xCFnearest)$ share the same class label as $\xF$, according the the model $f$. % returned from the explanation model $g$
%
% Henceforth, we refer to $\ballF$ as the $\Delta$-\emph{factual ball}.
%
Every contrastive explanation returned from the recourse API informs us that all instance surrounding the factual instance, up to a distance of $\Delta(\xF, \xCFnearest)$, share the same class label as $\xF$ according to $f$ (otherwise that instance would be the nearest contrastive explanation).
More formally, $f(\xF) = f(\xAny) ~ \forall ~ \xAny \in \ballF$, where $\ballF$ is referred to as the $\Delta$-\emph{factual ball}, centered at $\xF$ and with radius $\Delta(\xF, \xCFnearest)$.
The model extraction problem can thus be formulated as the number of factual balls needed to maximally pack all decision regions (see Fig. \ref{fig:ball_packing}): % (with centers sampled uniformly in $\mathbb{R}^d$)

\vspace{-2mm}

\begin{equation}
    \begin{aligned}
        \Pr \large[ \mathrm{Vol}(\mathcal{R}_l) - \large\cup_{i=1, {\ballF}_i \subseteq \mathcal{R}_l}^n \mathrm{Vol}({\ballF}_i) \le \epsilon \large] \ge 1 - \delta ~~ \forall ~~ l
    \end{aligned}
\end{equation}

% \begin{align}
%   \Pr \large[ \mathrm{Vol}(\mathcal{R}_l) - \large\cup_{i=1, {\ballF}_i \subseteq \mathcal{R}_l}^n \mathrm{Vol}({\ballF}_i) \le \epsilon \large] \ge 1 - \delta ~ \forall ~ l
% \end{align}

\vspace{-2mm}

% \amirm{why defence is imp}
% As in other extraction settings, an adversary may then use the learned approximation to game the system \hl{cite} or extract sensitive information about the dataset \hl{cite}.
% %
% Understanding these vulnerabilities and designing effective defence strategies (\hl{perhaps through...}) is necessary for the adoption of recourse by stakeholders concerned with IP theft. %in relation to many stakeholders' needs.

As in other extraction settings, $\hat{f}$ can then used to infer private information on individuals in the training set, to uncover exploitable system vulnerabilities, or for free internal use.
Understanding attack strategies may guide recourse policy and the design of defensive mechanisms to hinder the exploitation of such vulnerabilities.

\amirm{privacy violation}
Surprisingly, a model need not be extracted in the sense above to be revealing of sensitive information.
%
% Building on the intuition above, we note that explanations may indirectly expose the predictions of similar individuals.
%
% In more detail,
Building on the intuition above, we note that a single contrastive explanation informs the data-subject that there are no instances in a certain vicinity (i.e., within $\ballF$) such that their prediction is different. % (otherwise that would be the nearest contrastive explanation).
This information informs the data-subject about, e.g., whether their similar friend was also denied a loan, violating their predictive privacy.
Even under partial knowledge of the friend's attributes, an adversary may use the information about the shared predictions in $\ballF$ to perform membership inference attacks \citep{shokri2019privacy} or infer missing attributes \citep{dwork2018privacy}.
This problem is worsened when multiple diverse explanations are generated, and is an open problem.

% \paragraph{Recourse and privacy}
% \label{sec:recourse_privacy}
% \input{059recourseprivacy.tex}

\subsubsection{\textbf{Recourse and manipulation}}
\label{sec:recourse_privacy}
Although a central goal of recourse is to foster trust between an individual and an automated system, it would be simplistic to assume that all parties will act truthfully in this process.
For instance, having learned something about the decision-making process (perhaps through recommendations given to similar individuals), an individual may exaggerate some of their attributes for a better chance of favorable treatment~\citep{venkatasubramanianphilosophical}. % ; see earlier discussion
Trust can also be violated by the recourse-offering party. As discussed earlier, the multiplicity of recourse explanations/recommendations (see \S\ref{sec:optimal}) may allow for an organization to cherry-pick ``the most socially acceptable explanation out of many equally plausible ones'' \citep{hancox2020robustness, lakkaraju2020fool, barocas2020hidden} (see also, \emph{fairwashing} \citep{aivodji2019fairwashing}). % , e.g., provide explanations that puts less emphasis on sensitive attributes to prevent appearing unfair % may make it seem as though an organization is providing an explanation without disclosing potential other benefit system is offering recourse whereas \hl{different cost functions weighing attributes differently.}
In such cases of misaligned incentives, the oversight of a regulatory body, perhaps with random audits of either party, seems necessary.
%
% Another potential solution would be to mandate a minimum number of diverse recourse offerings.
Another solution may involve mandating a minimum number of diverse recourse offerings, which would conflict with security considerations.

% \vspace{1mm}
% % In summary, the examples above illustrate a tension between various desirable system properties as required by the many stakeholders involved in the recourse setting.
% In summary, the examples above illustrate the diversity of stakeholder needs and a tension between the desirable system properties we seek to offer in the recourse setting.
% %
% Satisfyingly addressing these needs and navigating such trade-offs requires new definitions and techniques, and relies on the cross-disciplinary expertise of a panel of technical and social scientists.

\subsection{\textbf{Towards unifying benchmarks}}
Table \ref{table:technical_summary} presented an overview of the diverse settings in which recourse is sought.
Despite the abundance of open-source implementations built on robust tools and working well in their respective settings, a comparative benchmark for recourse is lacking.
This problem is exacerbated for consequential recommendations which further rely on assumptions about the causal generative process. % missing due to identifiability issues \hl{cite}.
In order to make objective progress, however, new contributions should be evaluated against existing methods. 
Thus, a next step for the community is the curation of an online challenge (e.g., using Kaggle) to benchmark the performance of existing and new methods.
To broadly cover practical settings, we envision multiple tracks where authors can submit their generated explanations/recommendations given a fixed classifier, test dataset, and pre-determined \dist/\cost\, definition, and be evaluated using the metrics defined in \S\ref{sec:solutions_properties}.
Authors may also submit results that satisfy additional actionability, \hbox{plausibility, and diversity constraints, and be compared as such.}

% \newpage
\section{Conclusions}
\label{sec:060conclusions}
Our work started with a case study, of a 28-year-old female professional who was denied a loan by an automated decision-making system.
We aimed to assist this individual in overcoming their difficult situation, i.e., to achieve \emph{algorithmic recourse}, which was contingent on offering answers to two questions: why, and how?
%
% In reviewing relevant literature from psychology and philosophy,
We studied the relation between these questions, and arrived at distinct responses, namely, \emph{contrastive explanations} and \emph{consequential recommendations}.
Mindful of the goal of recourse, we emphasized \emph{minimal} consequential recommendations over \emph{nearest} contrastive explanations as the former directly optimizes for the least effort from the individual.
Furthermore, we noted that offering recourse recommendations automatically implied recourse explanations (through simulation of the causal effect of undertaken actions), whereas the converse would not.
In reviewing the literature, however, we observed an under-exploration of consequential recommendations, which we attribute to its reliance on additional assumptions at the level of the \hbox{causal generative process of the world in which actions take place.}

In addition to unifying and precisely defining recourse, we present an overview of the many constraints (e.g., actionability, plausibility, diversity, sparsity) that are needed to model realistic recourse settings.
With accompanying illustrative examples, we distinguish between the notions of \dist\, vs. \cost, and plausibility vs. actionability (feasibility), whose distinctions are largely ignored in the literature.
Throughout, we reiterate that these notions are individual-/context-dependent, and that formulations cannot arise from a technical perspective alone.
We summarize the technical literature in Table \ref{table:technical_summary}, as a guide for practitioners looking for methods that satisfy certain properties, and researchers that want to identify open problems and methods to further develop.

Finally, we identify a number of prospective research directions which challenge the assumptions of existing setups, and present extensions to better situate recourse in the broader ethical ML literature.
The presented examples and discussion serve to illustrate the diversity of stakeholder needs and a tension between the desirable system properties (fairness, security, privacy, robustness) which we seek to offer alongside recourse.
Satisfyingly addressing these needs and navigating the entailed trade-offs may require new definitions and techniques, and relies on the cross-disciplinary expertise of a panel of technical and social scientists.
We hope that this document may guide further discussion and progress in this direction.

\begin{acks}
AHK sincerely thanks the senior authors for encouraging him to undertake the daunting task of writing a first draft, which eventually resulted in this manuscript. AHK is also appreciative of Julius von K\"ugelgen and Umang Bhatt for fruitful discussions on recourse and fairness, Muhammad Waleed Gondal and the anonymous reviewers for constructive feedback throughout, and NSERC and CLS for generous funding support.
% who suggested the online competition for benchmarking recourse methods and for reviewing the manuscript.
\end{acks}

% \clearpage
\small
\bibliographystyle{ACM-Reference-Format}
\bibliography{references}


\begin{thebibliography}{211}

%%% the Web address.

\ifx \showCODEN    \undefined \def \showCODEN     #1{\unskip}     \fi
\ifx \showDOI      \undefined \def \showDOI       #1{#1}\fi
\ifx \showISBNx    \undefined \def \showISBNx     #1{\unskip}     \fi
\ifx \showISBNxiii \undefined \def \showISBNxiii  #1{\unskip}     \fi
\ifx \showISSN     \undefined \def \showISSN      #1{\unskip}     \fi
\ifx \showLCCN     \undefined \def \showLCCN      #1{\unskip}     \fi
\ifx \shownote     \undefined \def \shownote      #1{#1}          \fi
\ifx \showarticletitle \undefined \def \showarticletitle #1{#1}   \fi
\ifx \showURL      \undefined \def \showURL       {\relax}        \fi
% invisible to TeX
\providecommand\bibfield[2]{#2}
\providecommand\bibinfo[2]{#2}
\providecommand\natexlab[1]{#1}
\providecommand\showeprint[2][]{arXiv:#2}

\bibitem[\protect\citeauthoryear{Abdul, Vermeulen, Wang, Lim, and
  Kankanhalli}{Abdul et~al\mbox{.}}{2018}]%
        {abdul2018trends}
\bibfield{author}{\bibinfo{person}{Ashraf Abdul}, \bibinfo{person}{Jo
  Vermeulen}, \bibinfo{person}{Danding Wang}, \bibinfo{person}{Brian~Y Lim},
  {and} \bibinfo{person}{Mohan Kankanhalli}.} \bibinfo{year}{2018}\natexlab{}.
\newblock \showarticletitle{Trends and trajectories for explainable,
  accountable and intelligible systems: An hci research agenda}. In
  \bibinfo{booktitle}{\emph{Proceedings of the 2018 CHI conference on human
  factors in computing systems}}. \bibinfo{pages}{1--18}.
\newblock


\bibitem[\protect\citeauthoryear{Adadi and Berrada}{Adadi and Berrada}{2018}]%
        {adadi2018peeking}
\bibfield{author}{\bibinfo{person}{Amina Adadi} {and} \bibinfo{person}{Mohammed
  Berrada}.} \bibinfo{year}{2018}\natexlab{}.
\newblock \showarticletitle{Peeking inside the black-box: A survey on
  Explainable Artificial Intelligence (XAI)}.
\newblock \bibinfo{journal}{\emph{IEEE Access}}  \bibinfo{volume}{6}
  (\bibinfo{year}{2018}), \bibinfo{pages}{52138--52160}.
\newblock


\bibitem[\protect\citeauthoryear{{Adult data}}{{Adult data}}{1996}]%
        {adult_dataset}
\bibfield{author}{\bibinfo{person}{{Adult data}}.}
  \bibinfo{year}{1996}\natexlab{}.
\newblock
  \bibinfo{howpublished}{\href{https://archive.ics.uci.edu/ml/datasets/adult}{https://archive.ics.uci.edu/ml/datasets/adult}}.
\newblock


\bibitem[\protect\citeauthoryear{Aggarwal, Chen, and Han}{Aggarwal
  et~al\mbox{.}}{2010}]%
        {aggarwal2010inverse}
\bibfield{author}{\bibinfo{person}{Charu~C Aggarwal}, \bibinfo{person}{Chen
  Chen}, {and} \bibinfo{person}{Jiawei Han}.} \bibinfo{year}{2010}\natexlab{}.
\newblock \showarticletitle{The inverse classification problem}.
\newblock \bibinfo{journal}{\emph{Journal of Computer Science and Technology}}
  \bibinfo{volume}{25}, \bibinfo{number}{3} (\bibinfo{year}{2010}),
  \bibinfo{pages}{458--468}.
\newblock


\bibitem[\protect\citeauthoryear{Aguilar-Palacios, Mu{\~n}oz-Romero, and
  Rojo-{\'A}lvarez}{Aguilar-Palacios et~al\mbox{.}}{2020}]%
        {aguilar2020cold}
\bibfield{author}{\bibinfo{person}{Carlos Aguilar-Palacios},
  \bibinfo{person}{Sergio Mu{\~n}oz-Romero}, {and}
  \bibinfo{person}{Jos{\'e}~Luis Rojo-{\'A}lvarez}.}
  \bibinfo{year}{2020}\natexlab{}.
\newblock \showarticletitle{Cold-Start Promotional Sales Forecasting through
  Gradient Boosted-based Contrastive Explanations}.
\newblock \bibinfo{journal}{\emph{IEEE Access}} (\bibinfo{year}{2020}).
\newblock


\bibitem[\protect\citeauthoryear{A{\"\i}vodji, Arai, Fortineau, Gambs, Hara,
  and Tapp}{A{\"\i}vodji et~al\mbox{.}}{2019}]%
        {aivodji2019fairwashing}
\bibfield{author}{\bibinfo{person}{Ulrich A{\"\i}vodji},
  \bibinfo{person}{Hiromi Arai}, \bibinfo{person}{Olivier Fortineau},
  \bibinfo{person}{S{\'e}bastien Gambs}, \bibinfo{person}{Satoshi Hara}, {and}
  \bibinfo{person}{Alain Tapp}.} \bibinfo{year}{2019}\natexlab{}.
\newblock \showarticletitle{Fairwashing: the risk of rationalization}.
\newblock \bibinfo{journal}{\emph{arXiv preprint arXiv:1901.09749}}
  (\bibinfo{year}{2019}).
\newblock


\bibitem[\protect\citeauthoryear{A{\"\i}vodji, Bolot, and Gambs}{A{\"\i}vodji
  et~al\mbox{.}}{2020}]%
        {aivodji2020model}
\bibfield{author}{\bibinfo{person}{Ulrich A{\"\i}vodji},
  \bibinfo{person}{Alexandre Bolot}, {and} \bibinfo{person}{S{\'e}bastien
  Gambs}.} \bibinfo{year}{2020}\natexlab{}.
\newblock \showarticletitle{Model extraction from counterfactual explanations}.
\newblock \bibinfo{journal}{\emph{arXiv preprint arXiv:2009.01884}}
  (\bibinfo{year}{2020}).
\newblock


\bibitem[\protect\citeauthoryear{Alvarez-Melis and Jaakkola}{Alvarez-Melis and
  Jaakkola}{2018}]%
        {alvarez2018robustness}
\bibfield{author}{\bibinfo{person}{David Alvarez-Melis} {and}
  \bibinfo{person}{Tommi~S Jaakkola}.} \bibinfo{year}{2018}\natexlab{}.
\newblock \showarticletitle{On the robustness of interpretability methods}.
\newblock \bibinfo{journal}{\emph{arXiv preprint arXiv:1806.08049}}
  (\bibinfo{year}{2018}).
\newblock


\bibitem[\protect\citeauthoryear{Angrist, Imbens, and Rubin}{Angrist
  et~al\mbox{.}}{1996}]%
        {angrist1996identification}
\bibfield{author}{\bibinfo{person}{Joshua~D Angrist}, \bibinfo{person}{Guido~W
  Imbens}, {and} \bibinfo{person}{Donald~B Rubin}.}
  \bibinfo{year}{1996}\natexlab{}.
\newblock \showarticletitle{Identification of causal effects using instrumental
  variables}.
\newblock \bibinfo{journal}{\emph{Journal of the American statistical
  Association}} \bibinfo{volume}{91}, \bibinfo{number}{434}
  (\bibinfo{year}{1996}), \bibinfo{pages}{444--455}.
\newblock


\bibitem[\protect\citeauthoryear{Angwin, Larson, Mattu, and Kirchner}{Angwin
  et~al\mbox{.}}{2016}]%
        {angwin2016machine}
\bibfield{author}{\bibinfo{person}{Julia Angwin}, \bibinfo{person}{Jeff
  Larson}, \bibinfo{person}{Surya Mattu}, {and} \bibinfo{person}{Lauren
  Kirchner}.} \bibinfo{year}{2016}\natexlab{}.
\newblock \showarticletitle{Machine bias}.
\newblock \bibinfo{journal}{\emph{ProPublica, May}}  \bibinfo{volume}{23}
  (\bibinfo{year}{2016}), \bibinfo{pages}{2016}.
\newblock


\bibitem[\protect\citeauthoryear{Artelt and Hammer}{Artelt and Hammer}{2019a}]%
        {artelt2019efficient}
\bibfield{author}{\bibinfo{person}{Andr{\'e} Artelt} {and}
  \bibinfo{person}{Barbara Hammer}.} \bibinfo{year}{2019}\natexlab{a}.
\newblock \showarticletitle{Efficient computation of counterfactual
  explanations of LVQ models}.
\newblock \bibinfo{journal}{\emph{arXiv preprint arXiv:1908.00735}}
  (\bibinfo{year}{2019}).
\newblock


\bibitem[\protect\citeauthoryear{Artelt and Hammer}{Artelt and Hammer}{2019b}]%
        {artelt2019computation}
\bibfield{author}{\bibinfo{person}{Andr{\'e} Artelt} {and}
  \bibinfo{person}{Barbara Hammer}.} \bibinfo{year}{2019}\natexlab{b}.
\newblock \showarticletitle{On the computation of counterfactual
  explanations--A survey}.
\newblock \bibinfo{journal}{\emph{arXiv preprint arXiv:1911.07749}}
  (\bibinfo{year}{2019}).
\newblock


\bibitem[\protect\citeauthoryear{Artelt and Hammer}{Artelt and Hammer}{2020}]%
        {artelt2020convex}
\bibfield{author}{\bibinfo{person}{Andr{\'e} Artelt} {and}
  \bibinfo{person}{Barbara Hammer}.} \bibinfo{year}{2020}\natexlab{}.
\newblock \showarticletitle{Convex Density Constraints for Computing Plausible
  Counterfactual Explanations}.
\newblock \bibinfo{journal}{\emph{arXiv preprint arXiv:2002.04862}}
  (\bibinfo{year}{2020}).
\newblock


\bibitem[\protect\citeauthoryear{Arvanitidis, Hansen, and Hauberg}{Arvanitidis
  et~al\mbox{.}}{2017}]%
        {arvanitidis2017latent}
\bibfield{author}{\bibinfo{person}{Georgios Arvanitidis},
  \bibinfo{person}{Lars~Kai Hansen}, {and} \bibinfo{person}{S{\o}ren Hauberg}.}
  \bibinfo{year}{2017}\natexlab{}.
\newblock \showarticletitle{Latent space oddity: on the curvature of deep
  generative models}.
\newblock \bibinfo{journal}{\emph{arXiv preprint arXiv:1710.11379}}
  (\bibinfo{year}{2017}).
\newblock


\bibitem[\protect\citeauthoryear{Arvanitidis, Hauberg, and
  Sch{\"o}lkopf}{Arvanitidis et~al\mbox{.}}{2020}]%
        {arvanitidis2020geometrically}
\bibfield{author}{\bibinfo{person}{Georgios Arvanitidis},
  \bibinfo{person}{S{\o}ren Hauberg}, {and} \bibinfo{person}{Bernhard
  Sch{\"o}lkopf}.} \bibinfo{year}{2020}\natexlab{}.
\newblock \showarticletitle{Geometrically Enriched Latent Spaces}.
\newblock \bibinfo{journal}{\emph{arXiv preprint arXiv:2008.00565}}
  (\bibinfo{year}{2020}).
\newblock


\bibitem[\protect\citeauthoryear{Ates, Aksar, Leung, and Coskun}{Ates
  et~al\mbox{.}}{2020}]%
        {ates2020counterfactual}
\bibfield{author}{\bibinfo{person}{Emre Ates}, \bibinfo{person}{Burak Aksar},
  \bibinfo{person}{Vitus~J Leung}, {and} \bibinfo{person}{Ayse~K Coskun}.}
  \bibinfo{year}{2020}\natexlab{}.
\newblock \showarticletitle{Counterfactual Explanations for Machine Learning on
  Multivariate Time Series Data}.
\newblock \bibinfo{journal}{\emph{arXiv preprint arXiv:2008.10781}}
  (\bibinfo{year}{2020}).
\newblock


\bibitem[\protect\citeauthoryear{Bache and Lichman}{Bache and Lichman}{2013}]%
        {bache2013uci}
\bibfield{author}{\bibinfo{person}{Kevin Bache} {and} \bibinfo{person}{Moshe
  Lichman}.} \bibinfo{year}{2013}\natexlab{}.
\newblock \bibinfo{title}{{UCI} machine learning repository}.
\newblock
\newblock


\bibitem[\protect\citeauthoryear{Ballet, Renard, Aigrain, Laugel, Frossard, and
  Detyniecki}{Ballet et~al\mbox{.}}{2019}]%
        {ballet2019imperceptible}
\bibfield{author}{\bibinfo{person}{Vincent Ballet}, \bibinfo{person}{Xavier
  Renard}, \bibinfo{person}{Jonathan Aigrain}, \bibinfo{person}{Thibault
  Laugel}, \bibinfo{person}{Pascal Frossard}, {and} \bibinfo{person}{Marcin
  Detyniecki}.} \bibinfo{year}{2019}\natexlab{}.
\newblock \showarticletitle{Imperceptible Adversarial Attacks on Tabular Data}.
\newblock \bibinfo{journal}{\emph{arXiv preprint arXiv:1911.03274}}
  (\bibinfo{year}{2019}).
\newblock


\bibitem[\protect\citeauthoryear{Bareinboim, Correa, Ibeling, and
  Icard}{Bareinboim et~al\mbox{.}}{2020}]%
        {bareinboim2020pearl}
\bibfield{author}{\bibinfo{person}{E Bareinboim}, \bibinfo{person}{JD Correa},
  \bibinfo{person}{D Ibeling}, {and} \bibinfo{person}{T Icard}.}
  \bibinfo{year}{2020}\natexlab{}.
\newblock \showarticletitle{On Pearl’s hierarchy and the foundations of
  causal inference}.
\newblock \bibinfo{journal}{\emph{ACM Special Volume in Honor of Judea Pearl
  (provisional title)}} (\bibinfo{year}{2020}).
\newblock


\bibitem[\protect\citeauthoryear{Barocas, Hardt, and Narayanan}{Barocas
  et~al\mbox{.}}{2017}]%
        {barocas2017fairness}
\bibfield{author}{\bibinfo{person}{Solon Barocas}, \bibinfo{person}{Moritz
  Hardt}, {and} \bibinfo{person}{Arvind Narayanan}.}
  \bibinfo{year}{2017}\natexlab{}.
\newblock \showarticletitle{Fairness in machine learning}.
\newblock \bibinfo{journal}{\emph{NIPS Tutorial}}  \bibinfo{volume}{1}
  (\bibinfo{year}{2017}).
\newblock


\bibitem[\protect\citeauthoryear{Barocas, Selbst, and Raghavan}{Barocas
  et~al\mbox{.}}{2020}]%
        {barocas2020hidden}
\bibfield{author}{\bibinfo{person}{Solon Barocas}, \bibinfo{person}{Andrew~D
  Selbst}, {and} \bibinfo{person}{Manish Raghavan}.}
  \bibinfo{year}{2020}\natexlab{}.
\newblock \showarticletitle{The hidden assumptions behind counterfactual
  explanations and principal reasons}. In \bibinfo{booktitle}{\emph{Proceedings
  of the 2020 Conference on Fairness, Accountability, and Transparency}}.
  \bibinfo{pages}{80--89}.
\newblock


\bibitem[\protect\citeauthoryear{Barredo-Arrieta and Del~Ser}{Barredo-Arrieta
  and Del~Ser}{2020}]%
        {barredo2020plausible}
\bibfield{author}{\bibinfo{person}{Alejandro Barredo-Arrieta} {and}
  \bibinfo{person}{Javier Del~Ser}.} \bibinfo{year}{2020}\natexlab{}.
\newblock \showarticletitle{Plausible Counterfactuals: Auditing Deep Learning
  Classifiers with Realistic Adversarial Examples}.
\newblock \bibinfo{journal}{\emph{arXiv preprint arXiv:2003.11323}}
  (\bibinfo{year}{2020}).
\newblock


\bibitem[\protect\citeauthoryear{Barrett, Conway, Deters, Hadarean,
  Jovanovi{\'c}, King, Reynolds, and Tinelli}{Barrett et~al\mbox{.}}{2011}]%
        {barrett2011cvc4}
\bibfield{author}{\bibinfo{person}{Clark Barrett},
  \bibinfo{person}{Christopher~L Conway}, \bibinfo{person}{Morgan Deters},
  \bibinfo{person}{Liana Hadarean}, \bibinfo{person}{Dejan Jovanovi{\'c}},
  \bibinfo{person}{Tim King}, \bibinfo{person}{Andrew Reynolds}, {and}
  \bibinfo{person}{Cesare Tinelli}.} \bibinfo{year}{2011}\natexlab{}.
\newblock \showarticletitle{Cvc4}. In \bibinfo{booktitle}{\emph{International
  Conference on Computer Aided Verification}}. Springer,
  \bibinfo{pages}{171--177}.
\newblock


\bibitem[\protect\citeauthoryear{Bastani, Kim, and Bastani}{Bastani
  et~al\mbox{.}}{2017}]%
        {bastani2017interpretability}
\bibfield{author}{\bibinfo{person}{Osbert Bastani}, \bibinfo{person}{Carolyn
  Kim}, {and} \bibinfo{person}{Hamsa Bastani}.}
  \bibinfo{year}{2017}\natexlab{}.
\newblock \showarticletitle{Interpretability via model extraction}.
\newblock \bibinfo{journal}{\emph{arXiv preprint arXiv:1706.09773}}
  (\bibinfo{year}{2017}).
\newblock


\bibitem[\protect\citeauthoryear{Bechavod, Ligett, Roth, Waggoner, and
  Wu}{Bechavod et~al\mbox{.}}{2019}]%
        {bechavod2019equal}
\bibfield{author}{\bibinfo{person}{Yahav Bechavod}, \bibinfo{person}{Katrina
  Ligett}, \bibinfo{person}{Aaron Roth}, \bibinfo{person}{Bo Waggoner}, {and}
  \bibinfo{person}{Steven~Z Wu}.} \bibinfo{year}{2019}\natexlab{}.
\newblock \showarticletitle{Equal opportunity in online classification with
  partial feedback}. In \bibinfo{booktitle}{\emph{Advances in Neural
  Information Processing Systems}}. \bibinfo{pages}{8974--8984}.
\newblock


\bibitem[\protect\citeauthoryear{Beck and Teboulle}{Beck and Teboulle}{2009}]%
        {beck2009fast}
\bibfield{author}{\bibinfo{person}{Amir Beck} {and} \bibinfo{person}{Marc
  Teboulle}.} \bibinfo{year}{2009}\natexlab{}.
\newblock \showarticletitle{A fast iterative shrinkage-thresholding algorithm
  for linear inverse problems}.
\newblock \bibinfo{journal}{\emph{SIAM journal on imaging sciences}}
  \bibinfo{volume}{2}, \bibinfo{number}{1} (\bibinfo{year}{2009}),
  \bibinfo{pages}{183--202}.
\newblock


\bibitem[\protect\citeauthoryear{Begoli, Bhattacharya, and Kusnezov}{Begoli
  et~al\mbox{.}}{2019}]%
        {begoli2019need}
\bibfield{author}{\bibinfo{person}{Edmon Begoli}, \bibinfo{person}{Tanmoy
  Bhattacharya}, {and} \bibinfo{person}{Dimitri Kusnezov}.}
  \bibinfo{year}{2019}\natexlab{}.
\newblock \showarticletitle{The need for uncertainty quantification in
  machine-assisted medical decision making}.
\newblock \bibinfo{journal}{\emph{Nature Machine Intelligence}}
  \bibinfo{volume}{1}, \bibinfo{number}{1} (\bibinfo{year}{2019}),
  \bibinfo{pages}{20--23}.
\newblock


\bibitem[\protect\citeauthoryear{Bertossi}{Bertossi}{2020}]%
        {bertossi2020asp}
\bibfield{author}{\bibinfo{person}{Leopoldo Bertossi}.}
  \bibinfo{year}{2020}\natexlab{}.
\newblock \showarticletitle{An ASP-Based Approach to Counterfactual
  Explanations for Classification}.
\newblock \bibinfo{journal}{\emph{arXiv preprint arXiv:2004.13237}}
  (\bibinfo{year}{2020}).
\newblock


\bibitem[\protect\citeauthoryear{Bhatt, Andrus, Weller, and Xiang}{Bhatt
  et~al\mbox{.}}{2020a}]%
        {bhatt2020machine}
\bibfield{author}{\bibinfo{person}{Umang Bhatt}, \bibinfo{person}{McKane
  Andrus}, \bibinfo{person}{Adrian Weller}, {and} \bibinfo{person}{Alice
  Xiang}.} \bibinfo{year}{2020}\natexlab{a}.
\newblock \showarticletitle{Machine Learning Explainability for External
  Stakeholders}.
\newblock \bibinfo{journal}{\emph{arXiv preprint arXiv:2007.05408}}
  (\bibinfo{year}{2020}).
\newblock


\bibitem[\protect\citeauthoryear{Bhatt, Xiang, Sharma, Weller, Taly, Jia,
  Ghosh, Puri, Moura, and Eckersley}{Bhatt et~al\mbox{.}}{2020b}]%
        {bhatt2020explainable}
\bibfield{author}{\bibinfo{person}{Umang Bhatt}, \bibinfo{person}{Alice Xiang},
  \bibinfo{person}{Shubham Sharma}, \bibinfo{person}{Adrian Weller},
  \bibinfo{person}{Ankur Taly}, \bibinfo{person}{Yunhan Jia},
  \bibinfo{person}{Joydeep Ghosh}, \bibinfo{person}{Ruchir Puri},
  \bibinfo{person}{Jos{\'e}~MF Moura}, {and} \bibinfo{person}{Peter
  Eckersley}.} \bibinfo{year}{2020}\natexlab{b}.
\newblock \showarticletitle{Explainable machine learning in deployment}. In
  \bibinfo{booktitle}{\emph{Proceedings of the 2020 Conference on Fairness,
  Accountability, and Transparency}}. \bibinfo{pages}{648--657}.
\newblock


\bibitem[\protect\citeauthoryear{Biran and Cotton}{Biran and Cotton}{[n.d.]}]%
        {biran2017explanation}
\bibfield{author}{\bibinfo{person}{Or Biran} {and} \bibinfo{person}{Courtenay
  Cotton}.} \bibinfo{year}{[n.d.]}\natexlab{}.
\newblock \showarticletitle{Explanation and justification in machine learning:
  A survey}.
\newblock


\bibitem[\protect\citeauthoryear{Boyd, Boyd, and Vandenberghe}{Boyd
  et~al\mbox{.}}{2004}]%
        {boyd2004convex}
\bibfield{author}{\bibinfo{person}{Stephen Boyd}, \bibinfo{person}{Stephen~P
  Boyd}, {and} \bibinfo{person}{Lieven Vandenberghe}.}
  \bibinfo{year}{2004}\natexlab{}.
\newblock \bibinfo{booktitle}{\emph{Convex optimization}}.
\newblock \bibinfo{publisher}{Cambridge university press}.
\newblock


\bibitem[\protect\citeauthoryear{Breiman et~al\mbox{.}}{Breiman
  et~al\mbox{.}}{2001}]%
        {breiman2001statistical}
\bibfield{author}{\bibinfo{person}{Leo Breiman} {et~al\mbox{.}}}
  \bibinfo{year}{2001}\natexlab{}.
\newblock \showarticletitle{Statistical modeling: The two cultures (with
  comments and a rejoinder by the author)}.
\newblock \bibinfo{journal}{\emph{Statistical science}} \bibinfo{volume}{16},
  \bibinfo{number}{3} (\bibinfo{year}{2001}), \bibinfo{pages}{199--231}.
\newblock


\bibitem[\protect\citeauthoryear{Burrell}{Burrell}{2016}]%
        {burrell2016machine}
\bibfield{author}{\bibinfo{person}{Jenna Burrell}.}
  \bibinfo{year}{2016}\natexlab{}.
\newblock \showarticletitle{How the machine ‘thinks’: Understanding opacity
  in machine learning algorithms}.
\newblock \bibinfo{journal}{\emph{Big Data \& Society}} \bibinfo{volume}{3},
  \bibinfo{number}{1} (\bibinfo{year}{2016}),
  \bibinfo{pages}{2053951715622512}.
\newblock


\bibitem[\protect\citeauthoryear{Byrne}{Byrne}{2019}]%
        {byrne2019counterfactuals}
\bibfield{author}{\bibinfo{person}{Ruth~MJ Byrne}.}
  \bibinfo{year}{2019}\natexlab{}.
\newblock \showarticletitle{Counterfactuals in Explainable Artificial
  Intelligence (XAI): Evidence from Human Reasoning.}. In
  \bibinfo{booktitle}{\emph{IJCAI}}. \bibinfo{pages}{6276--6282}.
\newblock


\bibitem[\protect\citeauthoryear{Cao, Luo, and Zhang}{Cao
  et~al\mbox{.}}{2007}]%
        {cao2007knowledge}
\bibfield{author}{\bibinfo{person}{Longbing Cao}, \bibinfo{person}{Dan Luo},
  {and} \bibinfo{person}{Chengqi Zhang}.} \bibinfo{year}{2007}\natexlab{}.
\newblock \showarticletitle{Knowledge actionability: satisfying technical and
  business interestingness}.
\newblock \bibinfo{journal}{\emph{International Journal of Business
  Intelligence and Data Mining}} \bibinfo{volume}{2}, \bibinfo{number}{4}
  (\bibinfo{year}{2007}), \bibinfo{pages}{496--514}.
\newblock


\bibitem[\protect\citeauthoryear{Cao and Zhang}{Cao and Zhang}{2006}]%
        {cao2006domain}
\bibfield{author}{\bibinfo{person}{Longbing Cao} {and} \bibinfo{person}{Chengqi
  Zhang}.} \bibinfo{year}{2006}\natexlab{}.
\newblock \showarticletitle{Domain-driven actionable knowledge discovery in the
  real world}. In \bibinfo{booktitle}{\emph{Pacific-Asia Conference on
  Knowledge Discovery and Data Mining}}. Springer, \bibinfo{pages}{821--830}.
\newblock


\bibitem[\protect\citeauthoryear{Cao, Zhao, Zhang, Luo, Zhang, and Park}{Cao
  et~al\mbox{.}}{2009}]%
        {cao2009flexible}
\bibfield{author}{\bibinfo{person}{Longbing Cao}, \bibinfo{person}{Yanchang
  Zhao}, \bibinfo{person}{Huaifeng Zhang}, \bibinfo{person}{Dan Luo},
  \bibinfo{person}{Chengqi Zhang}, {and} \bibinfo{person}{Eun~Kyo Park}.}
  \bibinfo{year}{2009}\natexlab{}.
\newblock \showarticletitle{Flexible frameworks for actionable knowledge
  discovery}.
\newblock \bibinfo{journal}{\emph{IEEE Transactions on Knowledge and Data
  Engineering}} \bibinfo{volume}{22}, \bibinfo{number}{9}
  (\bibinfo{year}{2009}), \bibinfo{pages}{1299--1312}.
\newblock


\bibitem[\protect\citeauthoryear{Carlini and Wagner}{Carlini and
  Wagner}{2017}]%
        {carlini2017towards}
\bibfield{author}{\bibinfo{person}{Nicholas Carlini} {and}
  \bibinfo{person}{David Wagner}.} \bibinfo{year}{2017}\natexlab{}.
\newblock \showarticletitle{Towards evaluating the robustness of neural
  networks}. In \bibinfo{booktitle}{\emph{2017 ieee symposium on security and
  privacy (sp)}}. IEEE, \bibinfo{pages}{39--57}.
\newblock


\bibitem[\protect\citeauthoryear{Carmon, Raghunathan, Schmidt, Liang, and
  Duchi}{Carmon et~al\mbox{.}}{2019}]%
        {carmon2019unlabeled}
\bibfield{author}{\bibinfo{person}{Yair Carmon}, \bibinfo{person}{Aditi
  Raghunathan}, \bibinfo{person}{Ludwig Schmidt}, \bibinfo{person}{Percy
  Liang}, {and} \bibinfo{person}{John~C Duchi}.}
  \bibinfo{year}{2019}\natexlab{}.
\newblock \showarticletitle{Unlabeled data improves adversarial robustness}.
\newblock \bibinfo{journal}{\emph{arXiv preprint arXiv:1905.13736}}
  (\bibinfo{year}{2019}).
\newblock


\bibitem[\protect\citeauthoryear{Chapman-Rounds, Schulz, Pazos, and
  Georgatzis}{Chapman-Rounds et~al\mbox{.}}{2019}]%
        {chapman2019emap}
\bibfield{author}{\bibinfo{person}{Matt Chapman-Rounds},
  \bibinfo{person}{Marc-Andre Schulz}, \bibinfo{person}{Erik Pazos}, {and}
  \bibinfo{person}{Konstantinos Georgatzis}.} \bibinfo{year}{2019}\natexlab{}.
\newblock \showarticletitle{EMAP: Explanation by Minimal Adversarial
  Perturbation}.
\newblock \bibinfo{journal}{\emph{arXiv preprint arXiv:1912.00872}}
  (\bibinfo{year}{2019}).
\newblock


\bibitem[\protect\citeauthoryear{Chen, Zhang, Hu, and Wu}{Chen
  et~al\mbox{.}}{2020b}]%
        {chen2020boosting}
\bibfield{author}{\bibinfo{person}{Weilun Chen}, \bibinfo{person}{Zhaoxiang
  Zhang}, \bibinfo{person}{Xiaolin Hu}, {and} \bibinfo{person}{Baoyuan Wu}.}
  \bibinfo{year}{2020}\natexlab{b}.
\newblock \showarticletitle{Boosting decision-based black-box adversarial
  attacks with random sign flip}. In \bibinfo{booktitle}{\emph{Proceedings of
  the European Conference on Computer Vision}}.
\newblock


\bibitem[\protect\citeauthoryear{Chen, Wang, and Liu}{Chen
  et~al\mbox{.}}{2020a}]%
        {chen2020strategic}
\bibfield{author}{\bibinfo{person}{Yatong Chen}, \bibinfo{person}{Jialu Wang},
  {and} \bibinfo{person}{Yang Liu}.} \bibinfo{year}{2020}\natexlab{a}.
\newblock \showarticletitle{Strategic Recourse in Linear Classification}.
\newblock \bibinfo{journal}{\emph{arXiv preprint arXiv:2011.00355}}
  (\bibinfo{year}{2020}).
\newblock


\bibitem[\protect\citeauthoryear{Cheng, Ming, and Qu}{Cheng
  et~al\mbox{.}}{2020}]%
        {cheng2020dece}
\bibfield{author}{\bibinfo{person}{Furui Cheng}, \bibinfo{person}{Yao Ming},
  {and} \bibinfo{person}{Huamin Qu}.} \bibinfo{year}{2020}\natexlab{}.
\newblock \showarticletitle{DECE: Decision Explorer with Counterfactual
  Explanations for Machine Learning Models}.
\newblock \bibinfo{journal}{\emph{arXiv preprint arXiv:2008.08353}}
  (\bibinfo{year}{2020}).
\newblock


\bibitem[\protect\citeauthoryear{Cohen, Rosenfeld, and Kolter}{Cohen
  et~al\mbox{.}}{2019b}]%
        {cohen2019certified}
\bibfield{author}{\bibinfo{person}{Jeremy Cohen}, \bibinfo{person}{Elan
  Rosenfeld}, {and} \bibinfo{person}{Zico Kolter}.}
  \bibinfo{year}{2019}\natexlab{b}.
\newblock \showarticletitle{Certified adversarial robustness via randomized
  smoothing}. In \bibinfo{booktitle}{\emph{International Conference on Machine
  Learning}}. PMLR, \bibinfo{pages}{1310--1320}.
\newblock


\bibitem[\protect\citeauthoryear{Cohen, Lipton, and Mansour}{Cohen
  et~al\mbox{.}}{2019a}]%
        {cohen2019efficient}
\bibfield{author}{\bibinfo{person}{Lee Cohen}, \bibinfo{person}{Zachary~C
  Lipton}, {and} \bibinfo{person}{Yishay Mansour}.}
  \bibinfo{year}{2019}\natexlab{a}.
\newblock \showarticletitle{Efficient candidate screening under multiple tests
  and implications for fairness}.
\newblock \bibinfo{journal}{\emph{arXiv preprint arXiv:1905.11361}}
  (\bibinfo{year}{2019}).
\newblock


\bibitem[\protect\citeauthoryear{Cooper and Yoo}{Cooper and Yoo}{1999}]%
        {cooper1999causal}
\bibfield{author}{\bibinfo{person}{Gregory~F Cooper} {and}
  \bibinfo{person}{Changwon Yoo}.} \bibinfo{year}{1999}\natexlab{}.
\newblock \showarticletitle{Causal discovery from a mixture of experimental and
  observational data}. In \bibinfo{booktitle}{\emph{Proceedings of the
  Fifteenth conference on Uncertainty in artificial intelligence}}.
  \bibinfo{pages}{116--125}.
\newblock


\bibitem[\protect\citeauthoryear{Corbett-Davies and Goel}{Corbett-Davies and
  Goel}{2018}]%
        {corbett2018measure}
\bibfield{author}{\bibinfo{person}{Sam Corbett-Davies} {and}
  \bibinfo{person}{Sharad Goel}.} \bibinfo{year}{2018}\natexlab{}.
\newblock \showarticletitle{The measure and mismeasure of fairness: A critical
  review of fair machine learning}.
\newblock \bibinfo{journal}{\emph{arXiv preprint arXiv:1808.00023}}
  (\bibinfo{year}{2018}).
\newblock


\bibitem[\protect\citeauthoryear{Cormen, Leiserson, Rivest, and Stein}{Cormen
  et~al\mbox{.}}{2009}]%
        {cormen2009introduction}
\bibfield{author}{\bibinfo{person}{Thomas~H Cormen}, \bibinfo{person}{Charles~E
  Leiserson}, \bibinfo{person}{Ronald~L Rivest}, {and}
  \bibinfo{person}{Clifford Stein}.} \bibinfo{year}{2009}\natexlab{}.
\newblock \bibinfo{booktitle}{\emph{Introduction to algorithms}}.
\newblock \bibinfo{publisher}{MIT press}.
\newblock


\bibitem[\protect\citeauthoryear{Cplex}{Cplex}{2009}]%
        {cplex2009v12}
\bibfield{author}{\bibinfo{person}{IBM~ILOG Cplex}.}
  \bibinfo{year}{2009}\natexlab{}.
\newblock \showarticletitle{V12. 1: User’s Manual for CPLEX}.
\newblock \bibinfo{journal}{\emph{International Business Machines Corporation}}
  \bibinfo{volume}{46}, \bibinfo{number}{53} (\bibinfo{year}{2009}),
  \bibinfo{pages}{157}.
\newblock


\bibitem[\protect\citeauthoryear{Cui, Chen, He, and Chen}{Cui
  et~al\mbox{.}}{2015}]%
        {cui2015optimal}
\bibfield{author}{\bibinfo{person}{Zhicheng Cui}, \bibinfo{person}{Wenlin
  Chen}, \bibinfo{person}{Yujie He}, {and} \bibinfo{person}{Yixin Chen}.}
  \bibinfo{year}{2015}\natexlab{}.
\newblock \showarticletitle{Optimal action extraction for random forests and
  boosted trees}. In \bibinfo{booktitle}{\emph{Proceedings of the 21th ACM
  SIGKDD international conference on knowledge discovery and data mining}}.
  \bibinfo{pages}{179--188}.
\newblock


\bibitem[\protect\citeauthoryear{Dandl, Molnar, Binder, and Bischl}{Dandl
  et~al\mbox{.}}{2020}]%
        {dandl2020multi}
\bibfield{author}{\bibinfo{person}{Susanne Dandl}, \bibinfo{person}{Christoph
  Molnar}, \bibinfo{person}{Martin Binder}, {and} \bibinfo{person}{Bernd
  Bischl}.} \bibinfo{year}{2020}\natexlab{}.
\newblock \showarticletitle{Multi-Objective Counterfactual Explanations}.
\newblock \bibinfo{journal}{\emph{arXiv preprint arXiv:2004.11165}}
  (\bibinfo{year}{2020}).
\newblock


\bibitem[\protect\citeauthoryear{De~Moura and Bj{\o}rner}{De~Moura and
  Bj{\o}rner}{2008}]%
        {de2008z3}
\bibfield{author}{\bibinfo{person}{Leonardo De~Moura} {and}
  \bibinfo{person}{Nikolaj Bj{\o}rner}.} \bibinfo{year}{2008}\natexlab{}.
\newblock \showarticletitle{Z3: An efficient SMT solver}. In
  \bibinfo{booktitle}{\emph{International conference on Tools and Algorithms
  for the Construction and Analysis of Systems}}. Springer,
  \bibinfo{pages}{337--340}.
\newblock


\bibitem[\protect\citeauthoryear{Dean, Rich, and Recht}{Dean
  et~al\mbox{.}}{2020}]%
        {dean2020recommendations}
\bibfield{author}{\bibinfo{person}{Sarah Dean}, \bibinfo{person}{Sarah Rich},
  {and} \bibinfo{person}{Benjamin Recht}.} \bibinfo{year}{2020}\natexlab{}.
\newblock \showarticletitle{Recommendations and user agency: the reachability
  of collaboratively-filtered information}. In
  \bibinfo{booktitle}{\emph{Proceedings of the 2020 Conference on Fairness,
  Accountability, and Transparency}}. \bibinfo{pages}{436--445}.
\newblock


\bibitem[\protect\citeauthoryear{Dhurandhar, Chen, Luss, Tu, Ting, Shanmugam,
  and Das}{Dhurandhar et~al\mbox{.}}{2018}]%
        {dhurandhar2018explanations}
\bibfield{author}{\bibinfo{person}{Amit Dhurandhar}, \bibinfo{person}{Pin-Yu
  Chen}, \bibinfo{person}{Ronny Luss}, \bibinfo{person}{Chun-Chen Tu},
  \bibinfo{person}{Paishun Ting}, \bibinfo{person}{Karthikeyan Shanmugam},
  {and} \bibinfo{person}{Payel Das}.} \bibinfo{year}{2018}\natexlab{}.
\newblock \showarticletitle{Explanations based on the missing: Towards
  contrastive explanations with pertinent negatives}. In
  \bibinfo{booktitle}{\emph{Advances in Neural Information Processing
  Systems}}. \bibinfo{pages}{592--603}.
\newblock


\bibitem[\protect\citeauthoryear{Dhurandhar, Pedapati, Balakrishnan, Chen,
  Shanmugam, and Puri}{Dhurandhar et~al\mbox{.}}{2019}]%
        {dhurandhar2019model}
\bibfield{author}{\bibinfo{person}{Amit Dhurandhar}, \bibinfo{person}{Tejaswini
  Pedapati}, \bibinfo{person}{Avinash Balakrishnan}, \bibinfo{person}{Pin-Yu
  Chen}, \bibinfo{person}{Karthikeyan Shanmugam}, {and} \bibinfo{person}{Ruchir
  Puri}.} \bibinfo{year}{2019}\natexlab{}.
\newblock \showarticletitle{Model agnostic contrastive explanations for
  structured data}.
\newblock \bibinfo{journal}{\emph{arXiv preprint arXiv:1906.00117}}
  (\bibinfo{year}{2019}).
\newblock


\bibitem[\protect\citeauthoryear{Dombrowski, Alber, Anders, Ackermann,
  M{\"u}ller, and Kessel}{Dombrowski et~al\mbox{.}}{2019}]%
        {dombrowski2019explanations}
\bibfield{author}{\bibinfo{person}{Ann-Kathrin Dombrowski},
  \bibinfo{person}{Maximilian Alber}, \bibinfo{person}{Christopher~J Anders},
  \bibinfo{person}{Marcel Ackermann}, \bibinfo{person}{Klaus-Robert
  M{\"u}ller}, {and} \bibinfo{person}{Pan Kessel}.}
  \bibinfo{year}{2019}\natexlab{}.
\newblock \showarticletitle{Explanations can be manipulated and geometry is to
  blame}.
\newblock \bibinfo{journal}{\emph{arXiv preprint arXiv:1906.07983}}
  (\bibinfo{year}{2019}).
\newblock


\bibitem[\protect\citeauthoryear{Dong, Roth, Schutzman, Waggoner, and Wu}{Dong
  et~al\mbox{.}}{2018}]%
        {dong2018strategic}
\bibfield{author}{\bibinfo{person}{Jinshuo Dong}, \bibinfo{person}{Aaron Roth},
  \bibinfo{person}{Zachary Schutzman}, \bibinfo{person}{Bo Waggoner}, {and}
  \bibinfo{person}{Zhiwei~Steven Wu}.} \bibinfo{year}{2018}\natexlab{}.
\newblock \showarticletitle{Strategic classification from revealed
  preferences}. In \bibinfo{booktitle}{\emph{Proceedings of the 2018 ACM
  Conference on Economics and Computation}}. \bibinfo{pages}{55--70}.
\newblock


\bibitem[\protect\citeauthoryear{Doshi-Velez and Kim}{Doshi-Velez and
  Kim}{2017}]%
        {doshi2017towards}
\bibfield{author}{\bibinfo{person}{Finale Doshi-Velez} {and}
  \bibinfo{person}{Been Kim}.} \bibinfo{year}{2017}\natexlab{}.
\newblock \showarticletitle{Towards a rigorous science of interpretable machine
  learning}.
\newblock \bibinfo{journal}{\emph{arXiv preprint arXiv:1702.08608}}
  (\bibinfo{year}{2017}).
\newblock


\bibitem[\protect\citeauthoryear{Downs, Chu, Yacoby, Doshi-Velez, and
  Pan}{Downs et~al\mbox{.}}{[n.d.]}]%
        {downscruds}
\bibfield{author}{\bibinfo{person}{Michael Downs}, \bibinfo{person}{Jonathan~L
  Chu}, \bibinfo{person}{Yaniv Yacoby}, \bibinfo{person}{Finale Doshi-Velez},
  {and} \bibinfo{person}{Weiwei Pan}.} \bibinfo{year}{[n.d.]}\natexlab{}.
\newblock \showarticletitle{CRUDS: Counterfactual Recourse Using Disentangled
  Subspaces}.
\newblock  (\bibinfo{year}{[n.\,d.]}).
\newblock


\bibitem[\protect\citeauthoryear{Du, Hu, Ling, Fan, and Liu}{Du
  et~al\mbox{.}}{2011}]%
        {du2011efficient}
\bibfield{author}{\bibinfo{person}{Jianfeng Du}, \bibinfo{person}{Yong Hu},
  \bibinfo{person}{Charles~X Ling}, \bibinfo{person}{Ming Fan}, {and}
  \bibinfo{person}{Mei Liu}.} \bibinfo{year}{2011}\natexlab{}.
\newblock \showarticletitle{Efficient action extraction with many-to-many
  relationship between actions and features}. In
  \bibinfo{booktitle}{\emph{International Workshop on Logic, Rationality and
  Interaction}}. Springer, \bibinfo{pages}{384--385}.
\newblock


\bibitem[\protect\citeauthoryear{Dua and Graff}{Dua and Graff}{2017}]%
        {dua2019uci}
\bibfield{author}{\bibinfo{person}{Dheeru Dua} {and} \bibinfo{person}{Casey
  Graff}.} \bibinfo{year}{2017}\natexlab{}.
\newblock \bibinfo{title}{{UCI} Machine Learning Repository}.
\newblock
\newblock
\urldef\tempurl%
\url{http://archive.ics.uci.edu/ml}
\showURL{%
\tempurl}


\bibitem[\protect\citeauthoryear{Dwork and Feldman}{Dwork and Feldman}{2018}]%
        {dwork2018privacy}
\bibfield{author}{\bibinfo{person}{Cynthia Dwork} {and} \bibinfo{person}{Vitaly
  Feldman}.} \bibinfo{year}{2018}\natexlab{}.
\newblock \showarticletitle{Privacy-preserving prediction}.
\newblock \bibinfo{journal}{\emph{arXiv preprint arXiv:1803.10266}}
  (\bibinfo{year}{2018}).
\newblock


\bibitem[\protect\citeauthoryear{Dwork, Hardt, Pitassi, Reingold, and
  Zemel}{Dwork et~al\mbox{.}}{2012}]%
        {dwork2012fairness}
\bibfield{author}{\bibinfo{person}{Cynthia Dwork}, \bibinfo{person}{Moritz
  Hardt}, \bibinfo{person}{Toniann Pitassi}, \bibinfo{person}{Omer Reingold},
  {and} \bibinfo{person}{Richard Zemel}.} \bibinfo{year}{2012}\natexlab{}.
\newblock \showarticletitle{Fairness through awareness}. In
  \bibinfo{booktitle}{\emph{Proceedings of the 3rd innovations in theoretical
  computer science conference}}. ACM, \bibinfo{pages}{214--226}.
\newblock


\bibitem[\protect\citeauthoryear{Fawzi, Fawzi, and Frossard}{Fawzi
  et~al\mbox{.}}{2015}]%
        {fawzi2015fundamental}
\bibfield{author}{\bibinfo{person}{Alhussein Fawzi}, \bibinfo{person}{Omar
  Fawzi}, {and} \bibinfo{person}{Pascal Frossard}.}
  \bibinfo{year}{2015}\natexlab{}.
\newblock \showarticletitle{Fundamental limits on adversarial robustness}. In
  \bibinfo{booktitle}{\emph{Proc. ICML, Workshop on Deep Learning}}.
\newblock


\bibitem[\protect\citeauthoryear{Fernandez, Provost, and Han}{Fernandez
  et~al\mbox{.}}{2020}]%
        {fernandez2020explaining}
\bibfield{author}{\bibinfo{person}{Carlos Fernandez}, \bibinfo{person}{Foster
  Provost}, {and} \bibinfo{person}{Xintian Han}.}
  \bibinfo{year}{2020}\natexlab{}.
\newblock \showarticletitle{Explaining data-driven decisions made by ai
  systems: The counterfactual approach}.
\newblock \bibinfo{journal}{\emph{arXiv preprint arXiv:2001.07417}}
  (\bibinfo{year}{2020}).
\newblock


\bibitem[\protect\citeauthoryear{Freitas}{Freitas}{2014}]%
        {freitas2014comprehensible}
\bibfield{author}{\bibinfo{person}{Alex~A Freitas}.}
  \bibinfo{year}{2014}\natexlab{}.
\newblock \showarticletitle{Comprehensible classification models: a position
  paper}.
\newblock \bibinfo{journal}{\emph{ACM SIGKDD explorations newsletter}}
  \bibinfo{volume}{15}, \bibinfo{number}{1} (\bibinfo{year}{2014}),
  \bibinfo{pages}{1--10}.
\newblock


\bibitem[\protect\citeauthoryear{Gario and Micheli}{Gario and Micheli}{2015}]%
        {gario2015pysmt}
\bibfield{author}{\bibinfo{person}{Marco Gario} {and} \bibinfo{person}{Andrea
  Micheli}.} \bibinfo{year}{2015}\natexlab{}.
\newblock \showarticletitle{PySMT: a solver-agnostic library for fast
  prototyping of SMT-based algorithms}. In \bibinfo{booktitle}{\emph{SMT
  workshop}}, Vol.~\bibinfo{volume}{2015}.
\newblock


\bibitem[\protect\citeauthoryear{Gelman}{Gelman}{2011}]%
        {gelman2011causality}
\bibfield{author}{\bibinfo{person}{Andrew Gelman}.}
  \bibinfo{year}{2011}\natexlab{}.
\newblock \bibinfo{title}{Causality and statistical learning}.
\newblock
\newblock


\bibitem[\protect\citeauthoryear{Gerstenberg, Peterson, Goodman, Lagnado, and
  Tenenbaum}{Gerstenberg et~al\mbox{.}}{2017}]%
        {gerstenberg2017eye}
\bibfield{author}{\bibinfo{person}{Tobias Gerstenberg},
  \bibinfo{person}{Matthew~F Peterson}, \bibinfo{person}{Noah~D Goodman},
  \bibinfo{person}{David~A Lagnado}, {and} \bibinfo{person}{Joshua~B
  Tenenbaum}.} \bibinfo{year}{2017}\natexlab{}.
\newblock \showarticletitle{Eye-tracking causality}.
\newblock \bibinfo{journal}{\emph{Psychological science}} \bibinfo{volume}{28},
  \bibinfo{number}{12} (\bibinfo{year}{2017}), \bibinfo{pages}{1731--1744}.
\newblock


\bibitem[\protect\citeauthoryear{Ghazimatin, Balalau, Saha~Roy, and
  Weikum}{Ghazimatin et~al\mbox{.}}{2020}]%
        {ghazimatin2020prince}
\bibfield{author}{\bibinfo{person}{Azin Ghazimatin}, \bibinfo{person}{Oana
  Balalau}, \bibinfo{person}{Rishiraj Saha~Roy}, {and} \bibinfo{person}{Gerhard
  Weikum}.} \bibinfo{year}{2020}\natexlab{}.
\newblock \showarticletitle{PRINCE: Provider-side Interpretability with
  Counterfactual Explanations in Recommender Systems}. In
  \bibinfo{booktitle}{\emph{Proceedings of the 13th International Conference on
  Web Search and Data Mining}}. \bibinfo{pages}{196--204}.
\newblock


\bibitem[\protect\citeauthoryear{Ghorbani, Abid, and Zou}{Ghorbani
  et~al\mbox{.}}{2019}]%
        {ghorbani2019interpretation}
\bibfield{author}{\bibinfo{person}{Amirata Ghorbani}, \bibinfo{person}{Abubakar
  Abid}, {and} \bibinfo{person}{James Zou}.} \bibinfo{year}{2019}\natexlab{}.
\newblock \showarticletitle{Interpretation of neural networks is fragile}. In
  \bibinfo{booktitle}{\emph{Proceedings of the AAAI Conference on Artificial
  Intelligence}}, Vol.~\bibinfo{volume}{33}. \bibinfo{pages}{3681--3688}.
\newblock


\bibitem[\protect\citeauthoryear{Gilpin, Bau, Yuan, Bajwa, Specter, and
  Kagal}{Gilpin et~al\mbox{.}}{2018}]%
        {gilpin2018explaining}
\bibfield{author}{\bibinfo{person}{Leilani~H Gilpin}, \bibinfo{person}{David
  Bau}, \bibinfo{person}{Ben~Z Yuan}, \bibinfo{person}{Ayesha Bajwa},
  \bibinfo{person}{Michael Specter}, {and} \bibinfo{person}{Lalana Kagal}.}
  \bibinfo{year}{2018}\natexlab{}.
\newblock \showarticletitle{Explaining explanations: An overview of
  interpretability of machine learning}. In \bibinfo{booktitle}{\emph{2018 IEEE
  5th International Conference on data science and advanced analytics (DSAA)}}.
  IEEE, \bibinfo{pages}{80--89}.
\newblock


\bibitem[\protect\citeauthoryear{Gomez, Holter, Yuan, and Bertini}{Gomez
  et~al\mbox{.}}{2020}]%
        {gomez2020vice}
\bibfield{author}{\bibinfo{person}{Oscar Gomez}, \bibinfo{person}{Steffen
  Holter}, \bibinfo{person}{Jun Yuan}, {and} \bibinfo{person}{Enrico Bertini}.}
  \bibinfo{year}{2020}\natexlab{}.
\newblock \showarticletitle{ViCE: visual counterfactual explanations for
  machine learning models}. In \bibinfo{booktitle}{\emph{Proceedings of the
  25th International Conference on Intelligent User Interfaces}}.
  \bibinfo{pages}{531--535}.
\newblock


\bibitem[\protect\citeauthoryear{Goodfellow, Shlens, and Szegedy}{Goodfellow
  et~al\mbox{.}}{2014}]%
        {goodfellow2014explaining}
\bibfield{author}{\bibinfo{person}{Ian~J Goodfellow}, \bibinfo{person}{Jonathon
  Shlens}, {and} \bibinfo{person}{Christian Szegedy}.}
  \bibinfo{year}{2014}\natexlab{}.
\newblock \showarticletitle{Explaining and harnessing adversarial examples}.
\newblock \bibinfo{journal}{\emph{arXiv preprint arXiv:1412.6572}}
  (\bibinfo{year}{2014}).
\newblock


\bibitem[\protect\citeauthoryear{Goyal, Wu, Ernst, Batra, Parikh, and
  Lee}{Goyal et~al\mbox{.}}{2019}]%
        {goyal2019counterfactual}
\bibfield{author}{\bibinfo{person}{Yash Goyal}, \bibinfo{person}{Ziyan Wu},
  \bibinfo{person}{Jan Ernst}, \bibinfo{person}{Dhruv Batra},
  \bibinfo{person}{Devi Parikh}, {and} \bibinfo{person}{Stefan Lee}.}
  \bibinfo{year}{2019}\natexlab{}.
\newblock \showarticletitle{Counterfactual visual explanations}.
\newblock \bibinfo{journal}{\emph{arXiv preprint arXiv:1904.07451}}
  (\bibinfo{year}{2019}).
\newblock


\bibitem[\protect\citeauthoryear{Grath, Costabello, Van, Sweeney, Kamiab, Shen,
  and Lecue}{Grath et~al\mbox{.}}{2018}]%
        {grath2018interpretable}
\bibfield{author}{\bibinfo{person}{Rory~Mc Grath}, \bibinfo{person}{Luca
  Costabello}, \bibinfo{person}{Chan~Le Van}, \bibinfo{person}{Paul Sweeney},
  \bibinfo{person}{Farbod Kamiab}, \bibinfo{person}{Zhao Shen}, {and}
  \bibinfo{person}{Freddy Lecue}.} \bibinfo{year}{2018}\natexlab{}.
\newblock \showarticletitle{Interpretable credit application predictions with
  counterfactual explanations}.
\newblock \bibinfo{journal}{\emph{arXiv preprint arXiv:1811.05245}}
  (\bibinfo{year}{2018}).
\newblock


\bibitem[\protect\citeauthoryear{Grote and Berens}{Grote and Berens}{2020}]%
        {grote2020ethics}
\bibfield{author}{\bibinfo{person}{Thomas Grote} {and} \bibinfo{person}{Philipp
  Berens}.} \bibinfo{year}{2020}\natexlab{}.
\newblock \showarticletitle{On the ethics of algorithmic decision-making in
  healthcare}.
\newblock \bibinfo{journal}{\emph{Journal of medical ethics}}
  \bibinfo{volume}{46}, \bibinfo{number}{3} (\bibinfo{year}{2020}),
  \bibinfo{pages}{205--211}.
\newblock


\bibitem[\protect\citeauthoryear{Guidotti, Monreale, Matwin, and
  Pedreschi}{Guidotti et~al\mbox{.}}{2019}]%
        {guidotti2019black}
\bibfield{author}{\bibinfo{person}{Riccardo Guidotti}, \bibinfo{person}{Anna
  Monreale}, \bibinfo{person}{Stan Matwin}, {and} \bibinfo{person}{Dino
  Pedreschi}.} \bibinfo{year}{2019}\natexlab{}.
\newblock \showarticletitle{Black Box Explanation by Learning Image Exemplars
  in the Latent Feature Space}. In \bibinfo{booktitle}{\emph{Joint European
  Conference on Machine Learning and Knowledge Discovery in Databases}}.
  Springer, \bibinfo{pages}{189--205}.
\newblock


\bibitem[\protect\citeauthoryear{Guidotti, Monreale, Ruggieri, Pedreschi,
  Turini, and Giannotti}{Guidotti et~al\mbox{.}}{2018a}]%
        {guidotti2018local}
\bibfield{author}{\bibinfo{person}{Riccardo Guidotti}, \bibinfo{person}{Anna
  Monreale}, \bibinfo{person}{Salvatore Ruggieri}, \bibinfo{person}{Dino
  Pedreschi}, \bibinfo{person}{Franco Turini}, {and} \bibinfo{person}{Fosca
  Giannotti}.} \bibinfo{year}{2018}\natexlab{a}.
\newblock \showarticletitle{Local rule-based explanations of black box decision
  systems}.
\newblock \bibinfo{journal}{\emph{arXiv preprint arXiv:1805.10820}}
  (\bibinfo{year}{2018}).
\newblock


\bibitem[\protect\citeauthoryear{Guidotti, Monreale, Ruggieri, Turini,
  Giannotti, and Pedreschi}{Guidotti et~al\mbox{.}}{2018b}]%
        {guidotti2018survey}
\bibfield{author}{\bibinfo{person}{Riccardo Guidotti}, \bibinfo{person}{Anna
  Monreale}, \bibinfo{person}{Salvatore Ruggieri}, \bibinfo{person}{Franco
  Turini}, \bibinfo{person}{Fosca Giannotti}, {and} \bibinfo{person}{Dino
  Pedreschi}.} \bibinfo{year}{2018}\natexlab{b}.
\newblock \showarticletitle{A survey of methods for explaining black box
  models}.
\newblock \bibinfo{journal}{\emph{ACM computing surveys (CSUR)}}
  \bibinfo{volume}{51}, \bibinfo{number}{5} (\bibinfo{year}{2018}),
  \bibinfo{pages}{1--42}.
\newblock


\bibitem[\protect\citeauthoryear{Guo, Cheng, Li, Hahn, and Liu}{Guo
  et~al\mbox{.}}{2018}]%
        {guo2018survey}
\bibfield{author}{\bibinfo{person}{Ruocheng Guo}, \bibinfo{person}{Lu Cheng},
  \bibinfo{person}{Jundong Li}, \bibinfo{person}{P~Richard Hahn}, {and}
  \bibinfo{person}{Huan Liu}.} \bibinfo{year}{2018}\natexlab{}.
\newblock \showarticletitle{A survey of learning causality with data: Problems
  and methods}.
\newblock \bibinfo{journal}{\emph{arXiv preprint arXiv:1809.09337}}
  (\bibinfo{year}{2018}).
\newblock


\bibitem[\protect\citeauthoryear{Gupta, Nokhiz, Roy, and
  Venkatasubramanian}{Gupta et~al\mbox{.}}{2019}]%
        {gupta2019equalizing}
\bibfield{author}{\bibinfo{person}{Vivek Gupta}, \bibinfo{person}{Pegah
  Nokhiz}, \bibinfo{person}{Chitradeep~Dutta Roy}, {and}
  \bibinfo{person}{Suresh Venkatasubramanian}.}
  \bibinfo{year}{2019}\natexlab{}.
\newblock \showarticletitle{Equalizing Recourse across Groups}.
\newblock \bibinfo{journal}{\emph{arXiv preprint arXiv:1909.03166}}
  (\bibinfo{year}{2019}).
\newblock


\bibitem[\protect\citeauthoryear{Halpern and Pearl}{Halpern and Pearl}{2005}]%
        {halpern2005causes}
\bibfield{author}{\bibinfo{person}{Joseph~Y Halpern} {and}
  \bibinfo{person}{Judea Pearl}.} \bibinfo{year}{2005}\natexlab{}.
\newblock \showarticletitle{Causes and explanations: A structural-model
  approach. Part I: Causes}.
\newblock \bibinfo{journal}{\emph{The British journal for the philosophy of
  science}} \bibinfo{volume}{56}, \bibinfo{number}{4} (\bibinfo{year}{2005}),
  \bibinfo{pages}{843--887}.
\newblock


\bibitem[\protect\citeauthoryear{Hancox-Li}{Hancox-Li}{2020}]%
        {hancox2020robustness}
\bibfield{author}{\bibinfo{person}{Leif Hancox-Li}.}
  \bibinfo{year}{2020}\natexlab{}.
\newblock \showarticletitle{Robustness in machine learning explanations: does
  it matter?}. In \bibinfo{booktitle}{\emph{Proceedings of the 2020 Conference
  on Fairness, Accountability, and Transparency}}. \bibinfo{pages}{640--647}.
\newblock


\bibitem[\protect\citeauthoryear{Hardt, Megiddo, Papadimitriou, and
  Wootters}{Hardt et~al\mbox{.}}{2016}]%
        {hardt2016strategic}
\bibfield{author}{\bibinfo{person}{Moritz Hardt}, \bibinfo{person}{Nimrod
  Megiddo}, \bibinfo{person}{Christos Papadimitriou}, {and}
  \bibinfo{person}{Mary Wootters}.} \bibinfo{year}{2016}\natexlab{}.
\newblock \showarticletitle{Strategic classification}. In
  \bibinfo{booktitle}{\emph{Proceedings of the 2016 ACM conference on
  innovations in theoretical computer science}}. \bibinfo{pages}{111--122}.
\newblock


\bibitem[\protect\citeauthoryear{Hashemi and Fathi}{Hashemi and Fathi}{2020}]%
        {hashemi2020permuteattack}
\bibfield{author}{\bibinfo{person}{Masoud Hashemi} {and} \bibinfo{person}{Ali
  Fathi}.} \bibinfo{year}{2020}\natexlab{}.
\newblock \showarticletitle{PermuteAttack: Counterfactual Explanation of
  Machine Learning Credit Scorecards}.
\newblock \bibinfo{journal}{\emph{arXiv preprint arXiv:2008.10138}}
  (\bibinfo{year}{2020}).
\newblock


\bibitem[\protect\citeauthoryear{Hilton}{Hilton}{1990}]%
        {hilton1990conversational}
\bibfield{author}{\bibinfo{person}{Denis~J Hilton}.}
  \bibinfo{year}{1990}\natexlab{}.
\newblock \showarticletitle{Conversational processes and causal explanation.}
\newblock \bibinfo{journal}{\emph{Psychological Bulletin}}
  \bibinfo{volume}{107}, \bibinfo{number}{1} (\bibinfo{year}{1990}),
  \bibinfo{pages}{65}.
\newblock


\bibitem[\protect\citeauthoryear{Hilton and Slugoski}{Hilton and
  Slugoski}{1986}]%
        {hilton1986knowledge}
\bibfield{author}{\bibinfo{person}{Denis~J Hilton} {and} \bibinfo{person}{Ben~R
  Slugoski}.} \bibinfo{year}{1986}\natexlab{}.
\newblock \showarticletitle{Knowledge-based causal attribution: The abnormal
  conditions focus model.}
\newblock \bibinfo{journal}{\emph{Psychological review}} \bibinfo{volume}{93},
  \bibinfo{number}{1} (\bibinfo{year}{1986}), \bibinfo{pages}{75}.
\newblock


\bibitem[\protect\citeauthoryear{Holter, Gomez, and Bertini}{Holter
  et~al\mbox{.}}{[n.d.]}]%
        {holter2017fico}
\bibfield{author}{\bibinfo{person}{Steffen Holter}, \bibinfo{person}{Oscar
  Gomez}, {and} \bibinfo{person}{Enrico Bertini}.}
  \bibinfo{year}{[n.d.]}\natexlab{}.
\newblock \showarticletitle{FICO Explainable Machine Learning Challenge}.
\newblock  (\bibinfo{year}{[n.\,d.]}).
\newblock
\urldef\tempurl%
\url{https://community.fico.com/s/explainable-machine-learning-challenge}
\showURL{%
\tempurl}


\bibitem[\protect\citeauthoryear{Hu, Immorlica, and Vaughan}{Hu
  et~al\mbox{.}}{2019}]%
        {hu2019disparate}
\bibfield{author}{\bibinfo{person}{Lily Hu}, \bibinfo{person}{Nicole
  Immorlica}, {and} \bibinfo{person}{Jennifer~Wortman Vaughan}.}
  \bibinfo{year}{2019}\natexlab{}.
\newblock \showarticletitle{The disparate effects of strategic manipulation}.
  In \bibinfo{booktitle}{\emph{Proceedings of the Conference on Fairness,
  Accountability, and Transparency}}. \bibinfo{pages}{259--268}.
\newblock


\bibitem[\protect\citeauthoryear{Huk~Park, Anne~Hendricks, Akata, Rohrbach,
  Schiele, Darrell, and Rohrbach}{Huk~Park et~al\mbox{.}}{2018}]%
        {huk2018multimodal}
\bibfield{author}{\bibinfo{person}{Dong Huk~Park}, \bibinfo{person}{Lisa
  Anne~Hendricks}, \bibinfo{person}{Zeynep Akata}, \bibinfo{person}{Anna
  Rohrbach}, \bibinfo{person}{Bernt Schiele}, \bibinfo{person}{Trevor Darrell},
  {and} \bibinfo{person}{Marcus Rohrbach}.} \bibinfo{year}{2018}\natexlab{}.
\newblock \showarticletitle{Multimodal explanations: Justifying decisions and
  pointing to the evidence}. In \bibinfo{booktitle}{\emph{Proceedings of the
  IEEE Conference on Computer Vision and Pattern Recognition}}.
  \bibinfo{pages}{8779--8788}.
\newblock


\bibitem[\protect\citeauthoryear{Ilvento}{Ilvento}{2019}]%
        {ilvento2019metric}
\bibfield{author}{\bibinfo{person}{Christina Ilvento}.}
  \bibinfo{year}{2019}\natexlab{}.
\newblock \showarticletitle{Metric Learning for Individual Fairness}.
\newblock \bibinfo{journal}{\emph{arXiv preprint arXiv:1906.00250}}
  (\bibinfo{year}{2019}).
\newblock


\bibitem[\protect\citeauthoryear{Ilyas, Engstrom, Athalye, and Lin}{Ilyas
  et~al\mbox{.}}{2018}]%
        {ilyas2018black}
\bibfield{author}{\bibinfo{person}{Andrew Ilyas}, \bibinfo{person}{Logan
  Engstrom}, \bibinfo{person}{Anish Athalye}, {and} \bibinfo{person}{Jessy
  Lin}.} \bibinfo{year}{2018}\natexlab{}.
\newblock \showarticletitle{Black-box adversarial attacks with limited queries
  and information}.
\newblock \bibinfo{journal}{\emph{arXiv preprint arXiv:1804.08598}}
  (\bibinfo{year}{2018}).
\newblock


\bibitem[\protect\citeauthoryear{Joshi, Koyejo, Vijitbenjaronk, Kim, and
  Ghosh}{Joshi et~al\mbox{.}}{2019}]%
        {joshi2019towards}
\bibfield{author}{\bibinfo{person}{Shalmali Joshi}, \bibinfo{person}{Oluwasanmi
  Koyejo}, \bibinfo{person}{Warut Vijitbenjaronk}, \bibinfo{person}{Been Kim},
  {and} \bibinfo{person}{Joydeep Ghosh}.} \bibinfo{year}{2019}\natexlab{}.
\newblock \showarticletitle{{REVISE}: Towards Realistic Individual Recourse and
  Actionable Explanations in Black-Box Decision Making Systems}.
\newblock \bibinfo{journal}{\emph{arXiv preprint arXiv:1907.09615}}
  (\bibinfo{year}{2019}).
\newblock


\bibitem[\protect\citeauthoryear{Kanamori, Takagi, Kobayashi, and
  Arimura}{Kanamori et~al\mbox{.}}{[n.d.]}]%
        {kanamoridace}
\bibfield{author}{\bibinfo{person}{Kentaro Kanamori}, \bibinfo{person}{Takuya
  Takagi}, \bibinfo{person}{Ken Kobayashi}, {and} \bibinfo{person}{Hiroki
  Arimura}.} \bibinfo{year}{[n.d.]}\natexlab{}.
\newblock \showarticletitle{DACE: Distribution-Aware Counterfactual Explanation
  by Mixed-Integer Linear Optimization}.
\newblock  (\bibinfo{year}{[n.\,d.]}).
\newblock


\bibitem[\protect\citeauthoryear{Kang, Jung, Won, and Lee}{Kang
  et~al\mbox{.}}{2020}]%
        {kang2020counterfactual}
\bibfield{author}{\bibinfo{person}{Sin-Han Kang}, \bibinfo{person}{Hong-Gyu
  Jung}, \bibinfo{person}{Dong-Ok Won}, {and} \bibinfo{person}{Seong-Whan
  Lee}.} \bibinfo{year}{2020}\natexlab{}.
\newblock \showarticletitle{Counterfactual Explanation Based on Gradual
  Construction for Deep Networks}.
\newblock \bibinfo{journal}{\emph{arXiv preprint arXiv:2008.01897}}
  (\bibinfo{year}{2020}).
\newblock


\bibitem[\protect\citeauthoryear{Karim and Rahman}{Karim and Rahman}{2013}]%
        {karim2013decision}
\bibfield{author}{\bibinfo{person}{Masud Karim} {and}
  \bibinfo{person}{Rashedur~M Rahman}.} \bibinfo{year}{2013}\natexlab{}.
\newblock \showarticletitle{Decision tree and naive bayes algorithm for
  classification and generation of actionable knowledge for direct marketing}.
\newblock  (\bibinfo{year}{2013}).
\newblock


\bibitem[\protect\citeauthoryear{Karimi, Barthe, Balle, and Valera}{Karimi
  et~al\mbox{.}}{2020a}]%
        {karimi2020model}
\bibfield{author}{\bibinfo{person}{Amir-Hossein Karimi},
  \bibinfo{person}{Gilles Barthe}, \bibinfo{person}{Borja Balle}, {and}
  \bibinfo{person}{Isabel Valera}.} \bibinfo{year}{2020}\natexlab{a}.
\newblock \showarticletitle{Model-agnostic counterfactual explanations for
  consequential decisions}. In \bibinfo{booktitle}{\emph{International
  Conference on Artificial Intelligence and Statistics}}.
  \bibinfo{pages}{895--905}.
\newblock


\bibitem[\protect\citeauthoryear{Karimi, Sch{\"o}lkopf, and Valera}{Karimi
  et~al\mbox{.}}{2020b}]%
        {karimi2020mintrecourse}
\bibfield{author}{\bibinfo{person}{Amir-Hossein Karimi},
  \bibinfo{person}{Bernhard Sch{\"o}lkopf}, {and} \bibinfo{person}{Isabel
  Valera}.} \bibinfo{year}{2020}\natexlab{b}.
\newblock \showarticletitle{Algorithmic Recourse: from Counterfactual
  Explanations to Interventions}.
\newblock \bibinfo{journal}{\emph{arXiv preprint arXiv:2002.06278}}
  (\bibinfo{year}{2020}).
\newblock


\bibitem[\protect\citeauthoryear{Karimi, von K{\"u}gelgen, Sch{\"o}lkopf, and
  Valera}{Karimi et~al\mbox{.}}{2020c}]%
        {karimi2020caterecourse}
\bibfield{author}{\bibinfo{person}{Amir-Hossein Karimi},
  \bibinfo{person}{Julius von K{\"u}gelgen}, \bibinfo{person}{Bernhard
  Sch{\"o}lkopf}, {and} \bibinfo{person}{Isabel Valera}.}
  \bibinfo{year}{2020}\natexlab{c}.
\newblock \showarticletitle{Algorithmic recourse under imperfect causal
  knowledge: a probabilistic approach}.
\newblock \bibinfo{journal}{\emph{arXiv preprint arXiv:2006.06831}}
  (\bibinfo{year}{2020}).
\newblock


\bibitem[\protect\citeauthoryear{Katz, Barrett, Dill, Julian, and
  Kochenderfer}{Katz et~al\mbox{.}}{2017}]%
        {katz2017reluplex}
\bibfield{author}{\bibinfo{person}{Guy Katz}, \bibinfo{person}{Clark Barrett},
  \bibinfo{person}{David~L Dill}, \bibinfo{person}{Kyle Julian}, {and}
  \bibinfo{person}{Mykel~J Kochenderfer}.} \bibinfo{year}{2017}\natexlab{}.
\newblock \showarticletitle{Reluplex: An efficient SMT solver for verifying
  deep neural networks}. In \bibinfo{booktitle}{\emph{International Conference
  on Computer Aided Verification}}. Springer, \bibinfo{pages}{97--117}.
\newblock


\bibitem[\protect\citeauthoryear{Keane and Smyth}{Keane and Smyth}{2020}]%
        {keane2020good}
\bibfield{author}{\bibinfo{person}{Mark~T Keane} {and} \bibinfo{person}{Barry
  Smyth}.} \bibinfo{year}{2020}\natexlab{}.
\newblock \showarticletitle{Good Counterfactuals and Where to Find Them: A
  Case-Based Technique for Generating Counterfactuals for Explainable AI
  (XAI)}.
\newblock \bibinfo{journal}{\emph{arXiv preprint arXiv:2005.13997}}
  (\bibinfo{year}{2020}).
\newblock


\bibitem[\protect\citeauthoryear{Kilbertus, Rodriguez, Sch{\"o}lkopf, Muandet,
  and Valera}{Kilbertus et~al\mbox{.}}{2020}]%
        {kilbertus2020fair}
\bibfield{author}{\bibinfo{person}{Niki Kilbertus},
  \bibinfo{person}{Manuel~Gomez Rodriguez}, \bibinfo{person}{Bernhard
  Sch{\"o}lkopf}, \bibinfo{person}{Krikamol Muandet}, {and}
  \bibinfo{person}{Isabel Valera}.} \bibinfo{year}{2020}\natexlab{}.
\newblock \showarticletitle{Fair decisions despite imperfect predictions}. In
  \bibinfo{booktitle}{\emph{International Conference on Artificial Intelligence
  and Statistics}}. \bibinfo{pages}{277--287}.
\newblock


\bibitem[\protect\citeauthoryear{Kleinberg and Raghavan}{Kleinberg and
  Raghavan}{2020}]%
        {kleinberg2020classifiers}
\bibfield{author}{\bibinfo{person}{Jon Kleinberg} {and} \bibinfo{person}{Manish
  Raghavan}.} \bibinfo{year}{2020}\natexlab{}.
\newblock \showarticletitle{How Do Classifiers Induce Agents to Invest Effort
  Strategically?}
\newblock \bibinfo{journal}{\emph{ACM Transactions on Economics and Computation
  (TEAC)}} \bibinfo{volume}{8}, \bibinfo{number}{4} (\bibinfo{year}{2020}),
  \bibinfo{pages}{1--23}.
\newblock


\bibitem[\protect\citeauthoryear{Kovalev and Utkin}{Kovalev and Utkin}{2020}]%
        {kovalev2020counterfactual}
\bibfield{author}{\bibinfo{person}{Maxim~S Kovalev} {and}
  \bibinfo{person}{Lev~V Utkin}.} \bibinfo{year}{2020}\natexlab{}.
\newblock \showarticletitle{Counterfactual explanation of machine learning
  survival models}.
\newblock \bibinfo{journal}{\emph{arXiv preprint arXiv:2006.16793}}
  (\bibinfo{year}{2020}).
\newblock


\bibitem[\protect\citeauthoryear{Lage, Chen, He, Narayanan, Kim, Gershman, and
  Doshi-Velez}{Lage et~al\mbox{.}}{2019}]%
        {lage2019evaluation}
\bibfield{author}{\bibinfo{person}{Isaac Lage}, \bibinfo{person}{Emily Chen},
  \bibinfo{person}{Jeffrey He}, \bibinfo{person}{Menaka Narayanan},
  \bibinfo{person}{Been Kim}, \bibinfo{person}{Sam Gershman}, {and}
  \bibinfo{person}{Finale Doshi-Velez}.} \bibinfo{year}{2019}\natexlab{}.
\newblock \showarticletitle{An evaluation of the human-interpretability of
  explanation}.
\newblock \bibinfo{journal}{\emph{arXiv preprint arXiv:1902.00006}}
  (\bibinfo{year}{2019}).
\newblock


\bibitem[\protect\citeauthoryear{Lakkaraju and Bastani}{Lakkaraju and
  Bastani}{2020}]%
        {lakkaraju2020fool}
\bibfield{author}{\bibinfo{person}{Himabindu Lakkaraju} {and}
  \bibinfo{person}{Osbert Bastani}.} \bibinfo{year}{2020}\natexlab{}.
\newblock \showarticletitle{"How do I fool you?" Manipulating User Trust via
  Misleading Black Box Explanations}. In \bibinfo{booktitle}{\emph{Proceedings
  of the AAAI/ACM Conference on AI, Ethics, and Society}}.
  \bibinfo{pages}{79--85}.
\newblock


\bibitem[\protect\citeauthoryear{Larson, Mattu, Kirchner, and Angwin}{Larson
  et~al\mbox{.}}{2016}]%
        {propublica_compas}
\bibfield{author}{\bibinfo{person}{Jeff Larson}, \bibinfo{person}{Surya Mattu},
  \bibinfo{person}{Lauren Kirchner}, {and} \bibinfo{person}{Julia Angwin}.}
  \bibinfo{year}{2016}\natexlab{}.
\newblock
  \bibinfo{howpublished}{\href{https://github.com/propublica/compas-analysis}{https://github.com/propublica/compas-analysis}}.
\newblock


\bibitem[\protect\citeauthoryear{Lash, Lin, Street, Robinson, and Ohlmann}{Lash
  et~al\mbox{.}}{2017b}]%
        {lash2017generalized}
\bibfield{author}{\bibinfo{person}{Michael~T Lash}, \bibinfo{person}{Qihang
  Lin}, \bibinfo{person}{Nick Street}, \bibinfo{person}{Jennifer~G Robinson},
  {and} \bibinfo{person}{Jeffrey Ohlmann}.} \bibinfo{year}{2017}\natexlab{b}.
\newblock \showarticletitle{Generalized inverse classification}. In
  \bibinfo{booktitle}{\emph{Proceedings of the 2017 SIAM International
  Conference on Data Mining}}. SIAM, \bibinfo{pages}{162--170}.
\newblock


\bibitem[\protect\citeauthoryear{Lash, Lin, and Street}{Lash
  et~al\mbox{.}}{2018}]%
        {lash2018prophit}
\bibfield{author}{\bibinfo{person}{Michael~T Lash}, \bibinfo{person}{Qihang
  Lin}, {and} \bibinfo{person}{W~Nick Street}.}
  \bibinfo{year}{2018}\natexlab{}.
\newblock \showarticletitle{Prophit: Causal inverse classification for multiple
  continuously valued treatment policies}.
\newblock \bibinfo{journal}{\emph{arXiv preprint arXiv:1802.04918}}
  (\bibinfo{year}{2018}).
\newblock


\bibitem[\protect\citeauthoryear{Lash, Lin, Street, and Robinson}{Lash
  et~al\mbox{.}}{2017a}]%
        {lash2017budget}
\bibfield{author}{\bibinfo{person}{Michael~T Lash}, \bibinfo{person}{Qihang
  Lin}, \bibinfo{person}{W~Nick Street}, {and} \bibinfo{person}{Jennifer~G
  Robinson}.} \bibinfo{year}{2017}\natexlab{a}.
\newblock \showarticletitle{A budget-constrained inverse classification
  framework for smooth classifiers}. In \bibinfo{booktitle}{\emph{2017 IEEE
  International Conference on Data Mining Workshops (ICDMW)}}. IEEE,
  \bibinfo{pages}{1184--1193}.
\newblock


\bibitem[\protect\citeauthoryear{Laugel, Lesot, Marsala, and Detyniecki}{Laugel
  et~al\mbox{.}}{2019a}]%
        {laugel2019issues}
\bibfield{author}{\bibinfo{person}{Thibault Laugel},
  \bibinfo{person}{Marie-Jeanne Lesot}, \bibinfo{person}{Christophe Marsala},
  {and} \bibinfo{person}{Marcin Detyniecki}.} \bibinfo{year}{2019}\natexlab{a}.
\newblock \showarticletitle{Issues with post-hoc counterfactual explanations: a
  discussion}.
\newblock \bibinfo{journal}{\emph{arXiv preprint arXiv:1906.04774}}
  (\bibinfo{year}{2019}).
\newblock


\bibitem[\protect\citeauthoryear{Laugel, Lesot, Marsala, Renard, and
  Detyniecki}{Laugel et~al\mbox{.}}{2017}]%
        {laugel2017inverse}
\bibfield{author}{\bibinfo{person}{Thibault Laugel},
  \bibinfo{person}{Marie-Jeanne Lesot}, \bibinfo{person}{Christophe Marsala},
  \bibinfo{person}{Xavier Renard}, {and} \bibinfo{person}{Marcin Detyniecki}.}
  \bibinfo{year}{2017}\natexlab{}.
\newblock \showarticletitle{Inverse Classification for Comparison-based
  Interpretability in Machine Learning}.
\newblock \bibinfo{journal}{\emph{arXiv preprint arXiv:1712.08443}}
  (\bibinfo{year}{2017}).
\newblock


\bibitem[\protect\citeauthoryear{Laugel, Lesot, Marsala, Renard, and
  Detyniecki}{Laugel et~al\mbox{.}}{2019b}]%
        {laugel2019dangers}
\bibfield{author}{\bibinfo{person}{Thibault Laugel},
  \bibinfo{person}{Marie-Jeanne Lesot}, \bibinfo{person}{Christophe Marsala},
  \bibinfo{person}{Xavier Renard}, {and} \bibinfo{person}{Marcin Detyniecki}.}
  \bibinfo{year}{2019}\natexlab{b}.
\newblock \showarticletitle{The dangers of post-hoc interpretability:
  Unjustified counterfactual explanations}.
\newblock \bibinfo{journal}{\emph{arXiv preprint arXiv:1907.09294}}
  (\bibinfo{year}{2019}).
\newblock


\bibitem[\protect\citeauthoryear{Lawler and Wood}{Lawler and Wood}{1966}]%
        {lawler1966branch}
\bibfield{author}{\bibinfo{person}{Eugene~L Lawler} {and}
  \bibinfo{person}{David~E Wood}.} \bibinfo{year}{1966}\natexlab{}.
\newblock \showarticletitle{Branch-and-bound methods: A survey}.
\newblock \bibinfo{journal}{\emph{Operations research}} \bibinfo{volume}{14},
  \bibinfo{number}{4} (\bibinfo{year}{1966}), \bibinfo{pages}{699--719}.
\newblock


\bibitem[\protect\citeauthoryear{Lewis}{Lewis}{1973}]%
        {lewis1973counterfactuals}
\bibfield{author}{\bibinfo{person}{David~K Lewis}.}
  \bibinfo{year}{1973}\natexlab{}.
\newblock \bibinfo{booktitle}{\emph{Counterfactuals}}.
\newblock \bibinfo{publisher}{Harvard University Press}.
\newblock


\bibitem[\protect\citeauthoryear{Lewis}{Lewis}{1986}]%
        {lewis1986causal}
\bibfield{author}{\bibinfo{person}{David~K Lewis}.}
  \bibinfo{year}{1986}\natexlab{}.
\newblock \showarticletitle{Causal explanation}.
\newblock  (\bibinfo{year}{1986}).
\newblock


\bibitem[\protect\citeauthoryear{Lipton}{Lipton}{1990}]%
        {lipton1990contrastive}
\bibfield{author}{\bibinfo{person}{Peter Lipton}.}
  \bibinfo{year}{1990}\natexlab{}.
\newblock \showarticletitle{Contrastive explanation}.
\newblock  (\bibinfo{year}{1990}).
\newblock


\bibitem[\protect\citeauthoryear{Lipton}{Lipton}{2018}]%
        {lipton2018mythos}
\bibfield{author}{\bibinfo{person}{Zachary~C Lipton}.}
  \bibinfo{year}{2018}\natexlab{}.
\newblock \showarticletitle{The mythos of model interpretability}.
\newblock \bibinfo{journal}{\emph{Queue}} \bibinfo{volume}{16},
  \bibinfo{number}{3} (\bibinfo{year}{2018}), \bibinfo{pages}{31--57}.
\newblock


\bibitem[\protect\citeauthoryear{Liu, Wilson, Haghtalab, Kalai, Borgs, and
  Chayes}{Liu et~al\mbox{.}}{2020}]%
        {liu2020disparate}
\bibfield{author}{\bibinfo{person}{Lydia~T Liu}, \bibinfo{person}{Ashia
  Wilson}, \bibinfo{person}{Nika Haghtalab}, \bibinfo{person}{Adam~Tauman
  Kalai}, \bibinfo{person}{Christian Borgs}, {and} \bibinfo{person}{Jennifer
  Chayes}.} \bibinfo{year}{2020}\natexlab{}.
\newblock \showarticletitle{The disparate equilibria of algorithmic decision
  making when individuals invest rationally}. In
  \bibinfo{booktitle}{\emph{Proceedings of the 2020 Conference on Fairness,
  Accountability, and Transparency}}. \bibinfo{pages}{381--391}.
\newblock


\bibitem[\protect\citeauthoryear{Lowd and Meek}{Lowd and Meek}{2005}]%
        {lowd2005adversarial}
\bibfield{author}{\bibinfo{person}{Daniel Lowd} {and}
  \bibinfo{person}{Christopher Meek}.} \bibinfo{year}{2005}\natexlab{}.
\newblock \showarticletitle{Adversarial learning}. In
  \bibinfo{booktitle}{\emph{Proceedings of the eleventh ACM SIGKDD
  international conference on Knowledge discovery in data mining}}. ACM,
  \bibinfo{pages}{641--647}.
\newblock


\bibitem[\protect\citeauthoryear{Lucic, Haned, and de~Rijke}{Lucic
  et~al\mbox{.}}{2020}]%
        {lucic2020does}
\bibfield{author}{\bibinfo{person}{Ana Lucic}, \bibinfo{person}{Hinda Haned},
  {and} \bibinfo{person}{Maarten de Rijke}.} \bibinfo{year}{2020}\natexlab{}.
\newblock \showarticletitle{Why does my model fail? contrastive local
  explanations for retail forecasting}. In
  \bibinfo{booktitle}{\emph{Proceedings of the 2020 Conference on Fairness,
  Accountability, and Transparency}}. \bibinfo{pages}{90--98}.
\newblock


\bibitem[\protect\citeauthoryear{Lucic, Oosterhuis, Haned, and de~Rijke}{Lucic
  et~al\mbox{.}}{2019}]%
        {lucic2019actionable}
\bibfield{author}{\bibinfo{person}{Ana Lucic}, \bibinfo{person}{Harrie
  Oosterhuis}, \bibinfo{person}{Hinda Haned}, {and} \bibinfo{person}{Maarten de
  Rijke}.} \bibinfo{year}{2019}\natexlab{}.
\newblock \showarticletitle{Actionable Interpretability through Optimizable
  Counterfactual Explanations for Tree Ensembles}.
\newblock \bibinfo{journal}{\emph{arXiv preprint arXiv:1911.12199}}
  (\bibinfo{year}{2019}).
\newblock


\bibitem[\protect\citeauthoryear{Madumal, Miller, Sonenberg, and
  Vetere}{Madumal et~al\mbox{.}}{2019}]%
        {madumal2019explainable}
\bibfield{author}{\bibinfo{person}{Prashan Madumal}, \bibinfo{person}{Tim
  Miller}, \bibinfo{person}{Liz Sonenberg}, {and} \bibinfo{person}{Frank
  Vetere}.} \bibinfo{year}{2019}\natexlab{}.
\newblock \showarticletitle{Explainable reinforcement learning through a causal
  lens}.
\newblock \bibinfo{journal}{\emph{arXiv preprint arXiv:1905.10958}}
  (\bibinfo{year}{2019}).
\newblock


\bibitem[\protect\citeauthoryear{Madumal, Miller, Sonenberg, and
  Vetere}{Madumal et~al\mbox{.}}{2020}]%
        {madumal2020distal}
\bibfield{author}{\bibinfo{person}{Prashan Madumal}, \bibinfo{person}{Tim
  Miller}, \bibinfo{person}{Liz Sonenberg}, {and} \bibinfo{person}{Frank
  Vetere}.} \bibinfo{year}{2020}\natexlab{}.
\newblock \showarticletitle{Distal Explanations for Explainable Reinforcement
  Learning Agents}.
\newblock \bibinfo{journal}{\emph{arXiv preprint arXiv:2001.10284}}
  (\bibinfo{year}{2020}).
\newblock


\bibitem[\protect\citeauthoryear{Mahajan, Tan, and Sharma}{Mahajan
  et~al\mbox{.}}{2019}]%
        {mahajan2019preserving}
\bibfield{author}{\bibinfo{person}{Divyat Mahajan}, \bibinfo{person}{Chenhao
  Tan}, {and} \bibinfo{person}{Amit Sharma}.} \bibinfo{year}{2019}\natexlab{}.
\newblock \showarticletitle{Preserving Causal Constraints in Counterfactual
  Explanations for Machine Learning Classifiers}.
\newblock \bibinfo{journal}{\emph{arXiv preprint arXiv:1912.03277}}
  (\bibinfo{year}{2019}).
\newblock


\bibitem[\protect\citeauthoryear{Mannino and Koushik}{Mannino and
  Koushik}{2000}]%
        {mannino2000cost}
\bibfield{author}{\bibinfo{person}{Michael~V Mannino} {and}
  \bibinfo{person}{Murlidhar~V Koushik}.} \bibinfo{year}{2000}\natexlab{}.
\newblock \showarticletitle{The cost-minimizing inverse classification problem:
  a genetic algorithm approach}.
\newblock \bibinfo{journal}{\emph{Decision Support Systems}}
  \bibinfo{volume}{29}, \bibinfo{number}{3} (\bibinfo{year}{2000}),
  \bibinfo{pages}{283--300}.
\newblock


\bibitem[\protect\citeauthoryear{Martens and Provost}{Martens and
  Provost}{2014}]%
        {martens2014explaining}
\bibfield{author}{\bibinfo{person}{David Martens} {and} \bibinfo{person}{Foster
  Provost}.} \bibinfo{year}{2014}\natexlab{}.
\newblock \showarticletitle{Explaining data-driven document classifications}.
\newblock \bibinfo{journal}{\emph{Mis Quarterly}} \bibinfo{volume}{38},
  \bibinfo{number}{1} (\bibinfo{year}{2014}), \bibinfo{pages}{73--100}.
\newblock


\bibitem[\protect\citeauthoryear{McGarry}{McGarry}{[n.d.]}]%
        {mcgarry2005survey}
\bibfield{author}{\bibinfo{person}{Kenneth McGarry}.}
  \bibinfo{year}{[n.d.]}\natexlab{}.
\newblock \showarticletitle{A survey of interestingness measures for knowledge
  discovery}.
\newblock  (\bibinfo{year}{[n.\,d.]}).
\newblock


\bibitem[\protect\citeauthoryear{McGill and Klein}{McGill and Klein}{1993}]%
        {mcgill1993contrastive}
\bibfield{author}{\bibinfo{person}{Ann~L McGill} {and} \bibinfo{person}{Jill~G
  Klein}.} \bibinfo{year}{1993}\natexlab{}.
\newblock \showarticletitle{Contrastive and counterfactual reasoning in causal
  judgment.}
\newblock \bibinfo{journal}{\emph{Journal of Personality and Social
  Psychology}} \bibinfo{volume}{64}, \bibinfo{number}{6}
  (\bibinfo{year}{1993}), \bibinfo{pages}{897}.
\newblock


\bibitem[\protect\citeauthoryear{Melis and Jaakkola}{Melis and
  Jaakkola}{2018}]%
        {melis2018towards}
\bibfield{author}{\bibinfo{person}{David~Alvarez Melis} {and}
  \bibinfo{person}{Tommi Jaakkola}.} \bibinfo{year}{2018}\natexlab{}.
\newblock \showarticletitle{Towards robust interpretability with
  self-explaining neural networks}. In \bibinfo{booktitle}{\emph{Advances in
  Neural Information Processing Systems}}. \bibinfo{pages}{7775--7784}.
\newblock


\bibitem[\protect\citeauthoryear{Miller}{Miller}{1956}]%
        {miller1956magical}
\bibfield{author}{\bibinfo{person}{George~A Miller}.}
  \bibinfo{year}{1956}\natexlab{}.
\newblock \showarticletitle{The magical number seven, plus or minus two: Some
  limits on our capacity for processing information.}
\newblock \bibinfo{journal}{\emph{Psychological review}} \bibinfo{volume}{63},
  \bibinfo{number}{2} (\bibinfo{year}{1956}), \bibinfo{pages}{81}.
\newblock


\bibitem[\protect\citeauthoryear{Miller, Milli, and Hardt}{Miller
  et~al\mbox{.}}{2020}]%
        {miller2020strategic}
\bibfield{author}{\bibinfo{person}{John Miller}, \bibinfo{person}{Smitha
  Milli}, {and} \bibinfo{person}{Moritz Hardt}.}
  \bibinfo{year}{2020}\natexlab{}.
\newblock \showarticletitle{Strategic classification is causal modeling in
  disguise}. In \bibinfo{booktitle}{\emph{International Conference on Machine
  Learning}}. PMLR, \bibinfo{pages}{6917--6926}.
\newblock


\bibitem[\protect\citeauthoryear{Miller}{Miller}{2018}]%
        {miller2018contrastive}
\bibfield{author}{\bibinfo{person}{Tim Miller}.}
  \bibinfo{year}{2018}\natexlab{}.
\newblock \showarticletitle{Contrastive explanation: A structural-model
  approach}.
\newblock \bibinfo{journal}{\emph{arXiv preprint arXiv:1811.03163}}
  (\bibinfo{year}{2018}).
\newblock


\bibitem[\protect\citeauthoryear{Miller}{Miller}{2019}]%
        {miller2019explanation}
\bibfield{author}{\bibinfo{person}{Tim Miller}.}
  \bibinfo{year}{2019}\natexlab{}.
\newblock \showarticletitle{Explanation in artificial intelligence: Insights
  from the social sciences}.
\newblock \bibinfo{journal}{\emph{Artificial Intelligence}}
  \bibinfo{volume}{267} (\bibinfo{year}{2019}), \bibinfo{pages}{1--38}.
\newblock


\bibitem[\protect\citeauthoryear{Milli, Miller, Dragan, and Hardt}{Milli
  et~al\mbox{.}}{2019a}]%
        {milli2019social}
\bibfield{author}{\bibinfo{person}{Smitha Milli}, \bibinfo{person}{John
  Miller}, \bibinfo{person}{Anca~D Dragan}, {and} \bibinfo{person}{Moritz
  Hardt}.} \bibinfo{year}{2019}\natexlab{a}.
\newblock \showarticletitle{The social cost of strategic classification}. In
  \bibinfo{booktitle}{\emph{Proceedings of the Conference on Fairness,
  Accountability, and Transparency}}. \bibinfo{pages}{230--239}.
\newblock


\bibitem[\protect\citeauthoryear{Milli, Schmidt, Dragan, and Hardt}{Milli
  et~al\mbox{.}}{2019b}]%
        {milli2019model}
\bibfield{author}{\bibinfo{person}{Smitha Milli}, \bibinfo{person}{Ludwig
  Schmidt}, \bibinfo{person}{Anca~D Dragan}, {and} \bibinfo{person}{Moritz
  Hardt}.} \bibinfo{year}{2019}\natexlab{b}.
\newblock \showarticletitle{Model reconstruction from model explanations}. In
  \bibinfo{booktitle}{\emph{Proceedings of the Conference on Fairness,
  Accountability, and Transparency}}. \bibinfo{pages}{1--9}.
\newblock


\bibitem[\protect\citeauthoryear{Mittelstadt, Russell, and Wachter}{Mittelstadt
  et~al\mbox{.}}{2019}]%
        {mittelstadt2019explaining}
\bibfield{author}{\bibinfo{person}{Brent Mittelstadt}, \bibinfo{person}{Chris
  Russell}, {and} \bibinfo{person}{Sandra Wachter}.}
  \bibinfo{year}{2019}\natexlab{}.
\newblock \showarticletitle{Explaining explanations in AI}. In
  \bibinfo{booktitle}{\emph{Proceedings of the conference on fairness,
  accountability, and transparency}}. \bibinfo{pages}{279--288}.
\newblock


\bibitem[\protect\citeauthoryear{Mohammadi, Karimi, Barthe, and
  Valera}{Mohammadi et~al\mbox{.}}{2020}]%
        {mohammadi2020scalable}
\bibfield{author}{\bibinfo{person}{Kiarash Mohammadi},
  \bibinfo{person}{Amir-Hossein Karimi}, \bibinfo{person}{Gilles Barthe}, {and}
  \bibinfo{person}{Isabel Valera}.} \bibinfo{year}{2020}\natexlab{}.
\newblock \showarticletitle{Scaling Guarantees for Nearest Counterfactual
  Explanations}.
\newblock \bibinfo{journal}{\emph{arXiv preprint arXiv:2010.04965}}
  (\bibinfo{year}{2020}).
\newblock


\bibitem[\protect\citeauthoryear{Montavon, Samek, and M{\"u}ller}{Montavon
  et~al\mbox{.}}{2018}]%
        {montavon2018methods}
\bibfield{author}{\bibinfo{person}{Gr{\'e}goire Montavon},
  \bibinfo{person}{Wojciech Samek}, {and} \bibinfo{person}{Klaus-Robert
  M{\"u}ller}.} \bibinfo{year}{2018}\natexlab{}.
\newblock \showarticletitle{Methods for interpreting and understanding deep
  neural networks}.
\newblock \bibinfo{journal}{\emph{Digital Signal Processing}}
  \bibinfo{volume}{73} (\bibinfo{year}{2018}), \bibinfo{pages}{1--15}.
\newblock


\bibitem[\protect\citeauthoryear{Moosavi-Dezfooli, Fawzi, Fawzi, and
  Frossard}{Moosavi-Dezfooli et~al\mbox{.}}{2017}]%
        {moosavi2017universal}
\bibfield{author}{\bibinfo{person}{Seyed-Mohsen Moosavi-Dezfooli},
  \bibinfo{person}{Alhussein Fawzi}, \bibinfo{person}{Omar Fawzi}, {and}
  \bibinfo{person}{Pascal Frossard}.} \bibinfo{year}{2017}\natexlab{}.
\newblock \showarticletitle{Universal adversarial perturbations}. In
  \bibinfo{booktitle}{\emph{Proceedings of the IEEE conference on computer
  vision and pattern recognition}}. \bibinfo{pages}{1765--1773}.
\newblock


\bibitem[\protect\citeauthoryear{Moosavi-Dezfooli, Fawzi, and
  Frossard}{Moosavi-Dezfooli et~al\mbox{.}}{2016}]%
        {moosavi2016deepfool}
\bibfield{author}{\bibinfo{person}{Seyed-Mohsen Moosavi-Dezfooli},
  \bibinfo{person}{Alhussein Fawzi}, {and} \bibinfo{person}{Pascal Frossard}.}
  \bibinfo{year}{2016}\natexlab{}.
\newblock \showarticletitle{Deepfool: a simple and accurate method to fool deep
  neural networks}. In \bibinfo{booktitle}{\emph{Proceedings of the IEEE
  conference on computer vision and pattern recognition}}.
  \bibinfo{pages}{2574--2582}.
\newblock


\bibitem[\protect\citeauthoryear{Moraffah, Karami, Guo, Raglin, and
  Liu}{Moraffah et~al\mbox{.}}{2020}]%
        {moraffah2020causal}
\bibfield{author}{\bibinfo{person}{Raha Moraffah}, \bibinfo{person}{Mansooreh
  Karami}, \bibinfo{person}{Ruocheng Guo}, \bibinfo{person}{Adrienne Raglin},
  {and} \bibinfo{person}{Huan Liu}.} \bibinfo{year}{2020}\natexlab{}.
\newblock \showarticletitle{Causal Interpretability for Machine
  Learning-Problems, Methods and Evaluation}.
\newblock \bibinfo{journal}{\emph{ACM SIGKDD Explorations Newsletter}}
  \bibinfo{volume}{22}, \bibinfo{number}{1} (\bibinfo{year}{2020}),
  \bibinfo{pages}{18--33}.
\newblock


\bibitem[\protect\citeauthoryear{Mothilal, Sharma, and Tan}{Mothilal
  et~al\mbox{.}}{2019}]%
        {mothilal2019explaining}
\bibfield{author}{\bibinfo{person}{Ramaravind~Kommiya Mothilal},
  \bibinfo{person}{Amit Sharma}, {and} \bibinfo{person}{Chenhao Tan}.}
  \bibinfo{year}{2019}\natexlab{}.
\newblock \showarticletitle{{DiCE}: Explaining Machine Learning Classifiers
  through Diverse Counterfactual Explanations}.
\newblock \bibinfo{journal}{\emph{arXiv preprint arXiv:1905.07697}}
  (\bibinfo{year}{2019}).
\newblock


\bibitem[\protect\citeauthoryear{Mukerjee, Biswas, Deb, and Mathur}{Mukerjee
  et~al\mbox{.}}{2002}]%
        {mukerjee2002multi}
\bibfield{author}{\bibinfo{person}{Amitabha Mukerjee}, \bibinfo{person}{Rita
  Biswas}, \bibinfo{person}{Kalyanmoy Deb}, {and} \bibinfo{person}{Amrit~P
  Mathur}.} \bibinfo{year}{2002}\natexlab{}.
\newblock \showarticletitle{Multi--objective evolutionary algorithms for the
  risk--return trade--off in bank loan management}.
\newblock \bibinfo{journal}{\emph{International Transactions in operational
  research}} \bibinfo{volume}{9}, \bibinfo{number}{5} (\bibinfo{year}{2002}),
  \bibinfo{pages}{583--597}.
\newblock


\bibitem[\protect\citeauthoryear{Nabi and Shpitser}{Nabi and Shpitser}{2018}]%
        {nabi2018fair}
\bibfield{author}{\bibinfo{person}{Razieh Nabi} {and} \bibinfo{person}{Ilya
  Shpitser}.} \bibinfo{year}{2018}\natexlab{}.
\newblock \showarticletitle{Fair inference on outcomes}. In
  \bibinfo{booktitle}{\emph{Proceedings of the... AAAI Conference on Artificial
  Intelligence. AAAI Conference on Artificial Intelligence}},
  Vol.~\bibinfo{volume}{2018}. NIH Public Access, \bibinfo{pages}{1931}.
\newblock


\bibitem[\protect\citeauthoryear{Nazabal, Olmos, Ghahramani, and
  Valera}{Nazabal et~al\mbox{.}}{2020}]%
        {nazabal2020handling}
\bibfield{author}{\bibinfo{person}{Alfredo Nazabal}, \bibinfo{person}{Pablo~M
  Olmos}, \bibinfo{person}{Zoubin Ghahramani}, {and} \bibinfo{person}{Isabel
  Valera}.} \bibinfo{year}{2020}\natexlab{}.
\newblock \showarticletitle{Handling incomplete heterogeneous data using vaes}.
\newblock \bibinfo{journal}{\emph{Pattern Recognition}} (\bibinfo{year}{2020}),
  \bibinfo{pages}{107501}.
\newblock


\bibitem[\protect\citeauthoryear{Nguyen, Yosinski, and Clune}{Nguyen
  et~al\mbox{.}}{2015}]%
        {nguyen2015deep}
\bibfield{author}{\bibinfo{person}{Anh Nguyen}, \bibinfo{person}{Jason
  Yosinski}, {and} \bibinfo{person}{Jeff Clune}.}
  \bibinfo{year}{2015}\natexlab{}.
\newblock \showarticletitle{Deep neural networks are easily fooled: High
  confidence predictions for unrecognizable images}. In
  \bibinfo{booktitle}{\emph{Proceedings of the IEEE conference on computer
  vision and pattern recognition}}. \bibinfo{pages}{427--436}.
\newblock


\bibitem[\protect\citeauthoryear{Nocedal and Wright}{Nocedal and
  Wright}{2006}]%
        {nocedal2006numerical}
\bibfield{author}{\bibinfo{person}{Jorge Nocedal} {and}
  \bibinfo{person}{Stephen Wright}.} \bibinfo{year}{2006}\natexlab{}.
\newblock \bibinfo{booktitle}{\emph{Numerical optimization}}.
\newblock \bibinfo{publisher}{Springer Science \& Business Media}.
\newblock


\bibitem[\protect\citeauthoryear{OPTIMIZATION}{OPTIMIZATION}{2014}]%
        {optimization2014inc}
\bibfield{author}{\bibinfo{person}{GUROBI OPTIMIZATION}.}
  \bibinfo{year}{2014}\natexlab{}.
\newblock \showarticletitle{INC. Gurobi optimizer reference manual, 2015}.
\newblock \bibinfo{journal}{\emph{URL: http://www. gurobi. com}}
  (\bibinfo{year}{2014}), \bibinfo{pages}{29}.
\newblock


\bibitem[\protect\citeauthoryear{Papernot, McDaniel, Goodfellow, Jha, Celik,
  and Swami}{Papernot et~al\mbox{.}}{2017}]%
        {papernot2017practical}
\bibfield{author}{\bibinfo{person}{Nicolas Papernot}, \bibinfo{person}{Patrick
  McDaniel}, \bibinfo{person}{Ian Goodfellow}, \bibinfo{person}{Somesh Jha},
  \bibinfo{person}{Z~Berkay Celik}, {and} \bibinfo{person}{Ananthram Swami}.}
  \bibinfo{year}{2017}\natexlab{}.
\newblock \showarticletitle{Practical black-box attacks against machine
  learning}. In \bibinfo{booktitle}{\emph{Proceedings of the 2017 ACM on Asia
  conference on computer and communications security}}.
  \bibinfo{pages}{506--519}.
\newblock


\bibitem[\protect\citeauthoryear{Papernot, McDaniel, Jha, Fredrikson, Celik,
  and Swami}{Papernot et~al\mbox{.}}{2016}]%
        {papernot2016limitations}
\bibfield{author}{\bibinfo{person}{Nicolas Papernot}, \bibinfo{person}{Patrick
  McDaniel}, \bibinfo{person}{Somesh Jha}, \bibinfo{person}{Matt Fredrikson},
  \bibinfo{person}{Z~Berkay Celik}, {and} \bibinfo{person}{Ananthram Swami}.}
  \bibinfo{year}{2016}\natexlab{}.
\newblock \showarticletitle{The limitations of deep learning in adversarial
  settings}. In \bibinfo{booktitle}{\emph{2016 IEEE European symposium on
  security and privacy (EuroS\&P)}}. IEEE, \bibinfo{pages}{372--387}.
\newblock


\bibitem[\protect\citeauthoryear{Park and Boyd}{Park and Boyd}{2017}]%
        {park2017general}
\bibfield{author}{\bibinfo{person}{Jaehyun Park} {and} \bibinfo{person}{Stephen
  Boyd}.} \bibinfo{year}{2017}\natexlab{}.
\newblock \showarticletitle{General heuristics for nonconvex quadratically
  constrained quadratic programming}.
\newblock \bibinfo{journal}{\emph{arXiv preprint arXiv:1703.07870}}
  (\bibinfo{year}{2017}).
\newblock


\bibitem[\protect\citeauthoryear{Pawelczyk, Broelemann, and Kasneci}{Pawelczyk
  et~al\mbox{.}}{2020}]%
        {pawelczyk2020counterfactual}
\bibfield{author}{\bibinfo{person}{Martin Pawelczyk}, \bibinfo{person}{Klaus
  Broelemann}, {and} \bibinfo{person}{Gjergji Kasneci}.}
  \bibinfo{year}{2020}\natexlab{}.
\newblock \showarticletitle{On Counterfactual Explanations under Predictive
  Multiplicity}.
\newblock \bibinfo{journal}{\emph{arXiv preprint arXiv:2006.13132}}
  (\bibinfo{year}{2020}).
\newblock


\bibitem[\protect\citeauthoryear{Pawelczyk, Haug, Broelemann, and
  Kasneci}{Pawelczyk et~al\mbox{.}}{2019}]%
        {pawelczyk2019towards}
\bibfield{author}{\bibinfo{person}{Martin Pawelczyk}, \bibinfo{person}{Johannes
  Haug}, \bibinfo{person}{Klaus Broelemann}, {and} \bibinfo{person}{Gjergji
  Kasneci}.} \bibinfo{year}{2019}\natexlab{}.
\newblock \showarticletitle{Towards User Empowerment}.
\newblock \bibinfo{journal}{\emph{arXiv preprint arXiv:1910.09398}}
  (\bibinfo{year}{2019}).
\newblock


\bibitem[\protect\citeauthoryear{Pearl}{Pearl}{2000}]%
        {pearl2000causality}
\bibfield{author}{\bibinfo{person}{Judea Pearl}.}
  \bibinfo{year}{2000}\natexlab{}.
\newblock \bibinfo{booktitle}{\emph{Causality: models, reasoning and
  inference}}. Vol.~\bibinfo{volume}{29}.
\newblock \bibinfo{publisher}{Springer}.
\newblock


\bibitem[\protect\citeauthoryear{Pearl}{Pearl}{2010}]%
        {pearl2010foundations}
\bibfield{author}{\bibinfo{person}{Judea Pearl}.}
  \bibinfo{year}{2010}\natexlab{}.
\newblock \showarticletitle{The foundations of causal inference}.
\newblock \bibinfo{journal}{\emph{Sociological Methodology}}
  \bibinfo{volume}{40}, \bibinfo{number}{1} (\bibinfo{year}{2010}),
  \bibinfo{pages}{75--149}.
\newblock


\bibitem[\protect\citeauthoryear{Pearl, Glymour, and Jewell}{Pearl
  et~al\mbox{.}}{2016}]%
        {pearl2016causal}
\bibfield{author}{\bibinfo{person}{Judea Pearl}, \bibinfo{person}{Madelyn
  Glymour}, {and} \bibinfo{person}{Nicholas~P Jewell}.}
  \bibinfo{year}{2016}\natexlab{}.
\newblock \bibinfo{booktitle}{\emph{Causal inference in statistics: A primer}}.
\newblock \bibinfo{publisher}{John Wiley \& Sons}.
\newblock


\bibitem[\protect\citeauthoryear{Pennington, Socher, and Manning}{Pennington
  et~al\mbox{.}}{2014}]%
        {pennington2014glove}
\bibfield{author}{\bibinfo{person}{Jeffrey Pennington},
  \bibinfo{person}{Richard Socher}, {and} \bibinfo{person}{Christopher~D
  Manning}.} \bibinfo{year}{2014}\natexlab{}.
\newblock \showarticletitle{Glove: Global vectors for word representation}. In
  \bibinfo{booktitle}{\emph{Proceedings of the 2014 conference on empirical
  methods in natural language processing (EMNLP)}}.
  \bibinfo{pages}{1532--1543}.
\newblock


\bibitem[\protect\citeauthoryear{Peters, Janzing, and Sch{\"o}lkopf}{Peters
  et~al\mbox{.}}{2017}]%
        {peters2017elements}
\bibfield{author}{\bibinfo{person}{Jonas Peters}, \bibinfo{person}{Dominik
  Janzing}, {and} \bibinfo{person}{Bernhard Sch{\"o}lkopf}.}
  \bibinfo{year}{2017}\natexlab{}.
\newblock \bibinfo{booktitle}{\emph{Elements of causal inference}}.
\newblock \bibinfo{publisher}{The MIT Press}.
\newblock


\bibitem[\protect\citeauthoryear{Poyiadzi, Sokol, Santos-Rodriguez, De~Bie, and
  Flach}{Poyiadzi et~al\mbox{.}}{2019}]%
        {poyiadzi2019face}
\bibfield{author}{\bibinfo{person}{Rafael Poyiadzi}, \bibinfo{person}{Kacper
  Sokol}, \bibinfo{person}{Raul Santos-Rodriguez}, \bibinfo{person}{Tijl
  De~Bie}, {and} \bibinfo{person}{Peter Flach}.}
  \bibinfo{year}{2019}\natexlab{}.
\newblock \showarticletitle{{FACE}: Feasible and Actionable Counterfactual
  Explanations}.
\newblock \bibinfo{journal}{\emph{arXiv preprint arXiv:1909.09369}}
  (\bibinfo{year}{2019}).
\newblock


\bibitem[\protect\citeauthoryear{Ramakrishnan, Lee, and
  Albargouthi}{Ramakrishnan et~al\mbox{.}}{2019}]%
        {ramakrishnan2019synthesizing}
\bibfield{author}{\bibinfo{person}{Goutham Ramakrishnan},
  \bibinfo{person}{Yun~Chan Lee}, {and} \bibinfo{person}{Aws Albargouthi}.}
  \bibinfo{year}{2019}\natexlab{}.
\newblock \showarticletitle{Synthesizing Action Sequences for Modifying Model
  Decisions}.
\newblock \bibinfo{journal}{\emph{arXiv preprint arXiv:1910.00057}}
  (\bibinfo{year}{2019}).
\newblock


\bibitem[\protect\citeauthoryear{Ramon, Martens, Provost, and Evgeniou}{Ramon
  et~al\mbox{.}}{2019}]%
        {ramon2019counterfactual}
\bibfield{author}{\bibinfo{person}{Yanou Ramon}, \bibinfo{person}{David
  Martens}, \bibinfo{person}{Foster Provost}, {and} \bibinfo{person}{Theodoros
  Evgeniou}.} \bibinfo{year}{2019}\natexlab{}.
\newblock \showarticletitle{Counterfactual explanation algorithms for
  behavioral and textual data}.
\newblock \bibinfo{journal}{\emph{arXiv preprint arXiv:1912.01819}}
  (\bibinfo{year}{2019}).
\newblock


\bibitem[\protect\citeauthoryear{Rathi}{Rathi}{2019}]%
        {rathi2019generating}
\bibfield{author}{\bibinfo{person}{Shubham Rathi}.}
  \bibinfo{year}{2019}\natexlab{}.
\newblock \showarticletitle{Generating counterfactual and contrastive
  explanations using SHAP}.
\newblock \bibinfo{journal}{\emph{arXiv preprint arXiv:1906.09293}}
  (\bibinfo{year}{2019}).
\newblock


\bibitem[\protect\citeauthoryear{Rawal, Kamar, and Lakkaraju}{Rawal
  et~al\mbox{.}}{2020}]%
        {rawal2020can}
\bibfield{author}{\bibinfo{person}{Kaivalya Rawal}, \bibinfo{person}{Ece
  Kamar}, {and} \bibinfo{person}{Himabindu Lakkaraju}.}
  \bibinfo{year}{2020}\natexlab{}.
\newblock \showarticletitle{Can I Still Trust You?: Understanding the Impact of
  Distribution Shifts on Algorithmic Recourses}.
\newblock \bibinfo{journal}{\emph{arXiv preprint arXiv:2012.11788}}
  (\bibinfo{year}{2020}).
\newblock


\bibitem[\protect\citeauthoryear{Rawal and Lakkaraju}{Rawal and
  Lakkaraju}{2020}]%
        {rawal2020interpretable}
\bibfield{author}{\bibinfo{person}{Kaivalya Rawal} {and}
  \bibinfo{person}{Himabindu Lakkaraju}.} \bibinfo{year}{2020}\natexlab{}.
\newblock \showarticletitle{Beyond Individualized Recourse: Interpretable and
  Interactive Summaries of Actionable Recourses}.
\newblock \bibinfo{journal}{\emph{Advances in Neural Information Processing
  Systems}}  \bibinfo{volume}{33} (\bibinfo{year}{2020}).
\newblock


\bibitem[\protect\citeauthoryear{Reith, Schneider, and Tkachenko}{Reith
  et~al\mbox{.}}{2019}]%
        {reith2019efficiently}
\bibfield{author}{\bibinfo{person}{Robert~Nikolai Reith},
  \bibinfo{person}{Thomas Schneider}, {and} \bibinfo{person}{Oleksandr
  Tkachenko}.} \bibinfo{year}{2019}\natexlab{}.
\newblock \showarticletitle{Efficiently stealing your machine learning models}.
  In \bibinfo{booktitle}{\emph{Proceedings of the 18th ACM Workshop on Privacy
  in the Electronic Society}}. \bibinfo{pages}{198--210}.
\newblock


\bibitem[\protect\citeauthoryear{Ribeiro, Singh, and Guestrin}{Ribeiro
  et~al\mbox{.}}{2016}]%
        {ribeiro2016should}
\bibfield{author}{\bibinfo{person}{Marco~Tulio Ribeiro},
  \bibinfo{person}{Sameer Singh}, {and} \bibinfo{person}{Carlos Guestrin}.}
  \bibinfo{year}{2016}\natexlab{}.
\newblock \showarticletitle{" Why should I trust you?" Explaining the
  predictions of any classifier}. In \bibinfo{booktitle}{\emph{Proceedings of
  the 22nd ACM SIGKDD international conference on knowledge discovery and data
  mining}}. \bibinfo{pages}{1135--1144}.
\newblock


\bibitem[\protect\citeauthoryear{Robeer}{Robeer}{2018}]%
        {robeer2018contrastive}
\bibfield{author}{\bibinfo{person}{Marcel~Jurriaan Robeer}.}
  \bibinfo{year}{2018}\natexlab{}.
\newblock \emph{\bibinfo{title}{Contrastive explanation for machine learning}}.
\newblock \bibinfo{thesistype}{Master's\ thesis}.
\newblock


\bibitem[\protect\citeauthoryear{Rosenfeld, Hilgard, Ravindranath, and
  Parkes}{Rosenfeld et~al\mbox{.}}{2020}]%
        {rosenfeld2020predictions}
\bibfield{author}{\bibinfo{person}{Nir Rosenfeld}, \bibinfo{person}{Anna
  Hilgard}, \bibinfo{person}{Sai~Srivatsa Ravindranath}, {and}
  \bibinfo{person}{David~C Parkes}.} \bibinfo{year}{2020}\natexlab{}.
\newblock \showarticletitle{From Predictions to Decisions: Using Lookahead
  Regularization}.
\newblock \bibinfo{journal}{\emph{Advances in Neural Information Processing
  Systems}}  \bibinfo{volume}{33} (\bibinfo{year}{2020}).
\newblock


\bibitem[\protect\citeauthoryear{Ross, Lakkaraju, and Bastani}{Ross
  et~al\mbox{.}}{2020}]%
        {ross2020ensuring}
\bibfield{author}{\bibinfo{person}{Alexis Ross}, \bibinfo{person}{Himabindu
  Lakkaraju}, {and} \bibinfo{person}{Osbert Bastani}.}
  \bibinfo{year}{2020}\natexlab{}.
\newblock \showarticletitle{Ensuring Actionable Recourse via Adversarial
  Training}.
\newblock \bibinfo{journal}{\emph{arXiv preprint arXiv:2011.06146}}
  (\bibinfo{year}{2020}).
\newblock


\bibitem[\protect\citeauthoryear{Ruben}{Ruben}{2015}]%
        {ruben2015explaining}
\bibfield{author}{\bibinfo{person}{David-Hillel Ruben}.}
  \bibinfo{year}{2015}\natexlab{}.
\newblock \bibinfo{booktitle}{\emph{Explaining explanation}}.
\newblock \bibinfo{publisher}{Routledge}.
\newblock


\bibitem[\protect\citeauthoryear{Rudin}{Rudin}{2019}]%
        {rudin2019stop}
\bibfield{author}{\bibinfo{person}{Cynthia Rudin}.}
  \bibinfo{year}{2019}\natexlab{}.
\newblock \showarticletitle{Stop explaining black box machine learning models
  for high stakes decisions and use interpretable models instead}.
\newblock \bibinfo{journal}{\emph{Nature Machine Intelligence}}
  \bibinfo{volume}{1}, \bibinfo{number}{5} (\bibinfo{year}{2019}),
  \bibinfo{pages}{206--215}.
\newblock


\bibitem[\protect\citeauthoryear{Russell}{Russell}{2019}]%
        {russell2019efficient}
\bibfield{author}{\bibinfo{person}{Chris Russell}.}
  \bibinfo{year}{2019}\natexlab{}.
\newblock \showarticletitle{Efficient Search for Diverse Coherent
  Explanations}. In \bibinfo{booktitle}{\emph{Proceedings of the Conference on
  Fairness, Accountability, and Transparency}} \emph{(\bibinfo{series}{FAT*
  '19})}. \bibinfo{publisher}{ACM}, \bibinfo{pages}{20--28}.
\newblock
\showISBNx{978-1-4503-6125-5}
\urldef\tempurl%
\url{https://doi.org/10.1145/3287560.3287569}
\showDOI{\tempurl}


\bibitem[\protect\citeauthoryear{Sch{\"o}lkopf}{Sch{\"o}lkopf}{2019}]%
        {scholkopf2019causality}
\bibfield{author}{\bibinfo{person}{Bernhard Sch{\"o}lkopf}.}
  \bibinfo{year}{2019}\natexlab{}.
\newblock \showarticletitle{Causality for machine learning}.
\newblock \bibinfo{journal}{\emph{arXiv preprint arXiv:1911.10500}}
  (\bibinfo{year}{2019}).
\newblock


\bibitem[\protect\citeauthoryear{Schumann, Foster, Mattei, and
  Dickerson}{Schumann et~al\mbox{.}}{2020}]%
        {schumann2020we}
\bibfield{author}{\bibinfo{person}{Candice Schumann},
  \bibinfo{person}{Jeffrey~S Foster}, \bibinfo{person}{Nicholas Mattei}, {and}
  \bibinfo{person}{John~P Dickerson}.} \bibinfo{year}{2020}\natexlab{}.
\newblock \showarticletitle{We Need Fairness and Explainability in Algorithmic
  Hiring}. In \bibinfo{booktitle}{\emph{Proceedings of the 19th International
  Conference on Autonomous Agents and MultiAgent Systems}}.
  \bibinfo{pages}{1716--1720}.
\newblock


\bibitem[\protect\citeauthoryear{Selbst and Barocas}{Selbst and
  Barocas}{2018}]%
        {selbst2018intuitive}
\bibfield{author}{\bibinfo{person}{Andrew~D Selbst} {and}
  \bibinfo{person}{Solon Barocas}.} \bibinfo{year}{2018}\natexlab{}.
\newblock \showarticletitle{The intuitive appeal of explainable machines}.
\newblock \bibinfo{journal}{\emph{Fordham L. Rev.}}  \bibinfo{volume}{87}
  (\bibinfo{year}{2018}), \bibinfo{pages}{1085}.
\newblock


\bibitem[\protect\citeauthoryear{Sharma, Henderson, and Ghosh}{Sharma
  et~al\mbox{.}}{2019}]%
        {sharma2019certifai}
\bibfield{author}{\bibinfo{person}{Shubham Sharma}, \bibinfo{person}{Jette
  Henderson}, {and} \bibinfo{person}{Joydeep Ghosh}.}
  \bibinfo{year}{2019}\natexlab{}.
\newblock \showarticletitle{{CERTIFAI}: Counterfactual Explanations for
  Robustness, Transparency, Interpretability, and Fairness of Artificial
  Intelligence models}.
\newblock \bibinfo{journal}{\emph{arXiv preprint arXiv:1905.07857}}
  (\bibinfo{year}{2019}).
\newblock


\bibitem[\protect\citeauthoryear{Shokri, Strobel, and Zick}{Shokri
  et~al\mbox{.}}{2019}]%
        {shokri2019privacy}
\bibfield{author}{\bibinfo{person}{Reza Shokri}, \bibinfo{person}{Martin
  Strobel}, {and} \bibinfo{person}{Yair Zick}.}
  \bibinfo{year}{2019}\natexlab{}.
\newblock \showarticletitle{Privacy risks of explaining machine learning
  models}.
\newblock \bibinfo{journal}{\emph{arXiv preprint arXiv:1907.00164}}
  (\bibinfo{year}{2019}).
\newblock


\bibitem[\protect\citeauthoryear{Sokol and Flach}{Sokol and Flach}{2018}]%
        {sokol2018conversational}
\bibfield{author}{\bibinfo{person}{Kacper Sokol} {and} \bibinfo{person}{Peter~A
  Flach}.} \bibinfo{year}{2018}\natexlab{}.
\newblock \showarticletitle{Conversational Explanations of Machine Learning
  Predictions Through Class-contrastive Counterfactual Statements.}. In
  \bibinfo{booktitle}{\emph{IJCAI}}. \bibinfo{pages}{5785--5786}.
\newblock


\bibitem[\protect\citeauthoryear{Sokol and Flach}{Sokol and Flach}{2019}]%
        {sokol2019counterfactual}
\bibfield{author}{\bibinfo{person}{Kacper Sokol} {and} \bibinfo{person}{Peter~A
  Flach}.} \bibinfo{year}{2019}\natexlab{}.
\newblock \showarticletitle{Counterfactual explanations of machine learning
  predictions: opportunities and challenges for AI safety}. In
  \bibinfo{booktitle}{\emph{SafeAI@ AAAI}}.
\newblock


\bibitem[\protect\citeauthoryear{Sra, Nowozin, and Wright}{Sra
  et~al\mbox{.}}{2012}]%
        {sra2012optimization}
\bibfield{author}{\bibinfo{person}{Suvrit Sra}, \bibinfo{person}{Sebastian
  Nowozin}, {and} \bibinfo{person}{Stephen~J Wright}.}
  \bibinfo{year}{2012}\natexlab{}.
\newblock \bibinfo{booktitle}{\emph{Optimization for machine learning}}.
\newblock \bibinfo{publisher}{Mit Press}.
\newblock


\bibitem[\protect\citeauthoryear{Szegedy, Zaremba, Sutskever, Bruna, Erhan,
  Goodfellow, and Fergus}{Szegedy et~al\mbox{.}}{2013}]%
        {szegedy2013intriguing}
\bibfield{author}{\bibinfo{person}{Christian Szegedy},
  \bibinfo{person}{Wojciech Zaremba}, \bibinfo{person}{Ilya Sutskever},
  \bibinfo{person}{Joan Bruna}, \bibinfo{person}{Dumitru Erhan},
  \bibinfo{person}{Ian Goodfellow}, {and} \bibinfo{person}{Rob Fergus}.}
  \bibinfo{year}{2013}\natexlab{}.
\newblock \showarticletitle{Intriguing properties of neural networks}.
\newblock \bibinfo{journal}{\emph{arXiv preprint arXiv:1312.6199}}
  (\bibinfo{year}{2013}).
\newblock


\bibitem[\protect\citeauthoryear{Tian and Pearl}{Tian and Pearl}{2001}]%
        {tian2001causal}
\bibfield{author}{\bibinfo{person}{Jin Tian} {and} \bibinfo{person}{Judea
  Pearl}.} \bibinfo{year}{2001}\natexlab{}.
\newblock \showarticletitle{Causal discovery from changes}. In
  \bibinfo{booktitle}{\emph{Proceedings of the Seventeenth conference on
  Uncertainty in artificial intelligence}}. \bibinfo{pages}{512--521}.
\newblock


\bibitem[\protect\citeauthoryear{Tolomei, Silvestri, Haines, and
  Lalmas}{Tolomei et~al\mbox{.}}{2017}]%
        {tolomei2017interpretable}
\bibfield{author}{\bibinfo{person}{Gabriele Tolomei}, \bibinfo{person}{Fabrizio
  Silvestri}, \bibinfo{person}{Andrew Haines}, {and} \bibinfo{person}{Mounia
  Lalmas}.} \bibinfo{year}{2017}\natexlab{}.
\newblock \showarticletitle{Interpretable predictions of tree-based ensembles
  via actionable feature tweaking}. In \bibinfo{booktitle}{\emph{Proceedings of
  the 23rd ACM SIGKDD international conference on knowledge discovery and data
  mining}}. \bibinfo{pages}{465--474}.
\newblock


\bibitem[\protect\citeauthoryear{Tram{\`e}r, Zhang, Juels, Reiter, and
  Ristenpart}{Tram{\`e}r et~al\mbox{.}}{2016}]%
        {tramer2016stealing}
\bibfield{author}{\bibinfo{person}{Florian Tram{\`e}r}, \bibinfo{person}{Fan
  Zhang}, \bibinfo{person}{Ari Juels}, \bibinfo{person}{Michael~K Reiter},
  {and} \bibinfo{person}{Thomas Ristenpart}.} \bibinfo{year}{2016}\natexlab{}.
\newblock \showarticletitle{Stealing machine learning models via prediction
  apis}. In \bibinfo{booktitle}{\emph{25th $\{$USENIX$\}$ Security Symposium
  ($\{$USENIX$\}$ Security 16)}}. \bibinfo{pages}{601--618}.
\newblock


\bibitem[\protect\citeauthoryear{Tsirtsis and Gomez-Rodriguez}{Tsirtsis and
  Gomez-Rodriguez}{2020}]%
        {tsirtsis2020decisions}
\bibfield{author}{\bibinfo{person}{Stratis Tsirtsis} {and}
  \bibinfo{person}{Manuel Gomez-Rodriguez}.} \bibinfo{year}{2020}\natexlab{}.
\newblock \showarticletitle{Decisions, Counterfactual Explanations and
  Strategic Behavior}.
\newblock \bibinfo{journal}{\emph{arXiv preprint arXiv:2002.04333}}
  (\bibinfo{year}{2020}).
\newblock


\bibitem[\protect\citeauthoryear{Ustun, Spangher, and Liu}{Ustun
  et~al\mbox{.}}{2019}]%
        {ustun2019actionable}
\bibfield{author}{\bibinfo{person}{Berk Ustun}, \bibinfo{person}{Alexander
  Spangher}, {and} \bibinfo{person}{Yang Liu}.}
  \bibinfo{year}{2019}\natexlab{}.
\newblock \showarticletitle{Actionable recourse in linear classification}. In
  \bibinfo{booktitle}{\emph{Proceedings of the Conference on Fairness,
  Accountability, and Transparency}}. ACM, \bibinfo{pages}{10--19}.
\newblock


\bibitem[\protect\citeauthoryear{van~der Waa, Robeer, van Diggelen, Brinkhuis,
  and Neerincx}{van~der Waa et~al\mbox{.}}{2018}]%
        {van2018contrastive}
\bibfield{author}{\bibinfo{person}{Jasper van~der Waa}, \bibinfo{person}{Marcel
  Robeer}, \bibinfo{person}{Jurriaan van Diggelen}, \bibinfo{person}{Matthieu
  Brinkhuis}, {and} \bibinfo{person}{Mark Neerincx}.}
  \bibinfo{year}{2018}\natexlab{}.
\newblock \showarticletitle{Contrastive explanations with local foil trees}.
\newblock \bibinfo{journal}{\emph{arXiv preprint arXiv:1806.07470}}
  (\bibinfo{year}{2018}).
\newblock


\bibitem[\protect\citeauthoryear{Van~Looveren and Klaise}{Van~Looveren and
  Klaise}{2019}]%
        {van2019interpretable}
\bibfield{author}{\bibinfo{person}{Arnaud Van~Looveren} {and}
  \bibinfo{person}{Janis Klaise}.} \bibinfo{year}{2019}\natexlab{}.
\newblock \showarticletitle{Interpretable Counterfactual Explanations Guided by
  Prototypes}.
\newblock \bibinfo{journal}{\emph{arXiv preprint arXiv:1907.02584}}
  (\bibinfo{year}{2019}).
\newblock


\bibitem[\protect\citeauthoryear{Venkatasubramanian and
  Alfano}{Venkatasubramanian and Alfano}{2020}]%
        {venkatasubramanianphilosophical}
\bibfield{author}{\bibinfo{person}{Suresh Venkatasubramanian} {and}
  \bibinfo{person}{Mark Alfano}.} \bibinfo{year}{2020}\natexlab{}.
\newblock \showarticletitle{The philosophical basis of algorithmic recourse}.
  In \bibinfo{booktitle}{\emph{Proceedings of the Conference on Fairness,
  Accountability, and Transparency}}. ACM.
\newblock


\bibitem[\protect\citeauthoryear{Vermeire and Martens}{Vermeire and
  Martens}{2020}]%
        {vermeire2020explainable}
\bibfield{author}{\bibinfo{person}{Tom Vermeire} {and} \bibinfo{person}{David
  Martens}.} \bibinfo{year}{2020}\natexlab{}.
\newblock \showarticletitle{Explainable Image Classification with Evidence
  Counterfactual}.
\newblock \bibinfo{journal}{\emph{arXiv preprint arXiv:2004.07511}}
  (\bibinfo{year}{2020}).
\newblock


\bibitem[\protect\citeauthoryear{Voigt and Von~dem Bussche}{Voigt and Von~dem
  Bussche}{[n.d.]}]%
        {voigt2017eu}
\bibfield{author}{\bibinfo{person}{Paul Voigt} {and} \bibinfo{person}{Axel
  Von~dem Bussche}.} \bibinfo{year}{[n.d.]}\natexlab{}.
\newblock \showarticletitle{The {EU} General Data Protection Regulation
  ({GDPR})}.
\newblock  (\bibinfo{year}{[n.\,d.]}).
\newblock


\bibitem[\protect\citeauthoryear{von K{\"u}gelgen, Karimi, Bhatt, Valera,
  Weller, and Sch{\"o}lkopf}{von K{\"u}gelgen et~al\mbox{.}}{2020}]%
        {von2020fairness}
\bibfield{author}{\bibinfo{person}{Julius von K{\"u}gelgen},
  \bibinfo{person}{Amir-Hossein Karimi}, \bibinfo{person}{Umang Bhatt},
  \bibinfo{person}{Isabel Valera}, \bibinfo{person}{Adrian Weller}, {and}
  \bibinfo{person}{Bernhard Sch{\"o}lkopf}.} \bibinfo{year}{2020}\natexlab{}.
\newblock \showarticletitle{On the Fairness of Causal Algorithmic Recourse}.
\newblock \bibinfo{journal}{\emph{arXiv preprint arXiv:2010.06529}}
  (\bibinfo{year}{2020}).
\newblock


\bibitem[\protect\citeauthoryear{Wachter, Mittelstadt, and Russell}{Wachter
  et~al\mbox{.}}{2017}]%
        {wachter2017counterfactual}
\bibfield{author}{\bibinfo{person}{Sandra Wachter}, \bibinfo{person}{Brent
  Mittelstadt}, {and} \bibinfo{person}{Chris Russell}.}
  \bibinfo{year}{2017}\natexlab{}.
\newblock \showarticletitle{Counterfactual explanations without opening the
  black box: Automated decisions and the {GDPR}}.
\newblock \bibinfo{journal}{\emph{Harvard Journal of Law \& Technology}}
  \bibinfo{volume}{31}, \bibinfo{number}{2} (\bibinfo{year}{2017}).
\newblock


\bibitem[\protect\citeauthoryear{Wallin}{Wallin}{1992}]%
        {wallin1992legal}
\bibfield{author}{\bibinfo{person}{David~E Wallin}.}
  \bibinfo{year}{1992}\natexlab{}.
\newblock \showarticletitle{Legal recourse and the demand for auditing}.
\newblock \bibinfo{journal}{\emph{Accounting Review}} (\bibinfo{year}{1992}),
  \bibinfo{pages}{121--147}.
\newblock


\bibitem[\protect\citeauthoryear{Wang and Gong}{Wang and Gong}{2018}]%
        {wang2018stealing}
\bibfield{author}{\bibinfo{person}{Binghui Wang} {and}
  \bibinfo{person}{Neil~Zhenqiang Gong}.} \bibinfo{year}{2018}\natexlab{}.
\newblock \showarticletitle{Stealing hyperparameters in machine learning}. In
  \bibinfo{booktitle}{\emph{2018 IEEE Symposium on Security and Privacy (SP)}}.
  IEEE, \bibinfo{pages}{36--52}.
\newblock


\bibitem[\protect\citeauthoryear{Wang and Vasconcelos}{Wang and
  Vasconcelos}{2020}]%
        {wang2020scout}
\bibfield{author}{\bibinfo{person}{Pei Wang} {and} \bibinfo{person}{Nuno
  Vasconcelos}.} \bibinfo{year}{2020}\natexlab{}.
\newblock \showarticletitle{SCOUT: Self-aware Discriminant Counterfactual
  Explanations}. In \bibinfo{booktitle}{\emph{Proceedings of the IEEE/CVF
  Conference on Computer Vision and Pattern Recognition}}.
  \bibinfo{pages}{8981--8990}.
\newblock


\bibitem[\protect\citeauthoryear{Weller}{Weller}{2017}]%
        {weller2017challenges}
\bibfield{author}{\bibinfo{person}{Adrian Weller}.}
  \bibinfo{year}{2017}\natexlab{}.
\newblock \showarticletitle{Challenges for transparency}.
\newblock \bibinfo{journal}{\emph{arXiv preprint arXiv:1708.01870}}
  (\bibinfo{year}{2017}).
\newblock


\bibitem[\protect\citeauthoryear{Wexler, Pushkarna, Bolukbasi, Wattenberg,
  Vi{\'e}gas, and Wilson}{Wexler et~al\mbox{.}}{2019}]%
        {wexler2019if}
\bibfield{author}{\bibinfo{person}{James Wexler}, \bibinfo{person}{Mahima
  Pushkarna}, \bibinfo{person}{Tolga Bolukbasi}, \bibinfo{person}{Martin
  Wattenberg}, \bibinfo{person}{Fernanda Vi{\'e}gas}, {and}
  \bibinfo{person}{Jimbo Wilson}.} \bibinfo{year}{2019}\natexlab{}.
\newblock \showarticletitle{The What-If Tool: Interactive Probing of Machine
  Learning Models}.
\newblock \bibinfo{journal}{\emph{IEEE transactions on visualization and
  computer graphics}} \bibinfo{volume}{26}, \bibinfo{number}{1}
  (\bibinfo{year}{2019}), \bibinfo{pages}{56--65}.
\newblock


\bibitem[\protect\citeauthoryear{White and Garcez}{White and Garcez}{2019}]%
        {white2019measurable}
\bibfield{author}{\bibinfo{person}{Adam White} {and}
  \bibinfo{person}{Artur~d'Avila Garcez}.} \bibinfo{year}{2019}\natexlab{}.
\newblock \showarticletitle{Measurable counterfactual local explanations for
  any classifier}.
\newblock \bibinfo{journal}{\emph{arXiv preprint arXiv:1908.03020}}
  (\bibinfo{year}{2019}).
\newblock


\bibitem[\protect\citeauthoryear{Whitley}{Whitley}{1994}]%
        {whitley1994genetic}
\bibfield{author}{\bibinfo{person}{Darrell Whitley}.}
  \bibinfo{year}{1994}\natexlab{}.
\newblock \showarticletitle{A genetic algorithm tutorial}.
\newblock \bibinfo{journal}{\emph{Statistics and computing}}
  \bibinfo{volume}{4}, \bibinfo{number}{2} (\bibinfo{year}{1994}),
  \bibinfo{pages}{65--85}.
\newblock


\bibitem[\protect\citeauthoryear{Woodward}{Woodward}{2005}]%
        {woodward2005making}
\bibfield{author}{\bibinfo{person}{James Woodward}.}
  \bibinfo{year}{2005}\natexlab{}.
\newblock \bibinfo{booktitle}{\emph{Making things happen: A theory of causal
  explanation}}.
\newblock \bibinfo{publisher}{Oxford university press}.
\newblock


\bibitem[\protect\citeauthoryear{Woodward}{Woodward}{2016}]%
        {sep-causation-mani}
\bibfield{author}{\bibinfo{person}{James Woodward}.}
  \bibinfo{year}{2016}\natexlab{}.
\newblock \showarticletitle{{Causation and Manipulability}}.
\newblock In \bibinfo{booktitle}{\emph{The {Stanford} Encyclopedia of
  Philosophy} (\bibinfo{edition}{winter 2016} ed.)},
  \bibfield{editor}{\bibinfo{person}{Edward~N. Zalta}} (Ed.).
  \bibinfo{publisher}{Metaphysics Research Lab, Stanford University}.
\newblock


\bibitem[\protect\citeauthoryear{Xie, Wu, Maaten, Yuille, and He}{Xie
  et~al\mbox{.}}{2019}]%
        {xie2019feature}
\bibfield{author}{\bibinfo{person}{Cihang Xie}, \bibinfo{person}{Yuxin Wu},
  \bibinfo{person}{Laurens van~der Maaten}, \bibinfo{person}{Alan~L Yuille},
  {and} \bibinfo{person}{Kaiming He}.} \bibinfo{year}{2019}\natexlab{}.
\newblock \showarticletitle{Feature denoising for improving adversarial
  robustness}. In \bibinfo{booktitle}{\emph{Proceedings of the IEEE/CVF
  Conference on Computer Vision and Pattern Recognition}}.
  \bibinfo{pages}{501--509}.
\newblock


\bibitem[\protect\citeauthoryear{Yang, Yin, Ling, and Pan}{Yang
  et~al\mbox{.}}{2006}]%
        {yang2006extracting}
\bibfield{author}{\bibinfo{person}{Qiang Yang}, \bibinfo{person}{Jie Yin},
  \bibinfo{person}{Charles Ling}, {and} \bibinfo{person}{Rong Pan}.}
  \bibinfo{year}{2006}\natexlab{}.
\newblock \showarticletitle{Extracting actionable knowledge from decision
  trees}.
\newblock \bibinfo{journal}{\emph{IEEE Transactions on Knowledge and data
  Engineering}} \bibinfo{volume}{19}, \bibinfo{number}{1}
  (\bibinfo{year}{2006}), \bibinfo{pages}{43--56}.
\newblock


\bibitem[\protect\citeauthoryear{Yeh and Lien}{Yeh and Lien}{2009}]%
        {yeh2009comparisons}
\bibfield{author}{\bibinfo{person}{I-Cheng Yeh} {and} \bibinfo{person}{Che-hui
  Lien}.} \bibinfo{year}{2009}\natexlab{}.
\newblock \showarticletitle{The comparisons of data mining techniques for the
  predictive accuracy of probability of default of credit card clients}.
\newblock \bibinfo{journal}{\emph{Expert Systems with Applications}}
  \bibinfo{volume}{36}, \bibinfo{number}{2} (\bibinfo{year}{2009}),
  \bibinfo{pages}{2473--2480}.
\newblock


\bibitem[\protect\citeauthoryear{Zhang, Solar-Lezama, and Singh}{Zhang
  et~al\mbox{.}}{2018}]%
        {zhang2018interpreting}
\bibfield{author}{\bibinfo{person}{Xin Zhang}, \bibinfo{person}{Armando
  Solar-Lezama}, {and} \bibinfo{person}{Rishabh Singh}.}
  \bibinfo{year}{2018}\natexlab{}.
\newblock \showarticletitle{Interpreting neural network judgments via minimal,
  stable, and symbolic corrections}. In \bibinfo{booktitle}{\emph{Advances in
  Neural Information Processing Systems}}. \bibinfo{pages}{4874--4885}.
\newblock


\bibitem[\protect\citeauthoryear{Zhao}{Zhao}{2020}]%
        {zhao2020fast}
\bibfield{author}{\bibinfo{person}{Yunxia Zhao}.}
  \bibinfo{year}{2020}\natexlab{}.
\newblock \showarticletitle{Fast Real-time Counterfactual Explanations}.
\newblock \bibinfo{journal}{\emph{arXiv preprint arXiv:2007.05684}}
  (\bibinfo{year}{2020}).
\newblock


\bibitem[\protect\citeauthoryear{Zitzler and Thiele}{Zitzler and
  Thiele}{1998}]%
        {zitzler1998evolutionary}
\bibfield{author}{\bibinfo{person}{Eckart Zitzler} {and}
  \bibinfo{person}{Lothar Thiele}.} \bibinfo{year}{1998}\natexlab{}.
\newblock \showarticletitle{An evolutionary algorithm for multiobjective
  optimization: The strength pareto approach}.
\newblock \bibinfo{journal}{\emph{TIK-report}}  \bibinfo{volume}{43}
  (\bibinfo{year}{1998}).
\newblock


\end{thebibliography}

\end{document}